\documentclass[10pt,twocolumn]{article} 
\usepackage[a4paper,total={6.5in, 9in}]{geometry}

\usepackage{glossaries}
\newacronym{ap}{AP}{Average Precision}
\newacronym{auc}{AUROC}{Area Under the Receiver Operating Curve}
\newacronym{ace}{ACE}{Automatic Concept Extraction}
\newacronym{ecg}{ECG}{electro cardiogram}
\newacronym{nmf}{NMF}{Non-negative Matrix Factorization}
\newacronym{pca}{PCA}{Principal Component Analysis}
\newacronym{ai}{AI}{Artificial Intelligence}
\newacronym{dl}{DL}{Deep Learning}
\newacronym{lof}{LOF}{Local Outlier Factor}
\newacronym{dnn}{DNN}{Deep Neural Network}
\newacronym{lrp}{LRP}{Layer-wise Relevance Propagation}
\newacronym{fda}{FDA}{Fisher Discriminant Analysis}
\newacronym{lsb}{LSB}{least significant bit}
\newacronym{xai}{XAI}{eXplainable Artificial Intelligence}
\newacronym{crp}{CRP}{Concept Relevance Propagation}
\newacronym{amax}{ActMax}{Activation Maximization}
\newacronym{rmax}{RelMax}{Relevance Maximization}
\newacronym{dora}{DORA}{Data-agnOstic Representation Analysis}
\newacronym{aoc}{AOC}{Area Over Curve}
\newacronym{conv}{Conv}{convolutional}
\newacronym{svm}{SVM}{Support Vector Machine}
\newacronym{gmm}{GMM}{Gaussian Mixture Model}
\newacronym{roi}{ROI}{Region of Interest}
\newacronym{lcrp}{L-CRP}{CRP for Localization Models}
\newacronym{leace}{LEACE}{LEAst-squares Concept Erasure}
\newacronym{rrr}{RRR}{Right for the Right Reason}
\newacronym{cdep}{CDEP}{Contextual Decomposition Explanation Penalization}
\newacronym{clarc}{ClArC}{Class Artifact Compensation}
\newacronym{aclarc}{\mbox{A-ClArC}}{Augmentive ClArC}
\newacronym{pclarc}{\mbox{P-ClArC}}{Projective ClArC}
\newacronym{rpclarc}{\mbox{rP-ClArC}}{reactive P-ClArC}
\newacronym{rrclarc}{RR-ClArC}{Right Reason ClArC}
\newacronym{ml}{ML}{Machine Learning}
\newacronym{iou}{IoU}{Intersection over Union}
\newacronym{cse}{CSE}{complete skin examination}
\newacronym{cav}{CAV}{Concept Activation Vector}
\newacronym{cam}{CAM}{Class Activation Maps}
\newacronym{car}{CAR}{Concept Activation Region}
\newacronym{pcx}{PCX}{Prototypical Concept-based eXplanations}
\newacronym{tcav}{TCAV}{Testing with CAV}
\newacronym{spray}{SpRAy}{Spectral Relevance Analysis}
\newacronym{svd}{SVD}{Singular Value Decomposition}
\newacronym{nams}{n-AMS}{natural Activation-Maximization signals}
\newacronym{sae}{SAE}{Sparse Autoencoder}
\newacronym{shap}{SHAP}{SHapley Additive exPlanations}
\newacronym{iterrev}{IterRev}{Iteratively Revealing and Revising Spurious Model Behavior}
\newacronym{r2r}{R2R}{Reveal to Revise}
\newacronym{xil}{XIL}{eXplanatory Interactive Learning}
\newacronym{sem}{SEM}{Standard Error of the Mean}
\newacronym{se}{SE}{Standard Error}
\newacronym{lvh}{LVH}{left ventricular hypertrophy}
\newacronym{tsne}{t-SNE}{t-Distributed Stochastic Neighbor Embedding}
\newacronym{umap}{UMAP}{Uniform Manifold Approximation and Projection}
\newacronym{vit}{ViT}{Vision Transformer}
\newacronym{slic}{SLIC}{simple linear iterative clustering}

\usepackage{amsmath,amsfonts,bm}

\def\eqref#1{equation~\ref{#1}}

\def\1{\bm{1}}

\DeclareMathAlphabet{\mathsfit}{\encodingdefault}{\sfdefault}{m}{sl}
\SetMathAlphabet{\mathsfit}{bold}{\encodingdefault}{\sfdefault}{bx}{n}

\DeclareMathOperator*{\argmin}{arg\,min}

\makeatletter
\DeclareRobustCommand\onedot{\futurelet\@let@token\@onedot}
\def\@onedot{\ifx\@let@token.\else.\null\fi\xspace}

\def\eg{\emph{e.g}\onedot} 
\def\ie{\emph{i.e}\onedot}

\def\wrt{w.r.t\onedot} 

\def\argmin{\text{argmin}}

\def\x{\mathbf{x}}
\newcommand{\lossrr}{\mathcal{L}_\text{RR}}
\newcommand{\ba}{\mathbf{a}}
\newcommand{\bA}{\mathbf{A}}

\newcommand{\bh}{\mathbf{h}}
\newcommand{\bhr}{\mathbf{hr}}

\newcommand{\bx}{\mathbf{x}}
\newcommand{\bv}{\mathbf{v}}

\newcommand{\bmask}{\mathbf{m}}

\newcommand{\bbs}{s_{\text{bias}}}
\newcommand{\bsact}{\bbs}
\newcommand{\bsrel}{\bbs^{\text{rel}}}
\newcommand{\hpat}{\bh^{\text{pat}}}

\newcommand{\Int}{\mathbb{N}}
\newcommand{\Real}{\mathbb{R}}
\newcommand{\bfunc}{\bm{f}}
\newcommand{\bfeat}{\bm{a}_l}
\newcommand{\bfunct}{\bm{\Tilde{f}}}
\newcommand{\bneuron}{\bm{n}_l^i}
\newcommand{\setlatent}{\mathcal{A}_l}

\newcommand{\matR}{\bm{R}_l}
\newcommand{\matDl}{\bm{D}_l}
\newcommand{\matE}{\bm{E}}

\newcommand{\fnemb}{\bm{emb}}
\newcommand{\fnpwdist}{\bm{d}_p}
\newcommand{\fnrel}{\bm{r}^c_l}

\usepackage[numbers]{natbib}
\usepackage{multibib}
\usepackage{hyperref}
\usepackage{url}
\usepackage{graphicx}
\usepackage{duckuments}
\usepackage{lmodern}
\usepackage{anyfontsize}
\usepackage{bookmark}
\usepackage{xspace}
\usepackage{multirow} 
\usepackage{booktabs} 
\usepackage{tikz}
\usepackage{multicol}
\usepackage{algorithm2e}

\newcommand{\mycirc}[1][black]{\textcolor{#1}{\ensuremath\bullet}}

\definecolor{myblue}{HTML}{1F77B4}
\definecolor{mypurple}{HTML}{8172B3}
\definecolor{myorange}{HTML}{FF7F0E}
\definecolor{myred}{HTML}{C44E52}
\definecolor{mygrey}{HTML}{808080}
\definecolor{mygreen}{HTML}{2CA02C}

\definecolor{myredcircle}{HTML}{a12828} 
\definecolor{mybluecircle}{HTML}{5880A0} 

\newcommand{\circlednum}[1]{%
    \hspace{-0.5pt}\tikz[baseline=(char.base)]{
        \node[shape=circle, draw=myredcircle, fill=myredcircle, inner sep=1pt, minimum size=6pt] (char) {\textbf{\textcolor{white}{\textsf{\fontsize{5.5}{5.5}\selectfont #1}}}};
    }\hspace{-3.5pt}
}

\newcommand{\circlednumblue}[1]{%
    \hspace{-0.5pt}\tikz[baseline=(char.base)]{
        \node[shape=circle, draw=mybluecircle, fill=mybluecircle, inner sep=1pt, minimum size=6pt] (char) {\textbf{\textcolor{white}{\textsf{\fontsize{5.5}{5.5}\selectfont #1}}}};
    }\hspace{-3.5pt}
}

\newcommand{\circlednumbluesmall}[1]{%
    \hspace{-0.25pt}\tikz[baseline=(char.base)]{
        \node[shape=circle, draw=mybluecircle, fill=mybluecircle, inner sep=1pt, minimum size=7pt] (char) {\textbf{\textcolor{white}{\textsf{\fontsize{4.5}{5}\selectfont #1}}}};
    }\hspace{-2.75pt}
}

\usepackage{hyperref}
\glsdisablehyper

\newcites{app}{Additional References Appendix}

\title{\textbf{Ensuring Medical AI Safety: Interpretability-Driven Detection and Mitigation of Spurious Model Behavior and Associated Data
}}
\author{
    Frederik Pahde\textsuperscript{\rm 1},
    Thomas Wiegand\textsuperscript{\rm 1,2,3},
    Sebastian Lapuschkin\textsuperscript{\rm 1,$\dagger$},
    Wojciech Samek\textsuperscript{\rm 1,2,3,$\dagger$}
}
\date{\small
    \textsuperscript{\rm 1}Fraunhofer Heinrich Hertz Institut, Berlin, Germany\\
    \textsuperscript{\rm 2} Technische Universität Berlin, Berlin, Germany\\
    \textsuperscript{\rm 3}Berlin Institute for the Foundations of Learning and Data (BIFOLD), Berlin, Germany\\
    \textsuperscript{\rm $\dagger$}corresponding authors: \texttt{\{wojciech.samek,sebastian.lapuschkin\}@hhi.fraunhofer.de}\\
}

\begin{document}
\maketitle







\begin{abstract}
Deep neural networks are increasingly employed in high-stakes medical applications, despite their tendency for shortcut learning in the presence of spurious correlations, which can have potentially fatal consequences in practice.
Whereas a multitude of works address either the detection or mitigation of such shortcut behavior in isolation, the Reveal2Revise approach provides a comprehensive bias mitigation framework combining these steps.
However, effectively addressing these biases often requires substantial labeling efforts from domain experts. 
In this work, we review the steps of the Reveal2Revise framework and enhance it with semi-automated interpretability-based bias annotation capabilities.
This includes methods for the sample- and 
feature-level bias annotation, providing valuable information for bias mitigation methods to unlearn the undesired shortcut behavior.
We show the applicability of the framework using four medical datasets across two modalities, featuring controlled and real-world spurious correlations caused by data artifacts.
We successfully identify and mitigate these biases in VGG16, ResNet50, and contemporary Vision Transformer models, ultimately increasing their robustness and applicability for real-world medical tasks. 
Our code is available at \url{https://github.com/frederikpahde/medical-ai-safety}.
\end{abstract}

\section{Introduction}
In the past decade, \gls{ml} models have become ubiquitous in medical applications, often outperforming human experts in tasks like melanoma detection~\citep{brinker2019deep} and the prediction of cardiovascular diseases from \gls{ecg} data~\citep{strodthoff2020deep}.
However, the non-transparent nature of \gls{dnn} predictions poses challenges in safety-critical contexts, as their reasoning remains obscure to both clinicians and model developers.
This opacity is concerning, particularly since \glspl{dnn} are prone to exploit spurious correlations in the training data.
This can lead to shortcut learning~\citep{geirhos2020shortcut}, where models rely on (medically) irrelevant features, yet correlating with the target label.
Such shortcuts are not limited to protected attributes like gender or ethnicity, but include various confounders in the training data, such as objects (\eg, rulers or hair), color shifts, or watermarks added by scanning devices.
A well-known example are band-aids in dermoscopic images for melanoma detection dominantly occurring alongside benign lesions, causing \gls{ml} models to associate the presence of band-aids with benign lesions with potentially severe consequences in practice.
Similarly, DNNs trained to detect pneumonia from radiographs have been known to predict the hospital system used for the scan, as the prevalence varied across hospitals in the study~\citep{zech2018variable}.
Moreover, confounding shortcuts were learned over intended signals from computed tomography scans for COVID-19 detection~\citep{degrave2021ai}.

\begin{figure*}[t!]
    \centering
     \includegraphics[width=.95\textwidth]{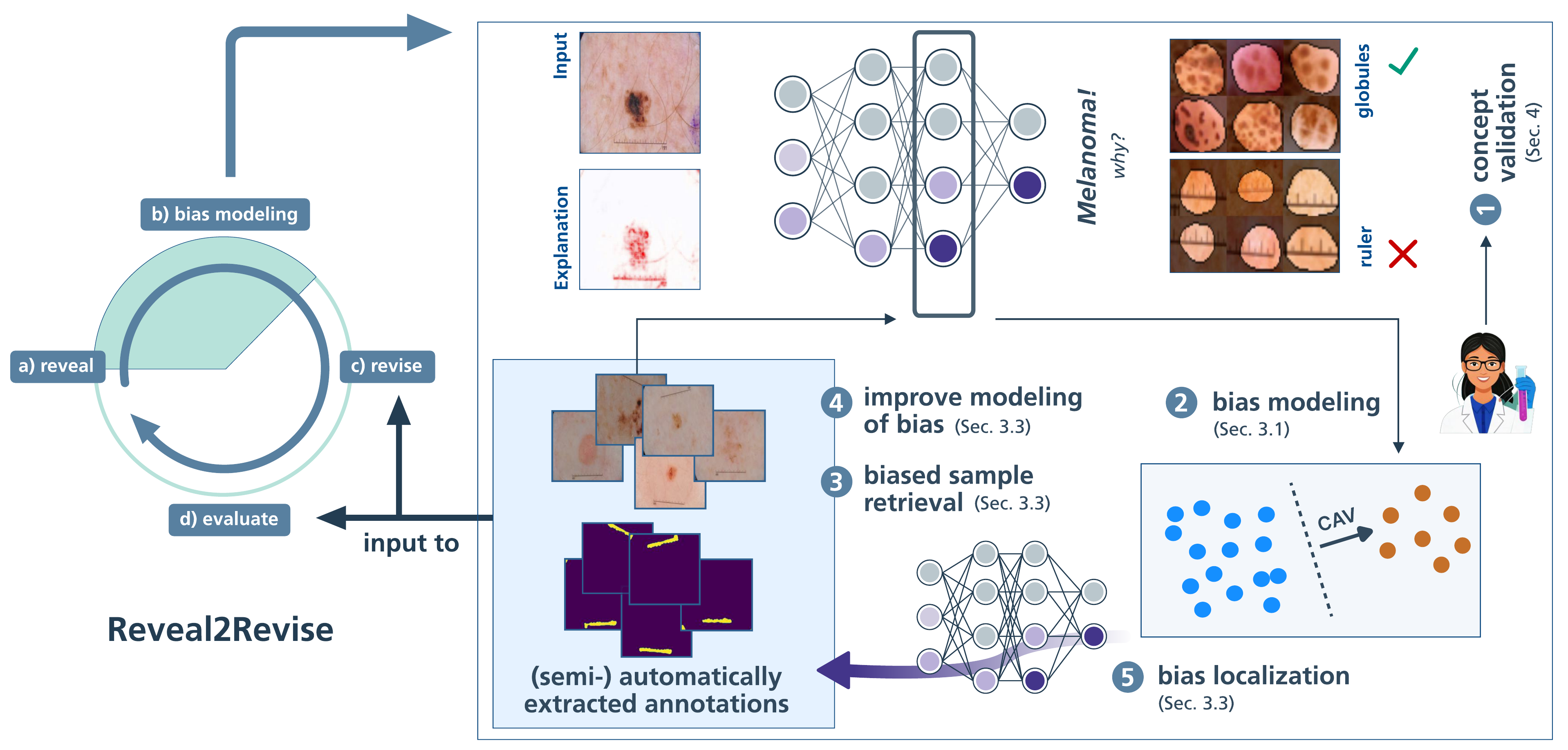}
    \caption{
    Extending the (a) reveal and (b) bias modeling steps of the Reveal2Revise framework, we demonstrate the power of \gls{xai} for bias detection, modeling, and data annotation.
    First, \mbox{(\protect\circlednumblue{1})}~bias identification approaches are leveraged to present outlier samples and concepts to model developers and domain experts for concept validation. 
    Identified samples representing biases can be used for \mbox{(\protect\circlednumblue{2})}~bias modeling using CAVs.
    This bias representation can be utilized for \mbox{(\protect\circlednumblue{3})}~the retrieval of biased samples, which, in turn, can be used in an \mbox{(\protect\circlednumblue{4})}~iterative process to improve the bias representation.
    Revised bias representations are further used for \mbox{(\protect\circlednumblue{5})}~spatial bias localization.
    These semi-automatically extracted annotations are input to the (c) revise and (d) re-evaluation steps of Reveal2Revise. 
    }
    \label{fig:title_figure}
\end{figure*}

The field of \gls{xai} sheds light onto the previously non-transparent prediction behavior of \glspl{dnn}, providing insights into their internal reasoning.
While traditional local \gls{xai} methods focus on feature importance for individual predictions, global \gls{xai} approaches 
aim to understand overall model behavior by explaining the roles of internal representations and encoded features~\citep{achtibat2023attribution,fel2023craft,zhang2021invertible}.
Recently, these insights have been utilized to uncover model weaknesses like shortcut behavior.
Current methods either detect outlier explanations for biased data samples~\citep{lapuschkin2019unmasking,anders2022finding,dreyer2024understanding} or outlier model concept representations~\citep{bykov2023dora,neuhaus2023spurious,pahde2023reveal}.
The purpose of this work is to review interpretability-driven shortcut detection and bias mitigation methods.
Specifically, we focus on comprehensive frameworks such as Reveal2Revise~\citep{pahde2023reveal}, which consists of the four steps (a) bias revealing, (b) bias modeling, \ie, learning accurate bias representations, (c) model revision to unlearn shortcuts, and (d) re-evaluation. 
While we review related work in the first three steps, we put special attention on the bias detection and modeling steps.
We further extend the framework with bias annotation capabilities to (semi-)automatically annotate and enrich datasets, leveraging insights from \gls{xai}, as shown in Fig.~\ref{fig:title_figure} (see Appendix~\ref{app:algorithm} for an algorithmic overview).
Specifically, we discuss bias identification from both data and model perspectives, enabling expert-guided validation of model behavior\footnote{Ideally, this is done by an interdisciplinary group consisting of model developers and domain experts.}~\mbox{(\circlednumblue{1})}. 
Moreover, we learn a model-internal bias representation, referred to as bias model, from an initial set of biased samples using \glspl{cav}~\mbox{(\circlednumblue{2})}. 
We then introduce the retrieval of biased data samples using the learned bias model~\mbox{(\circlednumblue{3})}, enabling its iterative improvement~\mbox{(\circlednumblue{4})}.
Further using the refined bias model for spatial bias localization~\mbox{(\circlednumblue{5})}, we enrich the dataset with extracted sample- and feature-level bias annotations, which are used in the bias mitigation and evaluation steps of the Reveal2Revise framework to improve the generalization capabilities of \gls{ml} models.

We demonstrate the applicability of the extended framework on four medical tasks across two modalities and provide annotations for the detected spurious concepts and data points\footnote{Both code and annotations are provided here: \texttt{https://github.com/frederikpahde/medical-ai-safety}}.
This includes image-based melanoma detection, the identification of gastrointestinal abnormalities, cardiomegaly prediction from chest radiographs, and cardiovascular disease prediction from \gls{ecg} data using VGG, ResNet, and \gls{vit} models. 
Utilizing Reveal2Revise, we identify and mitigate both intrinsic and artificially introduced biases across all datasets, and demonstrate the data annotation capabilities using concept-based bias representations, minimizing the need for human supervision to improve the validity and robustness of \gls{ml} models.

The paper is structured as follows: 
First, we summarize related work on comprehensive bias detection and mitigation frameworks in Sec.~\ref{sec:related_work}.
We then provide background on how (biased) concepts are encoded within \glspl{dnn} (Sec.~\ref{sec:bias_modeling}).
Our main contribution lies in utilizing these representations for sample- and feature-level bias annotation, specifically the (iterative) retrieval of biased samples (Sec.~\ref{sec:data_annotation}) and spatial bias localization (Sec.~\ref{sec:localization}).
We further discuss and categorize existing methods to detect (outlier) concepts and data points (Sec.~\ref{sec:bias_identification}), briefly describe existing bias mitigation methods (Sec.~\ref{sec:bias_mitigation}), and demonstrate the applicability of the extended Reveal2Revise framework in experiments (Sec.~\ref{sec:experiments}).
Lastly, we discuss limitations and conclusions in Secs.~\ref{sec:limitations}~and~\ref{sec:conclusions}, respectively.

\section{Related Bias Detection and Mitigation Frameworks}
\label{sec:related_work}
Many existing interpretability-based bias identification methods focus on structured data, which allows for clear concept specifications, particularly in tabular datasets~\citep{begley2020explainability,pradhan2022interpretable}. 
In contrast, works involving unstructured data such as images primarily concentrate on shortcut detection in pre-trained models using generic benchmark datasets like ImageNet~\citep{deng2009imagenet}.
One research directions aims to identify samples with outlier model behavior, measured via local attribution scores in input~\citep{lapuschkin2019unmasking} or latent~\citep{anders2022finding,dreyer2024understanding} space.
 Other methods seek to pinpoint spurious model representations like neurons~\citep{singla2022salient} or latent directions~\citep{neuhaus2023spurious}, and modify them for bias mitigation.
n a medical context, existing works 
study shortcut behavior related to sensitive attributes like gender or age by comparing the model performance in different sub-popluations~\citep{brown2023detecting}, 
manually annotate the dataset for spurious features~\citep{bissoto2020debiasing}, 
or define heuristics to automate the detection of specific artifacts~\citep{rieger2020interpretations}.
In contrast, our work emphasizes generic and comprehensive methods for bias detection and mitigation, such as Reveal2Revise. 
This iterative model correction framework consists of four steps: (1) \emph{Reveal} for bias identification, (2) bias modeling, (3) \emph{Revise} for bias mitigation, and (4) (re-)evaluation (see Appendix~\ref{app:algorithm}). 
Reveal2Revise enables the interpretability-based identification and mitigation of spurious correlations without requiring prior knowledge.
While the original Reveal2Revise work~\citep{pahde2023reveal} emphasizes bias mitigation (step 3), our paper reviews the interpretability-based bias identification and modeling (steps 1 and 2) in 
greater detail.
We further extend the framework with bias annotation techniques, enabling the (semi-)automated generation of sample- and feature-level bias annotations.
This significantly reduces the manual labeling efforts necessary for obtaining annotations required for bias mitigation methods. 
Similar to Reveal2Revise, \citet{schramowski2020making} provide a comprehensive framework that allows domain experts to interact with local explanations to align \gls{ml} models with scientifically expected behavior.

\section{From Bias Modeling to (Semi-)Automated Data Annotation}
\label{sec:bias_representation}
Both medically valid and biased concepts manifest in specific model components, such as individual neurons, model circuits, or directions in latent space.
In this section, we review methods from global \gls{xai} and mechanistic interpretability that aim to interpret model substructures, specifically those encoding biases stemming from data artifacts (step \mbox{(\circlednumblue{2})} in Fig.~\ref{fig:title_figure}).
We further utilize these model insights to extend the Reveal2Revise frameworks with bias annotation capabilities, including the detection of biased samples and (spatial) bias localization. 
Note that while we assume knowledge on the existence of biases in this section, the identification thereof (step \mbox{(\circlednumblue{1})} in Fig.~\ref{fig:title_figure}) is addressed in Sec.~\ref{sec:bias_identification}.

\paragraph{Considered Types of Data Artifacts}
In this work, we focus on data artifacts caused by spurious correlations, \ie, concepts unrelated to the (medical) task, yet correlating with the target label due to biases in the dataset curation process.
Whereas some artifacts are entirely irrelevant to the task, \eg, watermarks from medical devices, other artifacts can have a medical meaning but no causal impact on the predicted outcome, such as skin markers from dermatologists. 
We further distinguish between well-localized objects, such as band-aids or rulers, and non-localizable artifacts, \eg, slight color or brightness shifts caused by the usage of different medical scanners. 
Data artifacts can spatially overlap task-relevant information, such that masking out artifactual regions might remove important information.
Moreover, spurious features can be conceptually entangled with valid features. 
For example, in melanoma detection, model representations for specific color patterns indicative of lesions may be entangled with natural variations in skin tone.

\subsection{Background: Concept Representations in Neural Networks}

\label{sec:bias_modeling}
We define a \gls{dnn} as a function $\bfunc: \mathcal{X} \rightarrow \mathcal{Y}$ that maps input samples $\bx \in \mathcal{X}$ to target labels $y \in \mathcal{Y}$. 
We further assume that at any layer $l \in \Int$ with $m \in \Int$ neurons, $\bfunc$ can be split into a feature extractor $\bfeat: \mathcal{X} \rightarrow \mathbb{R}^m$, computing latent activations at layer $l$,
and a classifier head $\bfunct: \mathbb{R}^m \rightarrow \mathcal{Y}$, mapping latent activations to target labels.
Neurons in layer $l$ are denoted as $\bneuron$ with $i \in \Int$ indexing the neuron position in the respective layer.
We further assume the existence of binary (bias) concept labels $t \in \{0,1\}$.

\paragraph{Representing Concepts with Individual Neurons}
\label{sec:global_xai_neurons}
A common assumption is that neurons in robust models encode human-aligned concepts, particularly at layers close to the model head~\citep{olah2017feature,radford2017learning,bau2020understanding}. 
Hence, there might exist a neuron $\bneuron$ acting as feature extractor for a biased concept.
Various feature visualization approaches aim to globally explain the concept represented by a neuron by identifying inputs that maximally trigger the neuron.
Whereas one line of work generates inputs that maximally activate the selected neuron~\citep{erhan2009visualizing,olah2017feature,fel2024unlocking}, other approaches select natural images from a reference dataset, \eg, the training set. 
Specifically, while \gls{amax}~\citep{szegedy2014intriguing,borowski2020natural} selects samples that maximally \emph{activate} a given neuron, \gls{rmax}~\citep{achtibat2023attribution} selects samples for which the neuron is maximally \emph{relevant} for the classification task, as computed by a local explainability methods. 
In contrast to activations, the relevance scores are directly linked to the model prediction,  indicating the neuron's impact on a specified target label.
However, limitations of the mapping of concepts to individual neurons are redundancy, \ie, multiple neurons representing the same concept~\citep{denil2013predicting}, and polysemanticity~\citep{fong2018net2vec,olah2020zoom,elhage2022toy,dreyer2024pure}, \ie, neurons reacting to multiple, seemingly unrelated concepts.
Recent works aim to overcome these challenges by \emph{disentangling} learned concepts via \glspl{sae}~\citep{huben2023sparse,bricken2023towards}. 
Assuming there are more concepts than neurons, \glspl{sae} leverage sparse dictionary learning to find an overcomplete feature basis, allowing the usage of encoder neurons as monosemantic concept representation.

\paragraph{Representing Concepts with Directions}
\label{sec:global_xai_directions}
Given the aforementioned limitations of neurons and the fact that there are typically more concepts than neurons, it is assumed that concepts are encoded as linear combinations of neurons, \ie, directions in latent space, referred to as \emph{superposition}~\citep{olah2020zoom,elhage2022toy}. 
As an alternative to disentangling the latent space via \glspl{sae}, these directions can be obtained directly, either in supervised or unsupervised fashion, as outlined below. 
Notably, using linear directions does not require knowledge on the role of neurons and allows the representation of \emph{any} concept, even those not extracted by single neurons.

\begin{figure*}[t!]
    \centering
    \includegraphics[width=.26\textwidth]{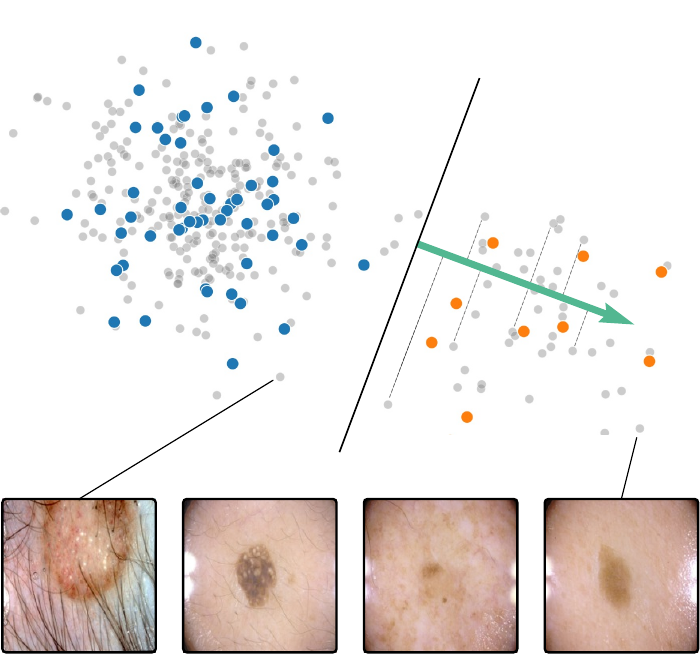}
    \quad\quad\quad\quad\quad\quad
    \includegraphics[width=.45\textwidth]{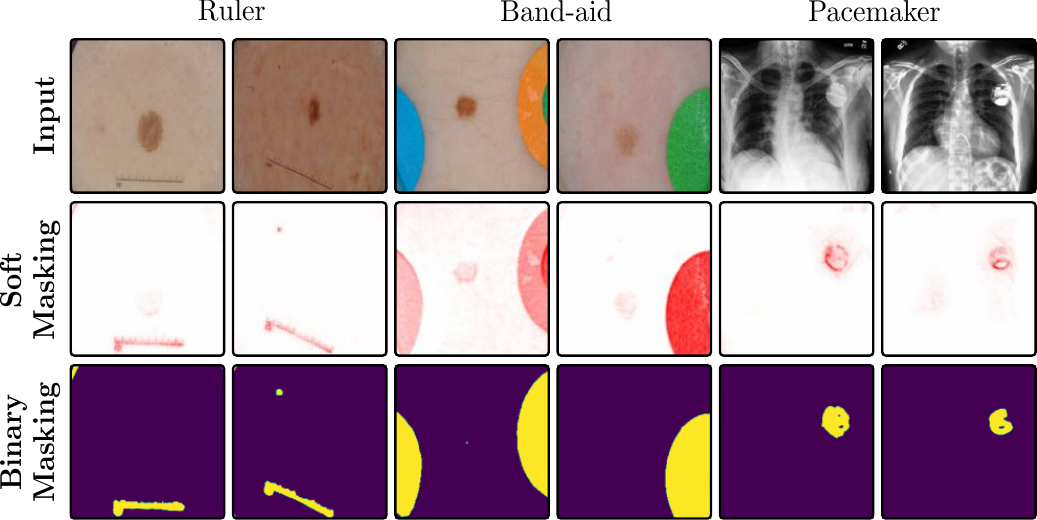}
    \caption{
    \emph{Left:} Usage of a \gls{cav} encoding the \texttt{reflection} concept, trained on known artifact ($\mycirc[myorange]$) and non-artifact ($\mycirc[myblue]$) samples, for the annotation of unknown ($\mycirc[mygrey]$) samples.
    We rank samples by their bias score, computed by projecting their activation onto the CAV $\bh_l$.
    \emph{Right:} Localization of data artifacts by computing relevance heatmaps for the \gls{cav} $\bh_l$ for soft masking and their binarization.
    }
    \label{fig:bias_modeling}
\end{figure*}

\textbf{Supervised Concept Estimation:}
When concept labels $t$ indicating the presence of a bias are known, concept directions can be estimated via Linear Probing~\citep{alain2016understanding,belinkov2022probing}. 
As such, \glspl{cav}, introduced for concept sensitivity testing~\citep{kim2018interpretability}, 
were originally computed as the weight vector $\mathbf{w}\in \Real^m$ from a linear classifier like a linear SVM, distinguishing latent activations $\setlatent^+$ on layer $l$ of samples with the concept from activations $\setlatent^-$ of samples without it.
However, recent work has shown that \glspl{cav} derived from linear classifiers can be influenced by distractors in the data, leading to inaccurate estimates of the concept \emph{signal} direction~\citep{pahde2025navigating}.
Instead of maximizing class separability, which is considered as the wrong optimization target, the work proposes Pattern-CAVs as robust alternative, 
as detailed in Appendix~\ref{app:sec:bias_modeling}.
\glspl{cav} can be computed on latent activations of arbitrary input shape, such as 3D representations in $\Real^{m\times h\times w}$, with $h\in\Real$ and $w\in\Real$ representing the height and width dimensions, respectively, as well as aggregated representations. 
This includes translation-invariant 1D or channel-invariant 2D representations by using max- or average pooling over spatial or channel dimensions~\citep{mikriukov2023evaluating}.
In this work, we use 1D-\glspl{cav} with max-pooling over spatial dimensions.
Beyond linear concept representations, Concept Activation Regions have been proposed~\citep{crabbe2022concept}, allowing for non-linear concept representations through a radial kernel function.

\textbf{Unsupervised Concept Discovery:}
When concept labels are not available, concepts can be discovered in unsupervised fashion.
An early approach, \gls{ace}~\citep{ghorbani2019towards}, extracts concepts by segmenting images into regions and clustering similar regions to identify potential visual concepts. 
In contrast, more recent approaches leverage matrix decomposition methods, such as \gls{pca}, \gls{svd}, or \gls{nmf}, on latent activations to identify meaningful directions in the model's latent space~\citep{zhang2021invertible,fel2023craft,graziani2023concept,neuhaus2023spurious}.
This leads to two matrices, with one matrix reinterpreted as the concept basis, \ie, each row can be considered as \gls{cav}, and the other matrix as the activations within that new basis~\citep{fel2024holistic}.
Another work proposes the discovery of linear subspaces as concept representations~\citep{vielhaben2023multi}.
However, unsupervised concept discovery for representing data artifacts has two drawbacks: 
First, while no labeling efforts are needed to find concept labels, manual inspection is required to determine which direction(s) accurately represent the desired concept, \ie, the artifact to be modeled. 
Second, matrix factorization approaches yield statistical groupings without guidance, such that there is no guarantee for the existence of a direction representing the targeted artifact concept.

\begin{figure*}[t!]
    \centering
    \includegraphics[width=.8\textwidth]{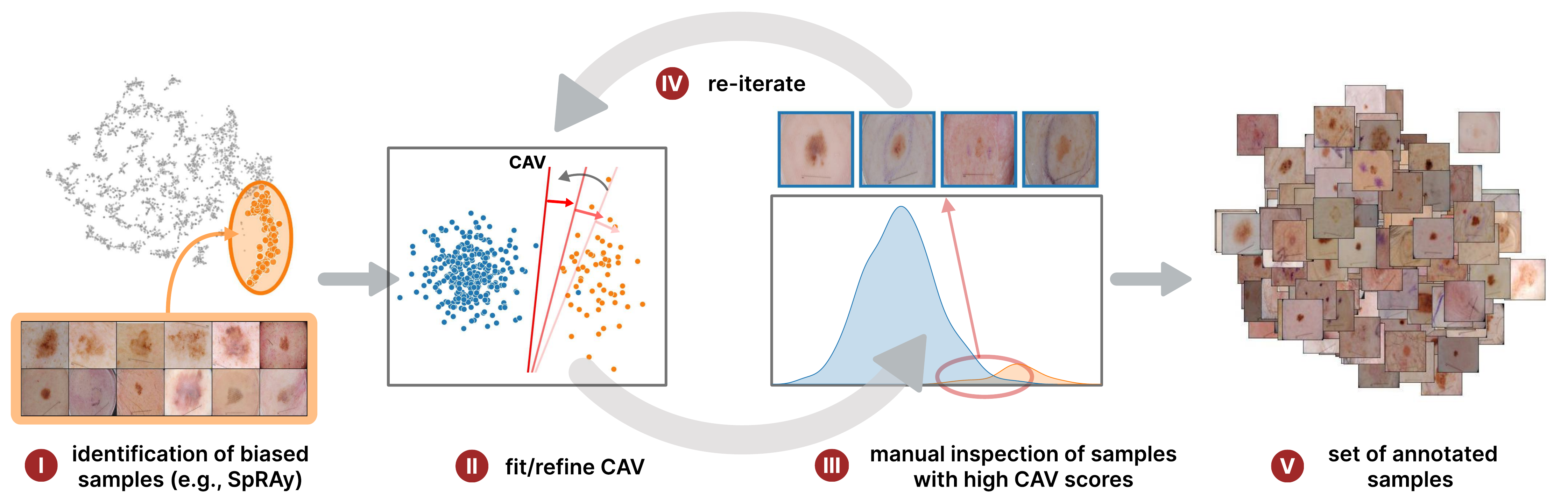}
    \caption{
    Iterative data annotation: (\protect\circlednum{I}) Given a small set of biased samples obtained via bias identification approaches, (\protect\circlednum{II}) a first \gls{cav} can be fitted.
    Using this CAV, (\protect\circlednum{III}) samples with high bias scores are subject to manual inspection to improve label quality.
    In an iterative process (\protect\circlednum{IV}), \glspl{cav} are refitted, and the manual inspection is repeated,  leading to an (\protect\circlednum{V}) improved set of annotated samples. 
    }    
    \label{fig:iterative_data_annotation}
\end{figure*}

\subsection{Sample-Level Bias Annotation via Data Retrieval}
\label{sec:data_annotation}
With a precise bias representation via \gls{cav} $\bh_l$ or neuron
$\bneuron$, the detection of artifactual samples can be further automated (step \mbox{(\circlednumblue{3})} in Fig.~\ref{fig:title_figure}).
Specifically, all samples from a dataset can be ranked by their similarity to the artifact representation, computed for example via cosine similarity, and presented to human annotators in that order, significantly supporting them in detecting artifact samples. 
Therefore, the data annotation process for concept representations based on a single neuron $\bneuron$, as for example suggested by \citet{singla2022salient}, is similar to global \gls{xai} methods explaining individual neurons via \gls{amax} or \gls{rmax}, which retrieve reference samples with maximal activation or relevance for the given neuron.
However, the limitations discussed above, namely redundancy and polysemanticity, affect the concept detection capabilities of individual neurons. 
To address this, we propose the extension of the artifact representation to linear directions in latent space via \glspl{cav}.
Specifically, given \gls{cav} $\bh_l$ and sample $\x \in \mathcal{X}$, we can compute a bias score $\bsact\in\Real$ by projecting latent activations $\bfeat(\x)$ for layer $l$ onto the \gls{cav}:

\begin{equation}
    \label{eq:annotation_act}
    \bsact=\bh_l^{\top}\bfeat(\x) ~\text{.}
\end{equation} 
High bias scores indicate a higher likelihood for the presence of the modeled bias.
Alternatively, inspired by \gls{rmax}, we can compute bias scores using \emph{relevance} scores instead of \emph{activations}, as outlined in Appendix~\ref{app:biased_sample_retrieval}.
Since relevance scores are computed class-specifically, this approach allows distinguishing concepts that are artifactual for certain classes but valid for others. 
An illustration is provided in Fig.~\ref{fig:bias_modeling} (\emph{left}), where a CAV has been trained to distinguish between known artifact ($\mycirc[myorange]$) and non-artifact samples ($\mycirc[myblue]$). This concept representation can subsequently be used to compute concept scores $\bsact$ for unknown samples ($\mycirc[mygrey]$), supporting annotators in the detection of further artifact samples.

\paragraph{Iterative Refinement of Bias Model}
Concept representations can be refined iteratively by correcting labeling errors in the data (step \mbox{(\circlednumblue{4})} in Fig.~\ref{fig:title_figure}).
Specifically, non-artifactual and unknown samples with \emph{high} bias scores are subject to manual inspection to improve the label quality and concept representation, as shown in Fig.~\ref{fig:iterative_data_annotation}.
Starting with a small set of bias samples obtained from bias identification methods~\mbox{(\circlednum{I})}, an initial \gls{cav} is fitted~\mbox{(\circlednum{II})}.
Next, manual inspection of samples with high bias scores improves the label quality~\mbox{(\circlednum{III})}.
The updated labels are used to iteratively refine the \gls{cav}~\mbox{(\circlednum{IV})}, resulting in a set of annotated bias samples~\mbox{(\circlednum{V})}.

\subsection{Feature-Level Bias Annotation via Spatial Localization}
\label{sec:localization}
Beyond detecting artifact samples, \gls{xai} insights can further reduce human labeling efforts by automating the \emph{spatial localization} of biased (and localizable) concepts within these samples (step \mbox{(\circlednumblue{5})} in Fig.~\ref{fig:title_figure}).
We assume the existence of a bias representation via \gls{cav} $\bh_l$ or neuron
$\bneuron$.
The latter can be represented as \gls{cav} $\bh_l$ by setting all values to zero, except for the one for neuron $\bneuron$, which is set to one. 
The targeted concept can then be localized in input space using local attribution methods, such as \gls{lrp}~\citep{bach2015pixel}.
Assuming singular neurons as concept representation, \citet{singla2022salient} leverage Class Activation Maps~\citep{zhou2016learning} to visualize the feature map for the given channel in input space and \citet{achtibat2023attribution} mask latent relevances for local attribution methods to compute channel-specific input heatmaps. 
This approach can be extended to biases represented as direction in latent space, \eg, using PCA \citep{neuhaus2023spurious} or \glspl{cav} \citep{anders2022finding,pahde2023reveal,de2024visual}.
Specifically, given the concept direction $\bh_l$ and latent activations $\bfeat(\bx)$, we can utilize local explanation approaches like \gls{lrp}~\citep{bach2015pixel} to explain the prediction of the function producing the bias score $\bsact$ as defined in Eq.~\ref{eq:annotation_act}.
This is equivalent to a 
local attribution method applied to sample $\bx$, with latent relevance scores $\matR(\bx)$ initialized as
\begin{equation}
    \label{eq:localization}
    \matR(\bx)=\bfeat(\bx) \circ \bh_l~\text{,}
\end{equation}
where $\circ$ denotes element-wise multiplication. 
This yields a heatmap that can be used as a soft mask or converted to a binary segmentation mask, for instance using thresholding techniques like Otsu's method~\citep{otsu1975threshold}.
An example is shown in Fig.~\ref{fig:bias_modeling} (\emph{right}), showing soft masks as heatmaps from concept-conditioned local attribution scores, along with binary masks for three known data artifacts: rulers and band-aids for skin cancer detection, and pacemakers in chest radiographs.

Notably, automated bias localization has the potential to substantially reduce manual feature-level annotation efforts.
These additional annotations can be utilized for various applications, \eg, as input for bias mitigation approaches requiring prior information on the bias to be unlearnt~(see Sec.~\ref{sec:bias_mitigation}).
Other applications include the design of metrics to measure artifact reliance
or the spatial isolation of the bias to copy-paste it onto non-artifactual samples to measure the model's sensitivity is towards the insertion of the artifact~\citep{pahde2023reveal}.

\section{Concept Validation: Detecting Spurious Behavior}
\label{sec:bias_identification}
Given the large number of model parameters, detecting biased model representations can be like searching for a needle in a haystack, especially without prior knowledge of spurious correlations.
To address this challenge, this section reviews and categorizes interpretability-driven approaches for detecting biased model behavior.
A common strategy is to identify outlier model behavior using a reference dataset.
Automated detection approaches typically focus on either analyzing post-hoc explanations for a set of reference images to find anomalous model behavior~\citep{lapuschkin2019unmasking,dreyer2024understanding} or identifying outlier representations within the model itself~\citep{bykov2023dora,neuhaus2023spurious}. 
For concept validation (step \mbox{(\circlednumblue{1})} in Fig.~\ref{fig:title_figure}), we distinguish between the \emph{data perspective} in Sec.~\ref{sec:detection:samples}, which focuses on detecting samples exhibiting outlier behavior, and the \emph{model perspective} in Sec.~\ref{sec:detection:concepts}, which aims to identify outlier concept representations within the model.
However, it is to note that outlier model reasoning is not necessarily caused by spurious correlations, but can be (clinically) valid model behavior that is rarely used.
Thus, detecting spurious correlations often requires manual inspection by domain experts to determine whether outlier behavior is valid or caused by spurious correlations.

\subsection{Data Perspective -- Detecting Spurious Samples}
\label{sec:detection:samples}
A first line of works assumes that models use a different behavior for spurious samples compared to ``clean'' samples. 
Concretely, model behavior can be estimated using local attribution methods, such as Input Gradients~\citep{morch1995visualization,simonyan2014very}, GradCAM~\citep{selvaraju2017grad}, or \gls{lrp}~\citep{bach2015pixel}. 
Note, that backpropagation-based attribution approaches distribute relevance scores from the output through all layers to the input, enabling the analysis of both \emph{input} heatmaps and \emph{latent} relevance scores.
This allows the analysis of prediction behavior at different abstraction levels, represented as relevance scores $\matR \in \Real^{n \times m  \times h \times w }$ for $n\in\Int$ samples in layer $l$ with $m$ channels and spatial dimension $h \times w$, or (spatially) aggregated representations. Using a distance function, such as cosine distance, we compute a pairwise distance matrix $\matDl \in \Real^{n \times n}$  for further inspection.

\paragraph{Analyzing Input Heatmaps} 
\gls{spray} groups input heatmaps using spectral clustering to identify outlier explanations that are likely to stem from spurious correlations~\citep{lapuschkin2019unmasking}.
This technique has uncovered various spurious correlations, such as the influence of photographers' watermarks for horse detection or other artifacts.
However, it is limited to well-localized biases that result in high relevance in similar spatial locations.

\paragraph{Analyzing Latent Relevances}
These limitations can be mitigated by applying \gls{spray} to relevances in \emph{latent} space~\citep{anders2022finding}.
Conveniently, many local attribution methods backpropagate relevance scores from the model output to the input, yielding scores for each neuron that indicate the importance of features extracted by those neurons.
Applying \gls{spray} allows clustering of latent relevance scores to identify typical and atypical model behavior, as shown in clinical gait analysis~\citep{slijepcevic2021explaining}. 
Note, that the clustering can be performed using relevance scores of shape $m \times h \times w$, as done by \citet{anders2022finding}, or in spatially aggregated manner using max- or average-pooling.
For example, \gls{pcx}~\citep{dreyer2024understanding} train Gaussian Mixture Models on max-pooled latent relevance scores and consider cluster means as stereotypical explanation, encoded as distribution over concepts (\ie, neurons).
\newline

\begin{figure*}[t!]
    \centering
    \includegraphics[width=.8\textwidth]{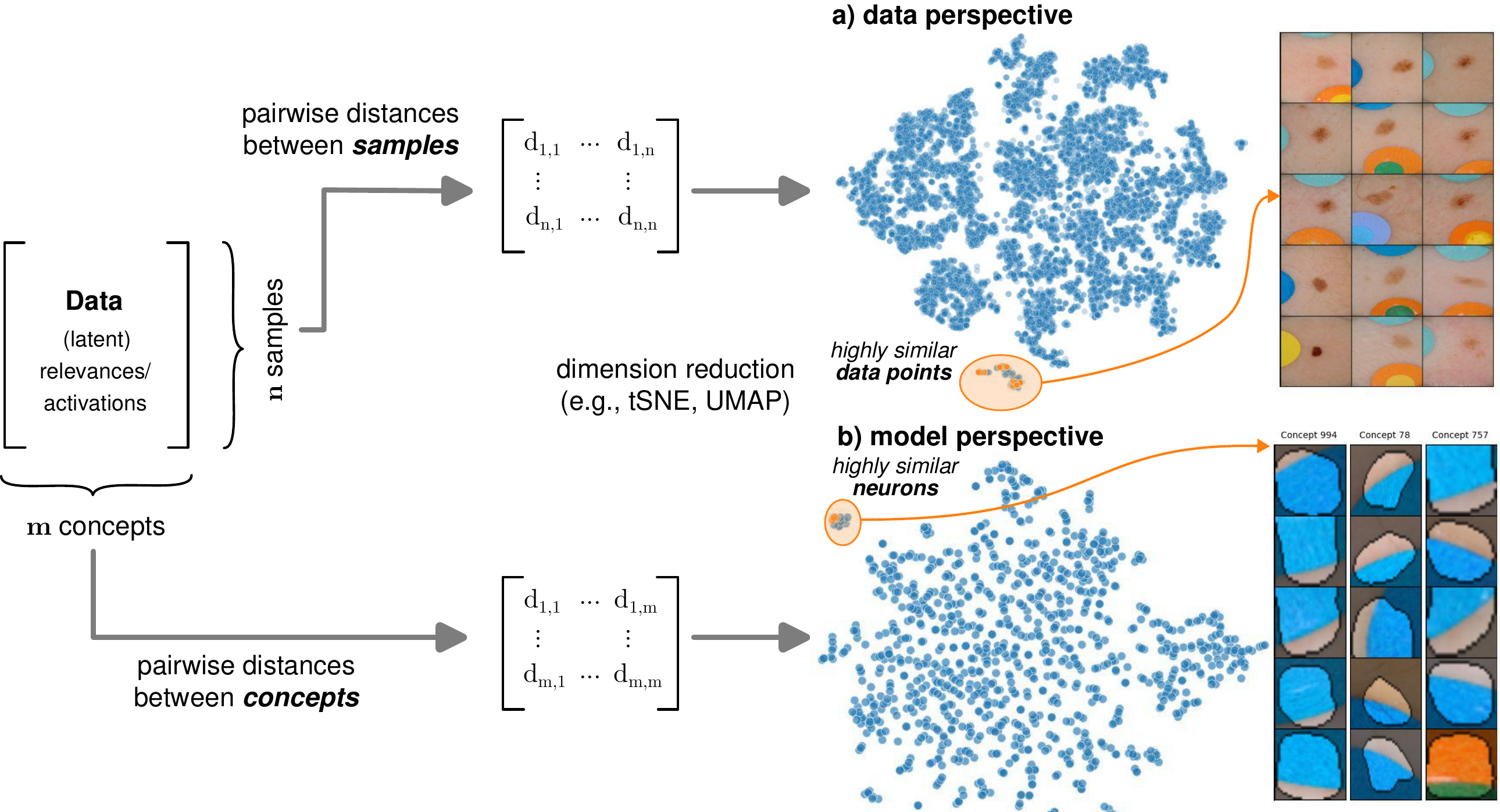} 
    \caption{
    Spurious correlations are identified by detecting outliers in model behavior. 
    This involves computing pairwise distances of (latent) activations or relevances from either the data (\emph{top}) or model (\emph{bottom}) perspective.
    The resulting $n \times n$ (data perspective) or $m \times m$ (model perspective) distance matrix is then reduced to 2D for visualization and outlier detection. 
    Human annotators determine whether detected outliers pose spurious correlations or valid prediction behavior. 
    }
    \label{fig:bias_detection}
\end{figure*}

Both input and latent relevance clustering require subsequent human supervision to determine whether outlier clusters represent valid or spurious behavior. To semi-automate this process, \citet{anders2022finding} propose using Fisher Discriminant Analysis~\citep{fisher1936use} to rank class-wise clusterings by linear separability, while \citet{dreyer2024understanding} computes similarities between prototypes.
The results of clustering approaches can serve as an initial set for bias modeling methods outlined in Sec.~\ref{sec:data_annotation}, which can be refined iteratively.
An example outlier cluster of latent relevances is shown in \mbox{Fig.~\ref{fig:bias_detection} (\emph{top})}, with all samples containing the spurious band-aid concept.

\subsection{Model Perspective -- Detecting Spurious Representations}
\label{sec:detection:concepts}
In contrast to detecting spurious \emph{samples}, recent work focuses on identifying spurious \emph{model internals} directly.
This aligns with mechanistic interpretability, which seeks to decipher the internal mechanics of \glspl{dnn}~\citep{olah2020zoom,elhage2022toy,bricken2023towards}.
As outlined in Sec.~\ref{sec:bias_modeling}, various global \gls{xai} methods aim to explain the role of individual neurons, and these insights can be leveraged to detect spurious model internals by clustering learned concepts and identifying outliers. 
Given input data $\mathcal{X}$ with $n$ samples, \citet{pahde2023reveal} compute spatially aggregated relevances $\matR \in \mathbb{R}^{n \times m}$.
Subsequently, they compute the pairwise cosine distance per column (\ie, channel/concept) and embed the resulting distance matrix $\matDl \in \mathbb{R}^{m \times m}$ in a low-dimensional space using dimension reduction techniques like \gls{tsne} \citep{van2008visualizing} or \gls{umap} \citep{mcinnes2018umap}.
This low-dimensional embedding can be visualized to identify outliers through human inspection or anomaly detection algorithms, such as the Local Outlier Factor~\citep{breunig2000lof}.
In summary, outlier representations can be identified in an embedding representation $\matE \in \Real^{m \times k}$, obtained as

\begin{equation}
\label{eq:cluster_concepts}
    \matE=\fnemb(\fnpwdist(\matDl))
\end{equation}
where $\fnemb(\cdot): \Real^{m \times m} \rightarrow \Real^{m \times k}$ reduces the dimension to $k \in \Int$ with $k \ll m$, and the pairwise distance function $\fnpwdist(\cdot)$ is applied along all channel dimensions in the latent representation $\matDl$, either given by activations or relevance scores for layer~$l$.
Note that this approach assumes over-parameterization resulting in redundant neurons, allowing to distinguish between similar and dissimilar concept representations.
An example is shown in \mbox{Fig.~\ref{fig:bias_detection} (\emph{bottom})}, where latent relevance scores from a ResNet50 model trained for melanoma detection are used to identify outlier concepts, specifically a cluster focusing on band-aids rather than clinically relevant features.
Notably, Eq.~\ref{eq:cluster_concepts} can easily be extended to find outlier \emph{directions} instead of neurons.
Specifically, this involves a linear transformation of latent representations $\matDl$ using the directions of interest, \eg, obtained in unsupervised manner as described in Sec.~\ref{sec:global_xai_directions}.

Similarly, \gls{dora} embeds a pairwise distance matrix of neuron representations into 2D, but proposes a data-agnostic approach and a tailored distance function~\citep{bykov2023dora}.
This involves generating 
\gls{amax} samples as concept representation for neurons, referred to as \gls{nams}. 
Each neuron $\bneuron$ is represented by a representation activation vector $\bv_i \in \Real^m$, measuring how much \emph{other} neurons activate on the given \gls{nams}, and compute pairwise distances between the vectors.
The resulting distance matrix $\matDl \in \Real^{m \times m}$ is embedded into lower dimension to identify outlier representations.
Notably, instead of generating \gls{amax} samples, this approach can also be applied on \emph{real} samples from a reference dataset.

Lastly, \citet{neuhaus2023spurious} use human supervision to identify spurious concepts represented as linear directions in latent activations obtained via \gls{pca}. 
To reduce manual labelling efforts, they propose an automated pre-selection of concept representations by focusing on the top 128 PCA components and raking them based on the model's confidence in classifying reference samples in the given direction.
Another promising direction is auto-labelling neurons, \eg, via foundation models, to search for expected valid or spurious concepts and to analyze unexpected concepts~\citep{hernandez2021natural,oikarinen2023clip,dreyer2025semanticlens,bykov2024labeling}.

\section{Bias Mitigation Methods}
\label{sec:bias_mitigation}
After detecting and annotating biases, our goal is to unlearn undesired behaviors. Therefore, we briefly review existing bias mitigation approaches in this section (see Appendix~\ref{app:bias_mitigation} for a more detailed overview).
A first line of approaches modifies the training data, \eg, by removing or manipulating biased samples, followed by retraining the model~\citep{wu2023discover,weng2025fast}.
While effectively mitigating biases, this method requires access to the full training set, can be costly, and may ignore valuable information, leading to practical limitations.
In an iterative model development life cycle like Reveal2Revise, with continuous model validation and bias mitigation, full re-training is often infeasible. 
Thus, we focus on efficient bias mitigation approaches, that either finetune the (biased) model with a custom loss function or modify the model post-hoc without additional training.
As such, \gls{rrr}~\citep{ross2017right} penalizes the alignment between the input gradient, \ie, the gradient of the prediction \wrt the input features, and ground truth masks localizing the artifact. 

\begin{figure*}[t!]
    \centering
    \includegraphics[width=.8\textwidth]{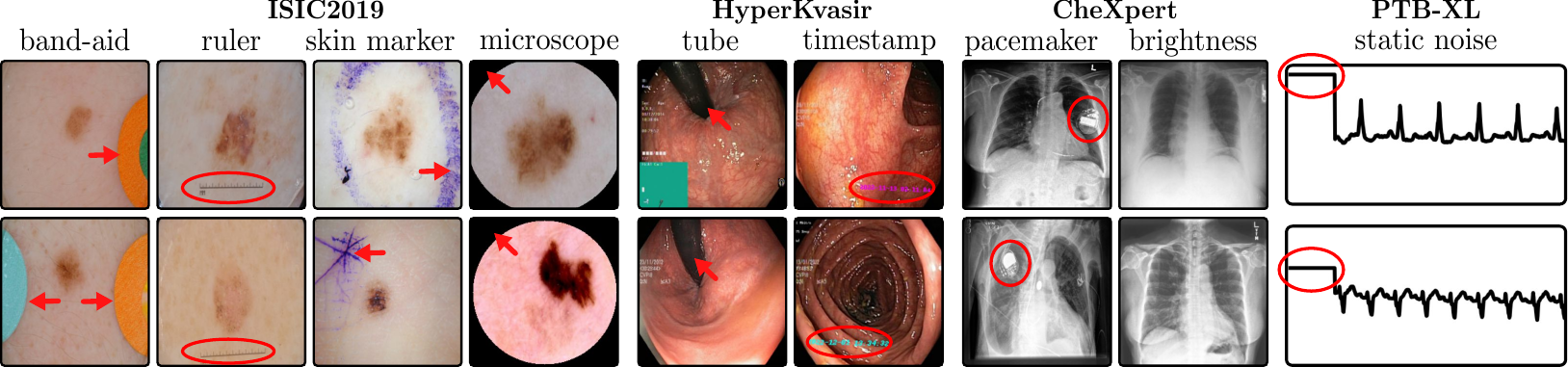}
    \caption{
    Examples for considered artifacts (f.l.t.r.): We use the real-world artifacts \texttt{band-aid}, \texttt{ruler}, and \texttt{skin marker} for ISIC2019, and an artificially inserted \texttt{microscope}-like black circle. In Hyper-Kvasir, we use \texttt{insertion tubes} and artificial \texttt{timestamps}. For CheXpert, we consider \texttt{pacemakers} and artificially increased \texttt{brightness}. Lastly, we insert \texttt{static noise} into one lead in PTB-XL.
    }
    \label{fig:artifacts}
\end{figure*}

As an alternative to input-level bias representations, the \gls{clarc} framework~\citep{anders2022finding} models biases in latent space using \glspl{cav}, requiring only binary labels per sample to indicate the presence of the artifact.
Inspired by \gls{rrr}, \gls{rrclarc}~\citep{dreyer2024hope} penalizes the feature use measured by the \emph{latent} gradient pointing into the direction of the bias, as modeled via the \gls{cav}.
Another research direction is training-free post-hoc model editing \citep{anders2022finding,neuhaus2023spurious,belrose2024leace}.
For instance, \gls{pclarc}~\citep{anders2022finding} removes activations in the bias direction modeled via \glspl{cav} during the prediction.
However, bias mitigation approaches risk ``collateral damage'', meaning that 
whereas the biased concept is successfully suppressed,
valid concepts entangled with the biased concept might be negatively impacted as well.
To address this, \gls{rpclarc} \citep{bareeva2024reactive} only targets samples meeting certain conditions, such as containing the artifact according to the \gls{cav}.

Note, that all presented approaches require either sample- or even feature-level bias annotations, for instance spatial bias localizations, which are costly to obtain.
However, the \mbox{(semi-)automated} bias annotation techniques outlined in Secs.~\ref{sec:data_annotation} and~\ref{sec:localization} reduce the manual data annotation efforts.

\section{Experiments}
\label{sec:experiments}
We evaluate the extended Reveal2Revise framework with four medical datasets from two modalities, namely vision and time-series. We describe the experimental setup~(Sec.~\ref{sec:experimental_setup}) and demonstrate the capabilities of the framework for bias identification~(Sec.~\ref{sec:exp_bias_identifcation}), the detection of biased samples~(Sec.~\ref{sec:exp_data_annotation}), bias localization~(Sec.~\ref{sec:exp_bias_localization}) and mitigation~(Sec.~\ref{sec:exp_bias_mitigation}).

\subsection{Experimental Setup}
\label{sec:experimental_setup}
The considered datasets include ISIC2019 for melanoma detection~\citep{codella2018skin,tschandl2018ham10000,combalia2019bcn20000}, HyperKvasir for the identification of gastrointestinal abnormalities~\citep{borgli2020hyperkvasir},  CheXpert with chest radiographs~\citep{irvin2019chexpert}, and the PTB-XL dataset~\citep{wagner2020ptb} with 12-lead \gls{ecg} (time series) data.
All vision datasets contain real-world artifacts that \glspl{dnn} may utilize as spurious correlation, \ie, features unrelated to the task, yet correlating with the target label. 
ISIC2019 is particularly known for various artifacts like colorful \texttt{band-aids} near benign lesions and \texttt{rulers} or \texttt{skin markers} beside malignant lesions~\citep{rieger2020interpretations,cassidy2022analysis,pahde2023reveal}.
Moreover, HyperKvasir contains \texttt{insertion tubes} predominantly in samples without abnormal conditions, while CheXpert samples with cardiomegaly contain \texttt{pacemakers} in radiographs more frequently than in healthy patients~\citep{weng2025fast}.
In addition, we insert controlled artifacts into a subset of images from exactly one class per dataset.
Specifically, we insert a \texttt{microscope}-like artifact into melanoma samples in ISIC2019.
Moreover, following \citet{dreyer2024hope}, we insert a \texttt{timestamp} into disease-samples from HyperKvasir, mimicking timestamps added by scanning devices.
For CheXpert, we increase the \texttt{brightness} of radiographs with cardiomegaly, while for PTB-XL, we insert a \texttt{static noise} into the first second of one lead for samples with \gls{lvh}.
Inserting these artifacts into $p\%$ of samples from exactly one class creates spurious correlations for that class.
Further dataset details are provided in Appendix~\ref{app:dataset_details} and
examples of the artifacts are shown in Fig.~\ref{fig:artifacts}.

\textbf{Model Details:}
For vision tasks, we use VGG16~\citep{simonyan2014very}, ResNet50~\citep{he2016deep}, and ViT-B-16~\citep{dosovitskiy2020image} model architectures with checkpoints pre-trained on ImageNet~\citep{deng2009imagenet} obtained from the PyTorch model zoo~\citep{paszke2019pytorch} and \texttt{timm}~\citep{rw2019timm}.
For \gls{ecg} data, we utilize a one-dimensional adaptation of  XResNet50~\citep{he2019bag}, following recent benchmarks~\citep{strodthoff2020deep,wagner2024explaining}.
We replace the last linear layer to match the number of classes and finetune the models with training details and model performance reported in Appendix~\ref{app:training_details}. 

\begin{figure*}[t!]
    \centering
    \includegraphics[width=.85\textwidth]{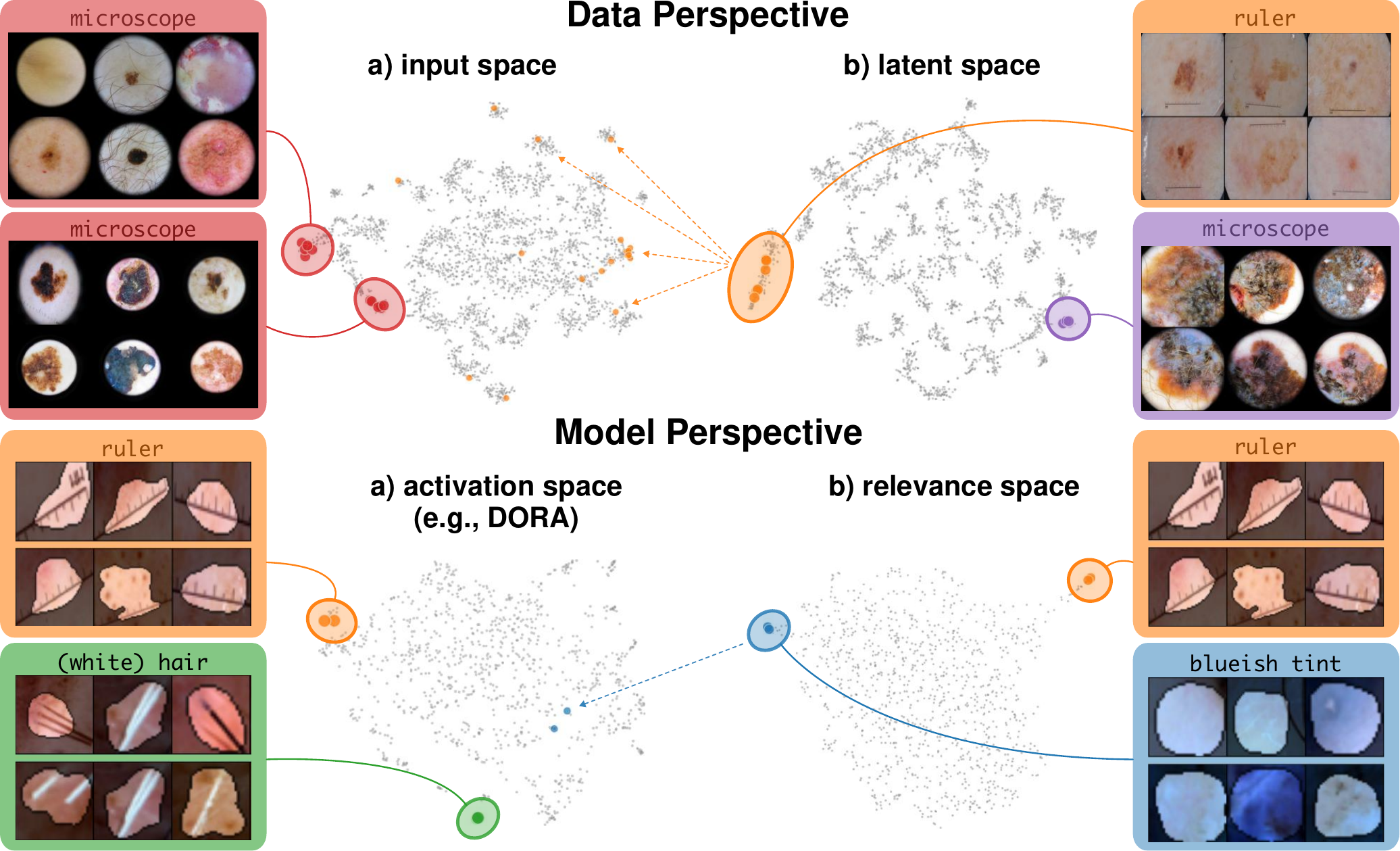}
    
    \caption{
    Detected outlier behavior for the prediction of melanoma 
    using a ResNet50 model trained on ISIC2019. 
    \emph{Top:} Bias identification methods from the data perspective, specifically \gls{spray} on input (\emph{left}) and latent (\emph{right}) relevance scores.
    \emph{Bottom}: Bias identification methods from the model perspective, specifically concept clustering with activation pattern  using \gls{dora} (\emph{left}) and with relevance pattern via cosine similarity (\emph{right}), reveal various bias concepts, \eg, ruler structures or hair.
    }
    \label{fig:reveal_isic_all_methods}
\end{figure*}

\subsection{Concept Validation: Identification of Spurious Behavior}
\label{sec:exp_bias_identifcation}
Given fitted models, we apply bias identification methods described in Sec.~\ref{sec:bias_identification} to detect the spurious model behavior. 
Throughout this section, we focus on the ResNet50 model trained on ISIC2019 with various confounders.
We further limit our analyses to samples from the melanoma class to prevent that clustering model behavior results in clusters representing different classes, and instead allowing us to identify spurious sub-strategies for predicting the considered class.
Results for other classes, model architectures, and datasets, including \gls{ecg} data, are presented in Appendix~\ref{app:sec:exp_bias_identification}.

\paragraph{Data Perspective} 
We first apply \gls{spray} in input and latent space, computing input feature importance scores using \gls{lrp} summed over color channels.  
To obtain latent relevances, we use intermediate relevance scores from the \gls{lrp} computation after the $3^\text{rd}$ (out of four) residual block,  max-pooled over spatial dimensions to yield relevance scores \mbox{$r_l \in \mathbb{R}^{m}$} for layer $l$ with $m$ channels. 
The clustering of pairwise cosine distances between heatmaps is shown in Fig.~\ref{fig:reveal_isic_all_methods} (\emph{top left}).
Detected outlier clusters contain samples with spatially coherent biases, \eg, the black circle around the lesions originating from \texttt{microscopes} ($\mycirc[myred]$). 
In contrast, clustering latent relevance scores reveals more complex, less spatially dependent clusters, as shown in Fig.~\ref{fig:reveal_isic_all_methods} (\emph{top right}), including the \texttt{ruler} artifact~($\mycirc[myorange]$) and the \texttt{microscope}~($\mycirc[mypurple]$). 
Compared to those in input space, the cluster for the \texttt{microscope} in latent space represents a more diverse high-level concept.
In the input space visualization, we further highlight samples from the \texttt{ruler} cluster detected in latent space. 
Instead of forming a cluster, they spread across the entire embedding space, indicating that the bias is too complex to be detected in input space.

\paragraph{Model Perspective} 
Next, we apply bias identification approaches from the model perspective by identifying outlier neurons based on \emph{activation} pattern via \gls{dora} and \emph{relevance} pattern by clustering pair-wise cosine distances between concept relevance scores.
We focus on latent activations and relevances after the third residual block.
\gls{dora} uses a distance function based on how neurons activate upon each others \gls{nams}, achieving high similarity when neurons activate upon similar input signal.
A 2D visualization of the resulting distance matrix is shown in Fig.~\ref{fig:reveal_isic_all_methods} (\emph{bottom left}). 
Identified outlier concepts include \texttt{ruler}~($\mycirc[myorange]$) and \texttt{(white) hair}~($\mycirc[mygreen]$). 
We further compute pairwise cosine distances between spatially aggregated latent relevance scores $\matR \in \Real^{n \times m}$
and apply \gls{umap} to embed the resulting distance matrix $\matDl \in \Real^{m \times m}$ in $\Real^{m \times 2}$. 
This results in high similarity between neurons (concepts) that the model uses similarly for predictions.
The concept clustering is visualized in Fig.~\ref{fig:reveal_isic_all_methods} (\emph{bottom right}), highlighting two outlier clusters focused on \texttt{rulers} ($\mycirc[myorange]$) and \texttt{blueish tint} ($\mycirc[myblue]$).

The additional experiments with \gls{ecg} data (see Appendix~\ref{app:sec:exp_bias_identification})
reveal the artificially inserted \texttt{static noise} in the attacked lead from both data and model perspectives. 
Moreover, additional vision experiments show that 
dominant spurious concepts, such as the artificial \texttt{timestamp} in HyperKvasir or the \texttt{static noise} in PTB-XL, may not be clearly detected as \emph{outlier} concepts.
In such cases, analyzing prediction sub-strategies via \gls{pcx} may provide additional insights on spurious \emph{inlier} behavior.
Hard-to-interpret concept representations pose another challenge for the model perspective. 
For example, the \texttt{brightness} artifact in CheXpert is not clearly visible in the concept \gls{umap} (see. Fig.~\ref{app:fig:reveal:chexpert_attacked}, \emph{right}), but can easily be detected using \gls{spray} (Fig.~\ref{app:fig:reveal:chexpert_attacked}, \emph{left}) or \gls{pcx} (Fig.~\ref{app:fig:reveal:pcx_chexpert}).
In summary, while all considered spurious features are detected, the choice of bias identification approach is crucial, as some shortcuts are easier to detect as outlier concept (\eg, \texttt{ruler}) and others via \gls{pcx} (\eg, \texttt{brightness}, \texttt{static noise} in ECG). 

\subsection{Biased Sample Retrieval}
\label{sec:exp_data_annotation}
\begin{figure*}[t!]
    \centering
    \includegraphics[width=.8\textwidth]{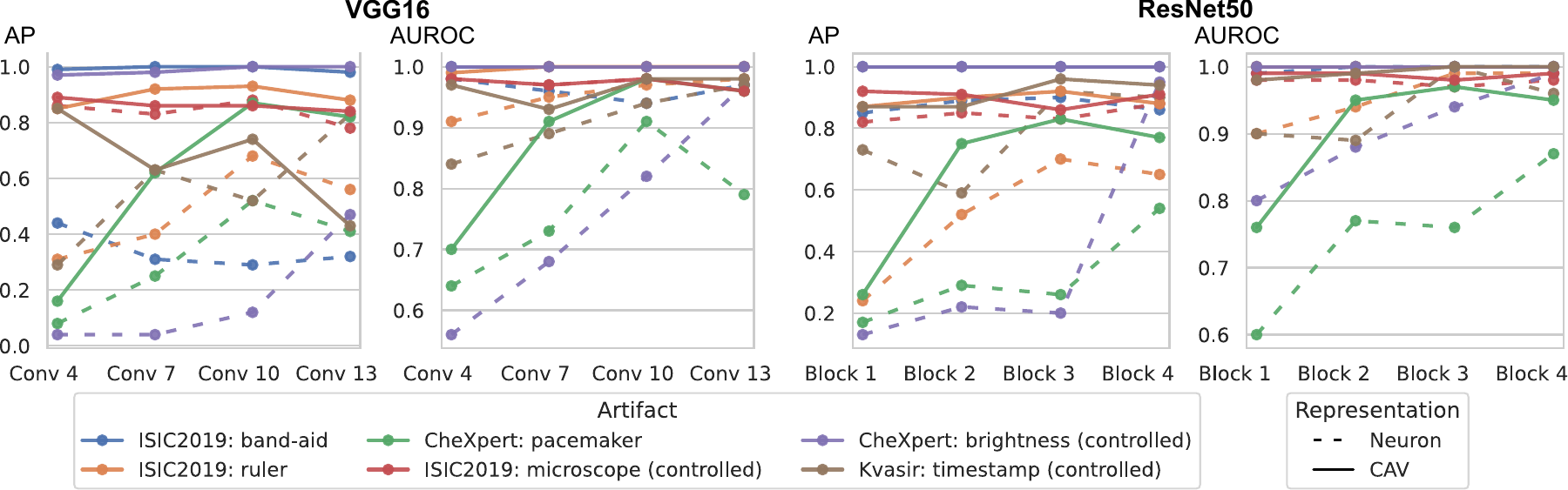}
    
    \caption{
    Quantitative data annotation results measuring the ranking capabilities via \gls{ap} and \gls{auc} for different layers of VGG16 (\emph{left}) and ResNet50 (\emph{right}) using artifacts from ISIC2019 (\text{band-aid}, \texttt{ruler}, \texttt{microscope}), HyperKvasir (\texttt{timestamp}) and CheXpert (\texttt{pacemaker}, \texttt{brightness}). 
    As concept representation, we use single neurons (\emph{dashed}) and \glspl{cav} (\emph{solid}).
    }
    \label{fig:data_annotation_quantitative}
\end{figure*}

In this section, we leverage latent bias representations, either via directions or individual neurons, to retrieve biased samples and measure how well bias samples are separated from clean samples. 
We compute bias scores $\bsact$ as defined in Eq.~\ref{eq:annotation_act} by projecting latent activations onto the bias direction.
As we are mostly interested in the ranking capabilities, \ie, artifact samples should be assigned higher bias scores than clean samples, we measure \gls{auc} and \gls{ap}, considering both real and controlled artifacts.
For real artifacts, we evaluate retrieval using existing artifact labels, while we have ground truth information for controlled experiments.
We train \glspl{cav} using \glspl{svm} on different layers of VGG16 and ResNet50 models for all datasets, reporting \gls{auc} and \gls{ap} on unseen test samples.
Note, that while \citet{pahde2025navigating} claim that classifier-based \gls{cav} computation can yield imprecise concept directions, they argue that SVM-\glspl{cav} are superior for predicting concept presence, the main goal of this experiment.
In Fig.~\ref{fig:data_annotation_quantitative}, we show the results for different layers of VGG16 and ResNet50 using single neurons (dashed line) and \glspl{cav} (solid line) as bias representation.
Best neurons are selected using the validation set.
The results indicate that \glspl{cav} generally outperform single neuron representations for sample retrieval, and the layer choice is crucial depending on the bias type.
For instance, while \glspl{cav} for layers closer to the model heads can detect \texttt{pacemaker} samples, they fail for earlier layers.

\begin{figure*}[t!]
    \centering
    \includegraphics[width=.33\textwidth]{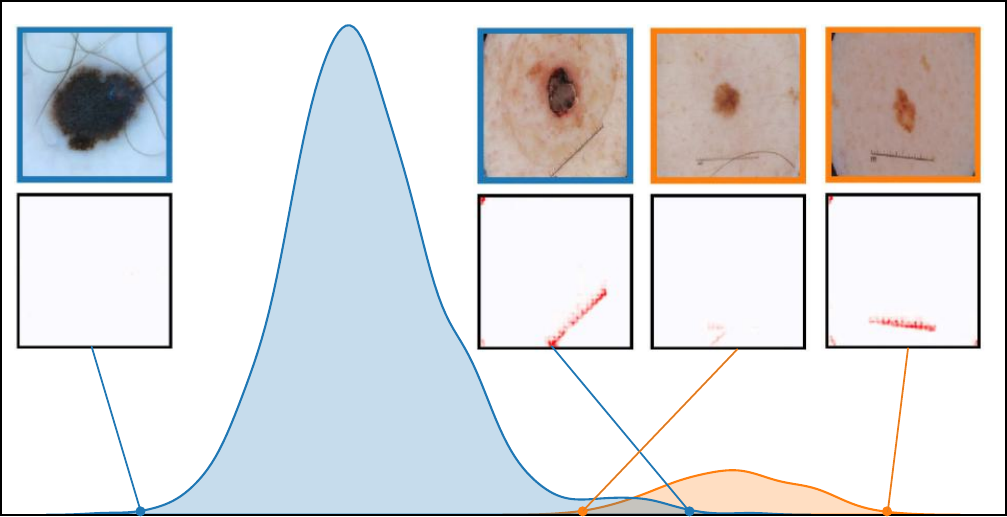}
    \quad\quad\quad\quad\quad
    \includegraphics[width=.33\textwidth]{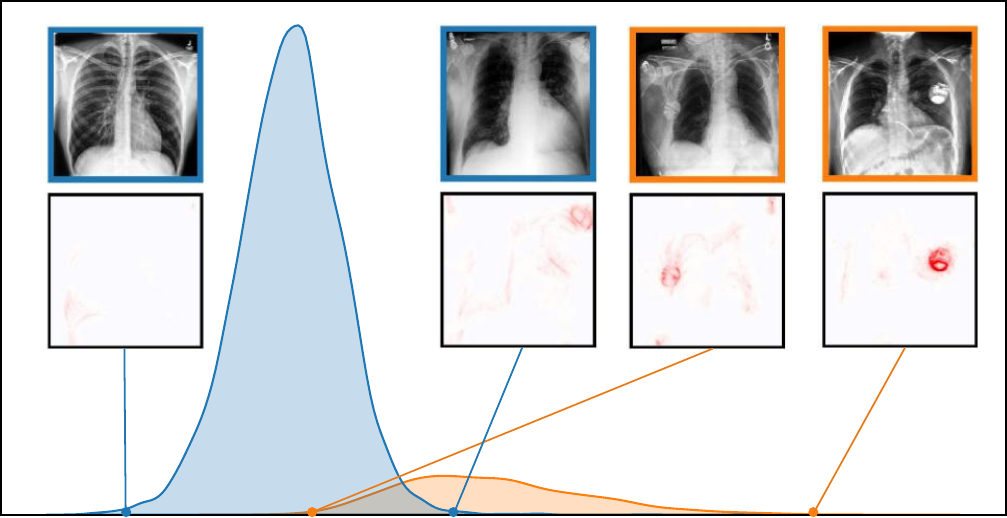}
    \caption{
    Distribution of latent activations projected onto CAV direction for known artifacts \texttt{ruler} ISIC2019 (\emph{left}) and \texttt{pacemaker} in CheXpert (\emph{right}), split into known artifact (\emph{orange}) and other (\emph{blue}) samples.
    We show samples at the 1- and 99-percentile of each set and the artifact localization using CAVs. In both cases, the samples in the 99-percentile of the set \emph{not} labeled as artifact are false negatives, i.e., artifact samples that have not been detected in the data annotation process. 
    }
    \label{fig:data_annotation_real}
\end{figure*}

We further plot the distribution of CAV-based bias scores for biased and clean samples for the real-world artifacts \texttt{ruler} (ISIC2019) and \texttt{pacemaker} (CheXpert) in Fig.~\ref{fig:data_annotation_real}.
We compute bias scores using the best performing \gls{cav} per artifact and use latent activations after the $3^{\text{rd}}$ residual block of the ResNet50 model for \texttt{ruler}, and the $10^{\text{th}}$ convolutional layer of the VGG16 model for \texttt{pacemaker}.
We show samples corresponding to the bias score in the \mbox{top-1~and~-99} percentiles of each set.
Samples near the decision boundary are particularly interesting, as they may arise from labeling errors.
Both examples demonstrate the retrieval of unlabeled artifact samples.
The distributions of bias scores for additional artifacts are shown in Appendix~\ref{app:sec:exp_data_annotation}.

\subsection{Spatial Bias Localization}
\label{sec:exp_bias_localization}
\begin{figure*}[t!]
    \centering
    \includegraphics[width=.75\textwidth]{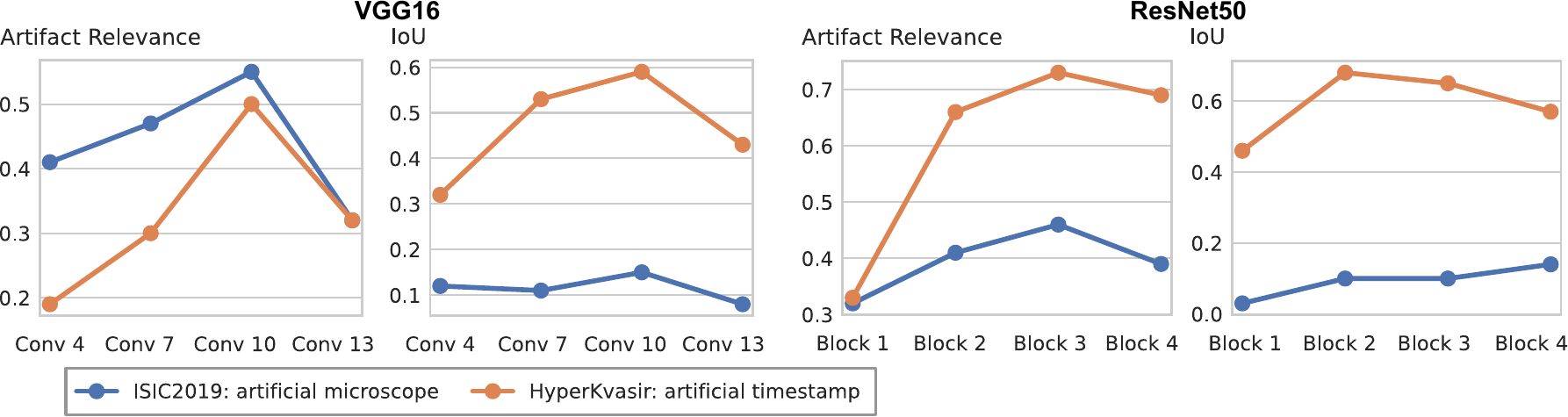}
    
    \caption{
    Bias localization results measuring the artifact relevance and \gls{iou} for various layers of VGG16 and ResNet50 using our controlled artifacts \texttt{microscope} (ISIC2019) and \texttt{timestamp} (HyperKvasir). 
    }
    \label{fig:bias_localization_quantitative}
\end{figure*}

To spatially localize biases in input space with \glspl{cav}, we compute local explanations for the element-wise product of latent activations $\bfeat(\bx)$ and concept direction $\bh_l$ (see Eq.~\ref{eq:localization}). 
We use the controlled artifacts, \ie, \texttt{timestamp} (HyperKvasir) and \texttt{micropscope} (ISIC2019) with ground truth concept localization masks for evaluation. 
We compute (1) the percentage of relevance \emph{within} the ground truth mask and (2) the Jaccard index, or \gls{iou}, of the binarized mask with the ground truth.
In Fig.~\ref{fig:bias_localization_quantitative}, we report both metrics using \glspl{cav} computed on different layers of VGG16 and ResNet50. 
The layer choice for concept representations is key, as for example middle layers perform better to localize \texttt{timestamps} and earlier layers are more effective to localize the \texttt{microscope}.
In comparison with Fig.~\ref{fig:data_annotation_quantitative}, we find that the optimal layer for bias localization may differ from the one for sample retrieval.
Interestingly, the \gls{iou} for the \texttt{microscope} artifact is consistently low, as models primarily focus on the border of the circle instead of the entire area, as indicated by qualitative results in Appendix~\ref{app:sec:exp_bias_localization}.
Unlike artifact relevance, the \gls{iou} metric also measures how much of the expected areas the computed mask does \emph{not} cover.

\subsection{Bias Mitigation}
\label{sec:exp_bias_mitigation}
We unlearn the detected biases using the methods summarized in Sec.~\ref{sec:bias_mitigation}.
We utilize \gls{rrr} as input-gradient-based bias mitigation approach for localizable artifacts and the \gls{clarc} framework for all artifacts, representing biases in latent space with \glspl{cav}. 
For the latter, we consider 
the penalty-based approach \gls{rrclarc} and the training-free model editing methods \gls{pclarc} and \gls{rpclarc}.
To measure the bias mitigation effect, we compute several metrics suggested by the Reveal2Revise framework.
First, we compare the accuracy on a clean (bias-free) test set and a biased test set, where the bias is artificially inserted into samples from all classes. 
Models impacted by spurious correlations are expected to perform poorly on the biased test set.
In addition, we measure the model's sensitivity towards the bias concept by computing (1) the percentage of relevance, measured via \gls{lrp}, on the artifact region using ground truth masks, and (2) the TCAV score~\citep{kim2018interpretability}.
The latter measures how sensitive the model is towards the artifact, represented as \gls{cav}, and is computed as the fraction of predictions that were positively influenced by the concept (see Appendix~\ref{app:sec:exp_bias_mitigation} for details).
It is reported as $\Delta \text{TCAV}=|\text{TCAV}-0.5|$, where 0 indicates no sensitivity and higher values reliance on the artifact.
Low scores are preferred after bias mitigation.
The results are compared to a Vanilla model that is finetuned without a bias mitigation loss term. 
In Tab.~\ref{tab:mitigation}, we report results for ResNet50 models in the controlled settings with ISIC2019, HyperKvasir, and CheXpert.
For \gls{rrr}, we use ground truth bias localization masks and refrain from reporting results for CheXpert, as we consider the \texttt{brightness} artifact unlocalizable in input space.
The results confirm that all models initially rely on the spurious correlation, indicated by a large gap between clean and biased accuracies for Vanilla models.
All bias mitigation approaches improve the accuracy on the biased test set while maintaining high accuracy on the clean test set, demonstrating reduced reliance on the targeted biases.
This is supported by decreased artifact relevance and $\Delta \text{TCAV}$. 
Qualitative results visualizing the decrease in
artifact reliance with input relevance heatmaps and additional quantitative results 
for other model architectures and \gls{ecg} data are provided in Appendix~\ref{app:sec:exp_bias_mitigation}. 

\begin{table*}[t]\centering
    \caption{Bias mitigation results with \gls{rrr}, \gls{rrclarc}, and \gls{pclarc} (plain and reactive) for ResNet50 models for controlled spurious correlations, specifically ISIC2019 (\texttt{microscope}) $|$ HyperKvasir (\texttt{timestamp}) $|$ CheXpert (\texttt{brightness}).  We report accuracy on a clean and biased test set, artifact relevance and $\Delta\text{TCAV}$ and arrows indicate whether high ($\uparrow$) or low ($\downarrow$) are better.}
\begin{tabular}{@{}
l@{\hspace{1em}}
c@{\hspace{1em}}
c@{\hspace{1em}}
c@{\hspace{1.5em}}
c@{\hspace{1.5em}}
}
        \toprule
Method &
Accuracy (clean) $\uparrow$ & Accuracy (biased) $\uparrow$ & Art. relevance  $\downarrow$& $\Delta\text{TCAV}$ 
$\downarrow$\\ 
\midrule
    
\emph{Vanilla} & ${0.87}$ $\,|\,$ ${0.97}$ $\,|\,$ ${0.81}$ & ${0.28}$ $\,|\,$ ${0.62}$ $\,|\,$ ${0.44}$ &  ${0.55}$ $\,|\,$ ${0.51}$ $\,|\,$  -  & ${0.17}$ $\,|\,$ ${0.30}$ $\,|\,$ ${0.33}$ \\
 RRR &  ${0.84}$ $\,|\,$ ${0.97}$ $\,|\,$ \hspace{6px}-\hspace{6px} &  ${0.51}$ $\,|\,$ ${0.82}$ $\,|\,$ \hspace{6px}-\hspace{6px} & ${0.53}$ $\,|\,$ ${0.45}$ $\,|\,$ - &  ${0.19}$ $\,|\,$ ${0.30}$ $\,|\,$ \hspace{6px}-\hspace{6px} \\
       RR-ClArC & ${0.86}$ $\,|\,$ ${0.97}$ $\,|\,$ ${0.82}$ & $\hspace{-1px}\mathbf{0.76}\hspace{-1px}$ $\,|\,$ $\hspace{-1px}\mathbf{0.96}\hspace{-1px}$ $\,|\,$ $\hspace{-1px}\mathbf{0.79}\hspace{-1px}$ &  $\hspace{-1px}\mathbf{0.42}\hspace{-1px}$ $\,|\,$ ${0.34}$ $\,|\,$  -  & $\hspace{-1px}\mathbf{0.00}\hspace{-1px}$ $\,|\,$ $\hspace{-1px}\mathbf{0.07}\hspace{-1px}$ $\,|\,$ $\hspace{-1px}\mathbf{0.00}\hspace{-1px}$ \\
        P-ClArC & ${0.82}$ $\,|\,$ ${0.83}$ $\,|\,$ ${0.72}$ & ${0.59}$ $\,|\,$ ${0.92}$ $\,|\,$ ${0.76}$ &  ${0.44}$ $\,|\,$ $\hspace{-1px}\mathbf{0.18}\hspace{-1px}$ $\,|\,$  -  & ${0.07}$ $\,|\,$ ${0.11}$ $\,|\,$ ${0.30}$ \\
rP-ClArC & ${0.87}$ $\,|\,$ ${0.97}$ $\,|\,$ ${0.81}$ & ${0.60}$ $\,|\,$ ${0.92}$ $\,|\,$ ${0.76}$ &  $\hspace{-1px}\mathbf{0.42}\hspace{-1px}$ $\,|\,$ ${0.19}$ $\,|\,$  -  & ${0.07}$ $\,|\,$ ${0.11}$ $\,|\,$ ${0.30}$ \\

    \bottomrule
\end{tabular}
    \label{tab:mitigation}
    \end{table*}

\section{Limitations}
\label{sec:limitations}
While all steps in the extended Reveal2Revise framework are semi-automated, they require human supervision from domain experts, \eg, to validate outlier concepts, inspect detected bias samples, and determine which concepts should be unlearned. 
Below, we discuss additional challenges associated with each step of the framework.

\begin{itemize}
    \item \emph{Concept Validation/Bias Identification:} When encoded biases dominate, they may not appear as outlier concepts or samples. 
    Identifying prediction sub-strategies, \eg, using \gls{pcx}, can overcome this challenge. 
    Moreover, it is to note that if the discussed approaches are not applied per class label, detected clusters might resemble clusters of classes instead of different sub-strategies.
    \item \emph{Bias Modeling:} Without explicit concept disentanglement (\eg, via \glspl{sae}), concept representations may be non-orthogonal, leading to entangled concepts that negatively impact data annotation and bias mitigation.
    Additionally, the linear separability assumption of \glspl{cav} poses limitations, as there is no guarantee for the existence of a layer perfectly separating concept samples.
    This makes the layer selection an additional hyperparameter, and a suboptimal choice can exacerbate issues like concept entanglement.
    Non-localizable artifacts (\eg, color shifts) are another challenge, as they cannot be effectively modeled using spatial masks in input space.
    Lastly, poorly chosen data samples representing the concept can limit the accuracy of bias representations.
    \item \emph{Bias Localization:} Even with accurate concept representations, explanations may not provide precise localizations.
    This discrepancy may occur when a model's perception of a concept differs from human expectation.
    For instance, in the case of the microscope-artifact, all models captured only the border of the circle rather than the entire area. 
    Consequently, if the model's understanding of a concept does not align with human expectations, the explanation may be ineffective for bias localization.
    
    \item \emph{Bias Mitigation:} With imperfect bias representations, bias mitigation approaches may cause collateral damage, \ie, unlearning of valid concepts entangled with biased concepts. 
    This can be overcome with improved concept disentanglement.
    Moreover, bias mitigation methods, especially post-hoc model editing, might not sufficiently foster valid behavior, but only unlearn invalid strategies. 
    Although successfully unlearning the targeted bias, this can lead to poor model performance on clean data.
\end{itemize}

\section{Conclusions}
\label{sec:conclusions}
In this work, we reviewed the steps of Reveal2Revise, a comprehensive interpretability-based framework for the detection and mitigation of spurious shortcut behavior in \glspl{dnn}, and enhanced the framework with bias annotation capabilities.
Specifically, we utilized concept-based bias representations for the semi-automated computation of sample- and feature-level bias annotations, 
providing valuable insights for the bias mitigation and re-evaluation steps. 
We successfully demonstrated the applicability of the extended Reveal2Revise framework by identifying and mitigating spurious correlations caused by controlled and real-world data artifacts in four medical datasets across two modalities, using VGG16, ResNet50, and \gls{vit} model architectures.
Future work may explore the identification and mitigation of biases in disentangled concept spaces, \eg, leveraging \glspl{sae}. 
Another promising direction is the integration of expected concepts for a more targeted search for unexpected concepts.


\paragraph{Acknowledgements}
We thank Melina Zeeb for assistance with visualizations.
This work was supported by
the Federal Ministry of Education and Research (BMBF) as grant BIFOLD (01IS18025A, 01IS180371I);
the German Research Foundation (DFG) as research unit DeSBi (KI-FOR 5363 $-$ project ID: 459422098);
the European Union’s Horizon Europe research and innovation programme (EU Horizon Europe) as grant TEMA (101093003);
the European Union’s Horizon 2020 research and innovation programme (EU Horizon 2020) as grant iToBoS (965221).

\bibliography{main}

\clearpage
\appendix
\section{Appendix}
In the following, we will provide an algorithmic overview of the extended Reveal2Revise framework in Sec.~\ref{app:algorithm}, additional details and mathematical foundations for \gls{cav}-based bias modeling approaches in Sec.~\ref{app:sec:bias_modeling}, biased sample retrieval in Sec.~\ref{app:biased_sample_retrieval}, and bias mitigation approaches in Sec.~\ref{app:bias_mitigation}.
We will then provide experimental details regarding the datasets in Sec.~\ref{app:dataset_details} and model training in Sec.~\ref{app:training_details}. 
This is followed by additional experimental results in Sec.~\ref{app:sec:experiments}, including \gls{xai}-driven shortcut identification for both ECG and vision data (Sec.~\ref{app:sec:exp_bias_identification}), biased sample retrieval (Sec.~\ref{app:sec:exp_data_annotation}), spatial bias localization (Sec.~\ref{app:sec:exp_bias_localization}), and bias mitigation (Sec.~\ref{app:sec:exp_bias_mitigation}).

\subsection{Reveal2Revise Framework} \label{app:algorithm}
In addition to the visualization in Fig.~\ref{fig:title_figure} (\emph{left}), we provide an algorithmic overview of the Reveal2Revise framework in Algorithm~\ref{app:alg:r2r}.
It includes the steps (a) \emph{reveal} to detect model biases, (b) \emph{bias modeling} to obtain a suitable representation of the bias, (c) \emph{revise} to correct the model behavior with a bias mitigation approach, and (d) \emph{evaluate} to measure if the spurious model behavior was unlearned successfully.

\begin{algorithm}[t]
    \label{app:alg:r2r}
    \vspace{1em}
    \DontPrintSemicolon
    \SetAlgoNlRelativeSize{-1}
    \KwIn{$\bm{f}$: (biased) model, $\bm{X}$: input features, $\mathbf{y}$: labels}
    \KwOut{$\bm{\hat{f}}$: corrected model}
    \;
    $\bm{\hat{f}} \gets \bm{f}$ \;\;

    \Repeat{$\bm{\hat{f}}$ converged}{
        \tcp{\textnormal{(a) Reveal \texttt{bias} either from model or data perspective}}
        $\texttt{bias}$ $\gets$ \textsc{Reveal}($\bm{\hat{f}}$, $\bm{X}$, $\mathbf{y}$) \;\;

        \tcp{\textnormal{(b) Compute \texttt{bias representation} (\gls{cav} and/or input localization)}}
        $\texttt{bias representation}$ $\gets$ \textsc{BiasModeling}($\bm{\hat{f}}$, $\bm{X}$, $\mathbf{y}$, $\texttt{bias}$) \;\;

        \tcp{\textnormal{(c) Revise model given \texttt{bias representation}}}
        $\bm{\hat{f}}$ $\gets$ \textsc{Revise}($\bm{\hat{f}}$, $\bm{X}$, $\mathbf{y}$, $\texttt{bias representation}$) \;\;
        
        \tcp{\textnormal{(d) (Re-)evaluate revised model}}
        $\texttt{metrics}$ $\gets$ \textsc{Evaluate}($\bm{\hat{f}}$, $\bm{X}$, $\mathbf{y}$) \;
    }\;
    
    \caption{Overview of the Reveal2Revise framework.}
\end{algorithm}

In addition, we provide an overivew of our proposed extension to the (a) \emph{reveal} and (b) \emph{bias modeling} steps in Algorithm~\ref{app:alg:r2r_extension}. 
Specifically, the application of bias identification approaches either from the model or data perspective provides an initial set of bias-affected samples, which can be utilized to construct the bias concept labels $\mathbf{t}$~\mbox{(\protect\circlednumblue{1})}.
The bias labels are leveraged to train a \gls{cav} $\bh_l$ as bias representation \mbox{(\protect\circlednumblue{2})}, which is subsequently employed to identify additional biased samples \mbox{(\protect\circlednumblue{3})}.
Specifically, as conceptualized in Fig.~\ref{fig:iterative_data_annotation}, this process involves calculating bias scores $\mathbf{s}_\text{bias}$ (see Eq.~\ref{eq:annotation_act}) for samples that are currently labeled as not containing the bias according to the current labels $\mathbf{t}$. 
The top $n_\text{inspect} \in \mathbb{N}$ high-scoring samples are subject to manual inspection to further refine the bias labels $\mathbf{t}$. 
The process is repeated until convergence~\mbox{(\protect\circlednumblue{4})}.
Finally, the updated \gls{cav} is utilized to spatially localize the bias in affected samples \mbox{(\protect\circlednumblue{5})}.
The algorithm outputs refined bias labels $\mathbf{t}$, \gls{cav} $\bh_l$ as bias representation, and spatial localization masks $\bm{M}$.

\begin{algorithm*}[t]
    \label{app:alg:r2r_extension}
    \vspace{1em}
    \DontPrintSemicolon
    \SetAlgoNlRelativeSize{-1}
    \KwIn{$\bm{f}$: model, $\mathcal{X}$: input data, $\mathbf{y}$: labels, $l$: layer} 
    \KwOut{$\mathbf{t}$: \texttt{bias} labels, $\bh_l$: \gls{cav} (bias representation), $\bm{M}$: bias localization}
    \;
     \tcp{\textnormal{(1) Identification of biased samples (\eg, via \gls{spray})}}
     $\texttt{bias}$ $\gets$ \textsc{Reveal}($\bm{\hat{f}}$, $\mathcal{X}$, $\mathbf{y}$)\;
    $\mathbf{t}_\text{init} \gets$ \textsc{IdentifyInitialSamples}($\bm{f}$, $\mathcal{X}, \mathbf{y}$, \texttt{bias})\;
    $\mathbf{t} \gets \mathbf{t}_\text{init}$\;\;
        
    \Repeat{$\mathbf{t}$ converged}{

        \tcp{\textnormal{(2) (Re-)compute CAV $\bh_l$ on layer $l$ as bias representation}}
        $\bh_l$ $\gets$ \textsc{ComputeCAV}($\bm{f}$, $\mathcal{X}$, $\mathbf{t}$, $l$) \;\;
        
        \tcp{\textnormal{(3) 
        Manual inspection of samples with high bias scores to update $\mathbf{t}$}}
        $\mathcal{X^{-}} \gets \{\bx_i \in \mathcal{X} \mid t_i = 0\}$\;
        $\mathbf{s}_\text{bias}$ $\gets$ \textsc{ComputeBiasScore}($\bm{f}$, $\mathcal{X^{-}}$, $\bh_l$) \;
        \texttt{candidates} $\gets$ \textsc{Top}($\mathcal{X^{-}}$,$\mathbf{s}_\text{bias}$, $n_\text{inspect}$)\;
        $\mathbf{t}$ $\gets$ \textsc{Inspect}(\texttt{candidates})\;\;
        \tcp{\textnormal{(4) Repeat until convergence to improve bias representation $\bh_l$}}
    }\;

    \tcp{\textnormal{(5) Compute spatial localization masks for samples that contain the bias}}
    $\mathcal{X^{+}} \gets \{\bx_i \in \mathcal{X} \mid t_i = 1\}$\;
        $\bm{M}$ $\gets$ \textsc{LocalizeBias}($\bm{\hat{f}}$, $\mathcal{X^{+}}$, $\mathbf{t}$, $\bh_l$) \; \;
    
    \caption{
    Overview of the bias annotation algorithm, 
    which extends the (a) \emph{reveal} and (b) \emph{bias modeling} steps of the Reveal2Revise framework (see Algorithm~\ref{app:alg:r2r}).
    It involves the identification of an initial set of bias-affected samples for a detected bias~\mbox{(\protect\circlednumbluesmall{1})}, the computation of a \gls{cav} as bias representation~\mbox{(\protect\circlednumbluesmall{2})}, the refinement of the bias labels by manually inspecting samples with high bias scores~\mbox{(\protect\circlednumbluesmall{3})}, and the iterative repetition of this process to improve the bias representation~\mbox{(\protect\circlednumbluesmall{4})}.
    Finally, the bias is spatially localized in bias-affected samples~\mbox{(\protect\circlednumbluesmall{5})}.
    The algorithm outputs bias labels $\mathbf{t}$, a bias representation via \gls{cav} $\bh_l$, and localization masks $\bm{M}$.}
\end{algorithm*}


   




\subsection{Bias Modeling}
\label{app:sec:bias_modeling}
In this work, we model biases either via individual neurons or directions in latent space, \ie, \glspl{cav}.
Traditionally, a \gls{cav} is computed by fitting a linear classifier separating samples \emph{with} the concept to be modeled from samples \emph{without} the concept.
The weight vector, \ie, the vector perpendicular to the decision hyperplane, is considered as the concept direction. 
However, \glspl{cav} obtained as weights from linear classifiers can be susceptible to distractor signals in the data and therefore fail in precisely estimating the concept \emph{signal} direction.
To tackle this, Pattern-CAVs have been proposed as an alternative~\citep{pahde2025navigating}.
Whereas \glspl{cav} obtained from \gls{svm} weights are superior in predicting the presence of a concept (\eg, for biased sample retrieval), Pattern-\glspl{cav} are more suitable to precisely model the concept direction (\eg, for \gls{clarc}-based bias mitigation).
In the following, we provide the mathematical foundation for both approaches.

\paragraph{SVM-CAV}
The most common choice for \glspl{cav} are weight vectors from \glspl{svm}.
Specifically,
\glspl{svm}~\citepapp{cortes1995support} 
find a hyperplane maximizing the margin between two classes using the hinge loss $l_h(y,\hat{y})=\text{max}(0,1-y\cdot\hat{y})$
and $L_2$-norm regularization with the following optimization objective:
\begin{align}
\textbf{h}, b &= \argmin_{\bh,b} \\
&\left\{\frac{1}{n} \sum_{\bx, t \in \mathcal{X}} l_h\left(t,\bh^\top \bx+b\right)
+ \lambda \sqrt{\sum_{j \in [m]} \bh_j^2} \right\} ~.
\end{align}
\paragraph{Pattern-CAV}
Pattern-\glspl{cav}~\citep{pahde2025navigating} can be computed based on the covariance between the latent activations $\ba(\bx)$ and the concept labels $t$ as:
\begin{equation}
     \hpat_l = 
    \frac{1}{\sigma_t^2|\mathcal{X}|} \sum_{\bx, t \in \mathcal{X}} (\bfeat
    (\bx) - \bar{\mathcal{A}}_l)(t - \bar t)
    \label{eq:pcav}
\end{equation}
with mean latent activation 
$\bar{\ba_l} \in \Real^m$,
mean concept label $\bar{t}\in\Real$ and sample concept label variance $\sigma_t^2$, which is equal to the sample covariance between the latent activations $\bfeat(\bx)$ and the concept labels $t$ divided by the sample concept label variance.

\subsection{Biased Sample Retrieval}
\label{app:biased_sample_retrieval}
Inspired by the idea behind \gls{rmax}, the identification of biased samples can also be implemented using relevance scores, as an alternative to the activation-based biased sample retrieval described in Eq.~\ref{eq:annotation_act} in the main paper.
Specifically, we can compute a bias score $\bsrel \in \Real$ by projecting the \emph{relevance} for the prediction of of a sample $\x$ onto the concept direction $\bhr^{c}_l$, computed similar as \glspl{cav}, but using latent relevances scores, instead of activations, on layer $l$ \wrt class $c$:

\begin{equation}
    \bsrel=\bhr_l^{c~\top}\fnrel(\x)
\end{equation}
with latent relevances $\fnrel(\x)$ for sample $\bx \in \mathcal{X}$ \wrt class $c$ computed on layer $l$ by a local attribution method, \eg, 
using \gls{lrp}.
As the relevance is computed in a class-specific manner, this approach further allows distinguishing concepts that are artifactual for certain classes but valid features for others. 

\subsection{Bias Mitigation}
\label{app:bias_mitigation}
\subsubsection{Regularization-based Bias Mitigation}
Recent approaches define a loss function designed to encourage or enforce pre-defined behavior. 
Therefore, additional prior information $\bA$ encoding \emph{expected} behavior is required.
This leads to an overall loss $\mathcal{L}_\text{total}$ based on a $\lambda$-weighted sum of the classification loss $\mathcal{L}_\text{class}$ and the newly constructed ``right-reason'' loss term $\lossrr$:
\begin{equation}
\label{eq:loss_rr}
    \mathcal{L}_\text{total}(\bx,y,\bA)=\mathcal{L}_\text{class}(\bx,y) + \lambda \lossrr(\bx,\bA)~\text{.}
\end{equation}

\paragraph{Input (heatmap-based)}
Bias mitigation approaches on input level require sample-wise pixel-level annotations, indicating the presence of a data artifact, available as prior knowledge.
This prior information can be used to align expected behavior with the model's prediction stategy.
Specifically, \gls{rrr}~\citep{ross2017right} introduces a loss term penalizing the alignment between the input gradient, \ie, the gradient of the prediction \wrt the input, and (binary) input masks localizing the data artifact. 
Hence, the model is penalized for paying attention on undesired regions.
Given an input sample $\bx$ along with bias localization mask $\bmask$, the \gls{rrr} loss term $\mathcal{L}_\text{RRR}$ is defined as follows:
\begin{equation}
    \mathcal{L}_\text{RRR}
    (\bx,\bmask)= (\boldsymbol{\nabla}_\x \log \left(\bfunc(\x)\right) \circ~\bmask)^2~\text{.}
\end{equation}

Alternatively, \citet{rieger2020interpretations} propose \gls{cdep}, using Contextual Decomposition importance scores~\citepapp{murdoch2018beyond} 
instead of the gradient to encode model behavior to be aligned with prior knowledge.

Note, however, that pixel-wise annotations for expected model behavior are expensive to obtain, leading to practical limitations. 
Concept localization approaches, as described in Sec.~\ref{sec:localization}, can be utilized to semi-automate this process.
Another limitation of input-level bias mitigation approaches is their inability to address spatially distributed and interconnected biases. Spurious correlations that are spread over the entire image, such as color shifts or overlapping other important concepts, cannot be effectively mitigated at the input level.

\paragraph{Latent space (CAV-based)} To overcome aforementioned limitations, recent work introduces the \gls{clarc} framework~\citep{anders2022finding} that leverages latent concept representations using \glspl{cav} for bias mitigation which has two advantages. 
First, using the latent space alleviates the reliance on spatial locations and therefore allows the correction of unlocalized biases, such as color shifts. 
Moreover, the annotation costs are substantially lower, since only sample-level concept labels are required for the computation of \glspl{cav}, instead of pixel-level artifact masks.
Given the bias representation in the form of a \gls{cav}, the \gls{clarc} framework unlearns related model behavior by either adding or removing activations in the artifact direction. 
Built upon this idea, \citet{dreyer2024hope} introduce \gls{rrclarc}, utilizing a loss function inspired by \gls{rrr} which penalizes the feature use measured through the \emph{latent} gradient pointing into the direction of the bias, as modeled via the \gls{cav}.
More precisely, given a bias direction $\bh_l$, the latent right reason loss $\mathcal{L}_\text{RR-ClArC}$ is defined as follows:

\begin{equation}
    \mathcal{L}_\text{RR-ClArC}
    (\bx,\bh_l)= \left( \boldsymbol\nabla_{\bm{a}} \bfunct(\bfeat(\bx) \cdot \bh_l \right)^2~\text{.}
\end{equation}
Intuitively, this loss term penalizes the model if it changes the prediction behavior when activations in the bias direction are added/subtracted, corresponding to the addition/removal of the modeled artifact.
Note, that the latent gradient is computed \wrt a chosen class, allowing for class-specific bias mitigation.

\subsubsection{Post-hoc Model Editing}
To further reduce the computational requirements for bias mitigation, another line of works removes undesired behavior by post-hoc model editing in training-free manner. 
This group of bias mitigation approaches commonly models the undesired concept as a linear direction in latent space, followed by a modification of latent representations
or model parameters to make model predictions invariant towards the modeled direction.
As such, \gls{pclarc}~\citep{anders2022finding} utilizes \glspl{cav} to model the concept to be erased and projects out activations into the concept direction during inference time.
Specifically, given \gls{cav} $\bh_l$ and latent activations $\bfeat(\bx)$ for layer $l$, the activations are overwritten to $\bfeat'(\bx)$ as
\begin{equation}
    \bfeat'(\bx) = \bfeat(\bx) - \lambda(\bx)\bh_l~\text{,}
\end{equation}
with perturbation strength $\lambda(\bx)$ based on the input sample $\bx$, for instance chosen such that activations in \gls{cav} direction are equal to the average value of clean (non-artifactual) samples. 
Intuitively, subtracting activations along the concept direction in latent space is equivalent to removing the concept in input space.
The modification can be performed either for all samples~\citep{anders2022finding} or in a reactive manner~\citep{bareeva2024reactive}, conditioned for example on the predicted class label or the existence of the spurious feature.

Similarly, SpuFix~\citep{neuhaus2023spurious} models concepts as PCA components in the activation space in layer $l$ and overwrites activations for spurious directions by $\min(\bm{\alpha}^c_k,0)$, with $\bm{\alpha}^c_k$ being the activations for the spurious PCA component $k$ for class $c$, thereby not allowing positive contributions in the spurious direction.
\citetapp{ravfogel2020null} 
perform an iterative null-space projection of linear directions encoding the undesired concept to make sure the concept cannot be recovered from other directions. 
\citetapp{santurkar2021editing} 
suggest classifier editing, a procedure modifying the model weights directly to shift the model's prediction behavior. 
Most recently, \gls{leace}~\citep{belrose2024leace} is a closed-form solution for concept erasure, minimally changing the model's internal representations on \emph{all} layers.
It is to note, however, that post-hoc editing approaches can lead to ``collateral damage'', meaning that whereas the concept is successfully suppressed for biased samples, the model modification can have a negative impact on clean samples, as further discussed by \citet{bareeva2024reactive}.

\subsection{Dataset Details}
\label{app:dataset_details}
Our experiments include the following datasets:
ISIC2019 for image-based melanoma detection, HyperKvasir for the identification of abnormal conditions in the gastrointestinal track, CheXpert with chest radiographs, and PTB-XL for the prediction of cardiovascular diseases from \gls{ecg} data.
ISIC2019 consists of 25,331 samples classified into eight lesion types, both malignant and benign.
The HyperKvasir dataset contains 10,662 samples labeled into 23 classes of findings.
We aggregate the findings into two classes representing findings with and without disease to consider the task as a binary classification task. 
The CheXpert dataset is a collection of 224,316 chest radiographs of 65,240 patients with labels for 14 conditions, including cardiomegaly.
We use a subset of 28,878 samples with additional annotations for the presence of pacemakers provided by \citet{weng2025fast}. 
Lastly, in addition to vision data, we use the PTB-XL dataset containing 21,837 records of 10 second 12-lead \gls{ecg} data (time series data) and labels for 23 cardiovascular conditions. 
We report details for all considered datasets in Tab.~\ref{app:tab:dataset_details}.
Specifically, we report the number of classes and samples, train/validation/test-splits and considered data artifacts.

We use ISIC2019 for multi-class classification and consider the real-world artifacts \texttt{ruler}, \texttt{skin-marker}, and \texttt{reflection}, as well as the artificial \texttt{brightness} artifact inserted into $20\%$ of samples of class ``melanoma''. 
For \texttt{ruler} (all classes) and \texttt{band-aid} (only occurring in ``melanocytic nevus''), we use labels indicating the existence of the artifact provided by \citet{anders2022finding}.
For HyperKvasir, we split the classes into ``normal''/``no-disease'' and ``abnormal''/``disease`` to implement the task as a binary classification task, as described in Tab.~\ref{app:tab:hyperkvasir_details}. 
We insert a \texttt{timestamp} artifact into $10\%$ of the the training samples with label ``disease'', and, in addition, consider the real-world artifact \texttt{insertion tube}. The latter was primarily detected in samples from class ``no disease''.
For CheXpert, we solve a binary classification task to predict the presence of the class ``cardiomegaly''.
We consider the \texttt{pacemaker} artifact dominantly occurring in class ``cardiomegaly'', with artifact labels provided by \citet{weng2025fast}.
In addition, we artificially increase the \texttt{brightness} in $10\%$ of ``cardiomegaly'' samples as a controlled artifact.
Lastly, we use the PTB-XL dataset with ECG-data for multi-class classification into cardiovascular conditions. 
As a controlled artifact, we insert a \texttt{static noise} (constant high value) into the first second of one lead (\texttt{II}-lead) for $50\%$ of samples of class ``left ventricular hypertrophy''.

\begin{table*}[t]\centering
    \caption{Details for considered datasets, including ISIC2019, HyperKvasir, CheXpert, and PTB-XL. We report the number of classes and samples, the train/val/test-splits, and the considered data artifacts. For the latter, we further report whether the artifact is controlled (or real-world) and, if applicable, in which class it occurs. For controlled artifacts, we further report the percentage of samples of the attacked class in which the artifact is inserted.}
\begin{tabular}{@{}
l@{\hspace{.5em}}c@{\hspace{.5em}}
c@{\hspace{.5em}}c@{\hspace{.5em}}
c@{\hspace{.5em}}c@{\hspace{0.5em}}
c@{\hspace{.5em}}
}
        \toprule
name & classes & samples & train/val/test & artifact & controlled & artifact class ($\%$)\\ 

\midrule
ISIC2019 & 8  & 25,331 & $0.8/0.1/0.1$ & \texttt{microscope} & yes & melanoma ($20\%$)\\
& & & & \texttt{ruler} & no & -\\
& & & & \texttt{band-aid} & no & \emph{melanocytic nevus}\\
& & & & \texttt{skin marker} & no & -\\
& & & & \texttt{reflection} & no & \emph{mainly benign keratosis}\\
\midrule
HyperKvasir & 2 & 10,662 & $0.8/0.1/0.1$ & \texttt{timestamp} & yes & disease ($10\%$) \\
  &  &  &  & \texttt{insertion tube} & no & \emph{mainly no disease} \\
\midrule
CheXpert & 2 & 28,878 & $0.8/0.1/0.1$ & \texttt{brightness} & yes & cardiomegaly ($10\%$) \\
         &  &  &  & \texttt{pacemaker} & no &  \emph{mainly cardiomegaly}\\

\midrule
PTB-XL & 23 & 21,837 & $0.8/0.1/0.1$ & \shortstack[c]{\texttt{static noise} \\ (\texttt{II}-lead)} & yes & \shortstack[c]{left ventricular \\hypertrophy ($50\%$)} \\
    \bottomrule
\end{tabular}
    \label{app:tab:dataset_details}
    \end{table*}

\begin{table*}[t]\centering
    \caption{Categorization of classes in HyperKvasir into ``disease'' and ``no-disease''.}
    \begin{tabular}{@{}
l@{\hspace{2em}}l@{\hspace{1em}}
}
        \toprule
class & sub-classes\\ 

\midrule
disease & 
 \shortstack[l]{
``\textit{barretts}'',
``\textit{short-segment-barretts}'',
``\textit{oesophagitis-a}'',
``\textit{oesophagitis-b-d}'',\\
``\textit{hemorrhoids}'',
``\textit{hemorroids}'',
``\textit{polyp}'',
``\textit{ulcerative-colitis-grade-0-1}'',\\
``\textit{ulcerative-colitis-grade-1-2}'',
``\textit{ulcerative-colitis-grade-2-3}'',\\
``\textit{ulcerative-colitis-grade-1}'',
``\textit{ulcerative-colitis-grade-2}'',
``\textit{ulcerative-colitis-grade-3}'',\\
``\textit{dyed-lifted-polyps}'',
``\textit{impacted-stoo}l''
}
    \\ \midrule
no-disease & 
\shortstack[l]{
``\textit{bbps-0-1}'',
``\textit{bbps-2-3}'',
``\textit{dyed-resection-margins}'',
``\textit{ileum}'',
``\textit{retroflex-rectum}'', \\
``\textit{retroflex-stomach}'',
``\textit{normal-cecum}'',
``\textit{normal-pylorus}'',
``\textit{normal-z-lin}e''
}
\\
    \bottomrule
\end{tabular}
    \label{app:tab:hyperkvasir_details}
    \end{table*}

\subsection{Training Details}
\label{app:training_details}
We provide training details for all considered models in Tab.~\ref{app:tab:training_details}.
This includes VGG16, ResNet50 and \gls{vit} model architecture for vision datasets, \ie, ISIC2019, HyperKvasir, and CheXpert, each both with and without controlled artifact. 
We use model checkpoints from the PyTorch model zoo (\texttt{torchvision})~\citep{paszke2019pytorch} for VGG16 (weights: \texttt{VGG16\_Weights.IMAGENET1K\_V1}) and ViT-B-16 (weights: \texttt{ViT\_B\_16\_Weights.IMAGENET1K\_V1}) models, as well as from \texttt{timm}~\citep{rw2019timm}, specifically the checkpoint named \texttt{resnet50d.a3\_in1k} for ResNet50.
Moreover, we train a XResNet1d50 model for the (controlled) PTB-XL dataset.
We report accuracy and false positive rate (FPR) for the affected class on a clean and, for controlled settings, an attacked test set, with artifacts inserted into \emph{all} samples. 
We train models with an initial learning rate $\alpha \in \{0.005,0.001,0.005\}$, either with SGD or Adam optimizers. 
We divide the learning rate by $10$ after $150$ and $250$ epochs, respectively, for vision datasets ($50$/$75$ for PTB-XL), and select $\alpha$ and the optimizer for the final model based on the performance on the validation set.
Note, that a large gap between accuracy and FPR on the \emph{clean} compared to the \emph{attacked} test set indicates that the model is sensitive towards the artifact, \ie, it picks up the spurious correlation.

\begin{table*}[t]
    \centering
    \caption{Model performance and training details for all considered datasets and architectures. 
    We train VGG16, ResNet50, and \gls{vit} models for vision datasets, \ie, ISIC2019, HyperKvasir, and CheXpert, in the original and controlled version, respectively. 
    Moreover, we train an XResNet1d50 model for \gls{ecg} data in PTB-XL.
    We report the accuracy and false positive rate (FPR) on a clean and, for controlled datasets, an attacked test set. 
    The FPR is computed \wrt the attacked class in the controlled settings, and for classes ``melanoma'' (ISIC2019),  ``Disease'' (HyperKvasir) and  ``cardiomegaly'' (CheXpert) for the original datasets.
    In addition, we report training details including optimizer, learning rate (LR) and number of training epochs. The former two are selected based on the performance on the validation set.}
\begin{tabular}{@{}
l@{\hspace{1em}}c@{\hspace{1em}}|
c@{\hspace{.5em}}c@{\hspace{.5em}}|
c@{\hspace{.5em}}c@{\hspace{.5em}}|
c@{\hspace{1em}}c@{\hspace{1em}}c@{\hspace{1em}}
}
        \toprule
        &        &\multicolumn{2}{c}{Accuracy $\uparrow$} & \multicolumn{2}{c}{FPR $\downarrow$} & \multicolumn{3}{c}{Training details}\\
Dataset & Model  & clean & attacked & clean & attacked & Optim. & LR & epochs\\ 

\midrule
\multirow{3}{*}{ISIC2019} 
& VGG16 & 0.82& -& 0.09& -& SGD& 0.001 & 300\\
& ResNet50 & 0.88& -& 0.03 & - & Adam & 0.0005& 300\\
& ViT & 0.79& -& 0.05& -& SGD& 0.005& 300\\
\midrule
\multirow{3}{*}{\shortstack[c]{ISIC2019\\(attacked)}} 
& VGG16 & 0.85& 0.25& 0.04& 0.88& SGD& 0.001 & 300\\
& ResNet50 & 0.87& 0.28& 0.02 & 0.85 & Adam & 0.001& 300\\
& ViT & 0.81& 0.27& 0.04& 0.85& SGD& 0.005& 300\\
\midrule
\multirow{3}{*}{\shortstack[c]{Hyper-\\Kvasir}} & VGG16 & 0.96& -& 0.02& -& SGD& 0.001 & 300\\
                    & ResNet50 & 0.97& -& 0.01& -& Adam& 0.0005& 300 \\
                    & ViT & 0.95& -& 0.02& -& SGD& 0.005 & 300\\
\midrule
\multirow{3}{*}{\shortstack[c]{Hyper-\\Kvasir\\(attacked)}} & VGG16 & 0.97& 0.68& 0.02& 0.39& SGD& 0.005 & 300\\
                    & ResNet50 & 0.97& 0.60& 0.02& 0.48& Adam& 0.001& 300 \\
                    & ViT & 0.93& 0.42& 0.01& 0.70& SGD& 0.001 & 300\\
\midrule
\multirow{3}{*}{CheXpert} & VGG16 & 0.83& -& 0.08& -& SGD& 0.001& 300\\
& ResNet50 & 0.82& -& 0.08& -& Adam& 0.001 &300\\
& ViT & 0.80& -& 0.09& -& SGD& 0.005& 300 \\
\midrule
\multirow{3}{*}{\shortstack[c]{CheXpert\\(attacked)}} & VGG16 & 0.83& 0.52& 0.07& 0.58& SGD& 0.005& 300\\
& ResNet50 & 0.81& 0.42& 0.09& 0.73& Adam& 0.005 &300\\
& ViT & 0.80& 0.25& 0.12& 0.97& SGD& 0.0005& 300 \\
\midrule
{\shortstack[c]{PTB-XL\\(attacked)}} & XResNet1d50 & 0.96& 0.94& 0.00& 0.43& Adam& 0.001& 100\\
    \bottomrule
\end{tabular}
    \label{app:tab:training_details}
    \end{table*}

\subsection{Additional Results}
\label{app:sec:experiments}
This section will provide additional experimental results. 
Specifically, we report results for the \gls{xai}-driven shortcut identification for both ECG and vision data in Sec.~\ref{app:sec:exp_bias_identification}, the biased sample retrieval in Sec.~\ref{app:sec:exp_data_annotation}, the spatial bias localization in Sec.~\ref{app:sec:exp_bias_localization}, and, lastly, bias mitigation in Sec.~\ref{app:sec:exp_bias_mitigation}.

\subsubsection{Detection of Spurious Model Behavior}
\label{app:sec:exp_bias_identification}
In the following, we apply the considered spurious behavior detection approaches, from both data and model perspective, to all datasets, including \gls{ecg} and vision data.

\begin{figure*}[t!]
    \centering
    \includegraphics[width=.6\textwidth]{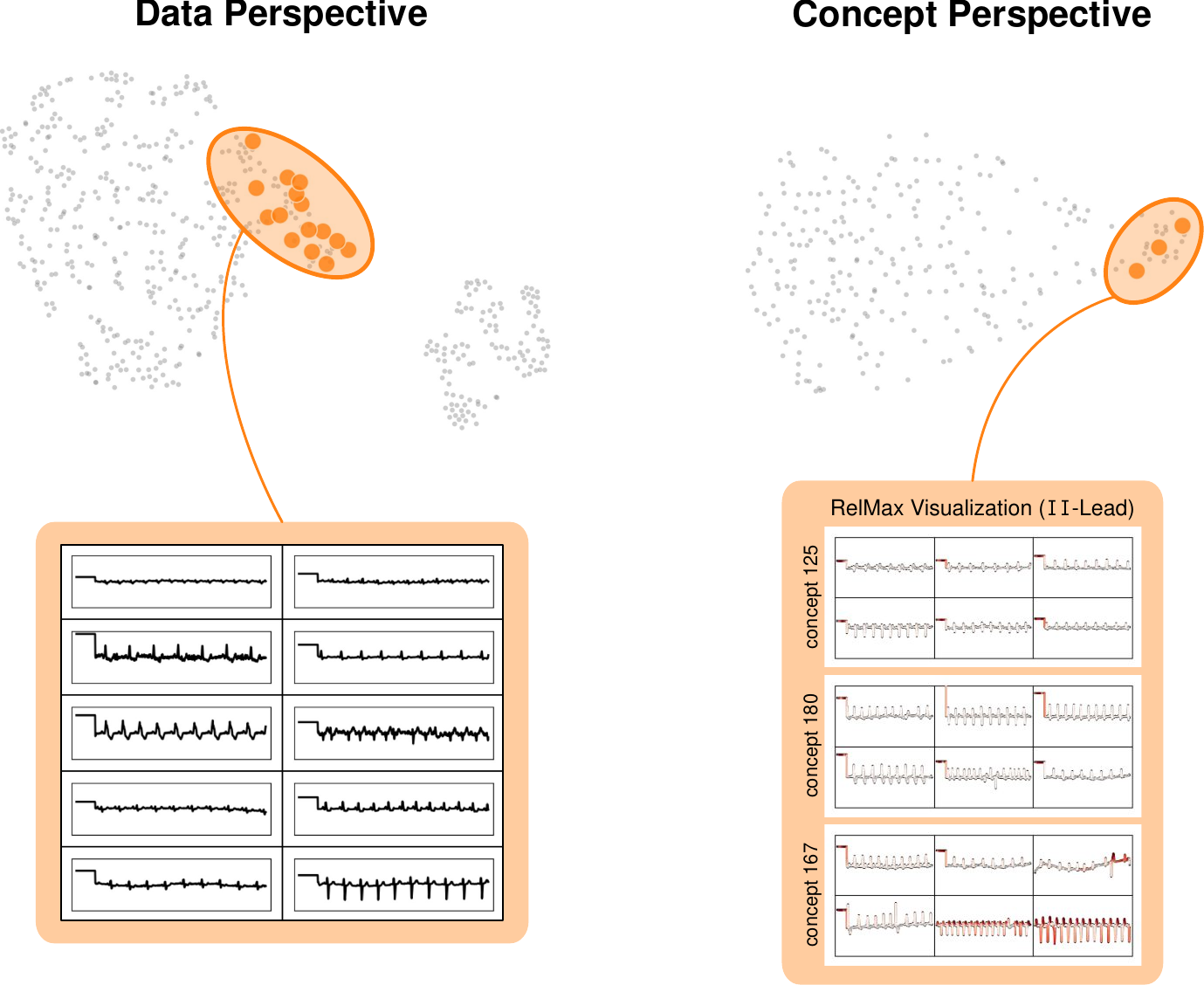}
    
    \caption{
    Artifact identification on the last conv layer of a XResNet50 model trained on ECG data from the PTB-XL dataset with samples from the ``\gls{lvh}''-class using \gls{spray} on latent relevances for the data perspective (\emph{left}) and pair-wise cosine similarities for the model perspective (\emph{right}). 
    For the data perspective, we identify a coherence set of samples containing the inserted \texttt{static noise} artifact in the poisoned lead. 
    However, as the controlled artifact is very dominant, the prediction behavior manifests as \emph{inlier}, not as \emph{outlier}.
    From the model perspective, we identify a set of outlier concepts focusing on the artifact in the \texttt{II}-lead.
    }
    \label{app:fig:reveal_ptb}
\end{figure*}

\begin{figure*}[t!]
    \centering
    \includegraphics[width=.4\textwidth]{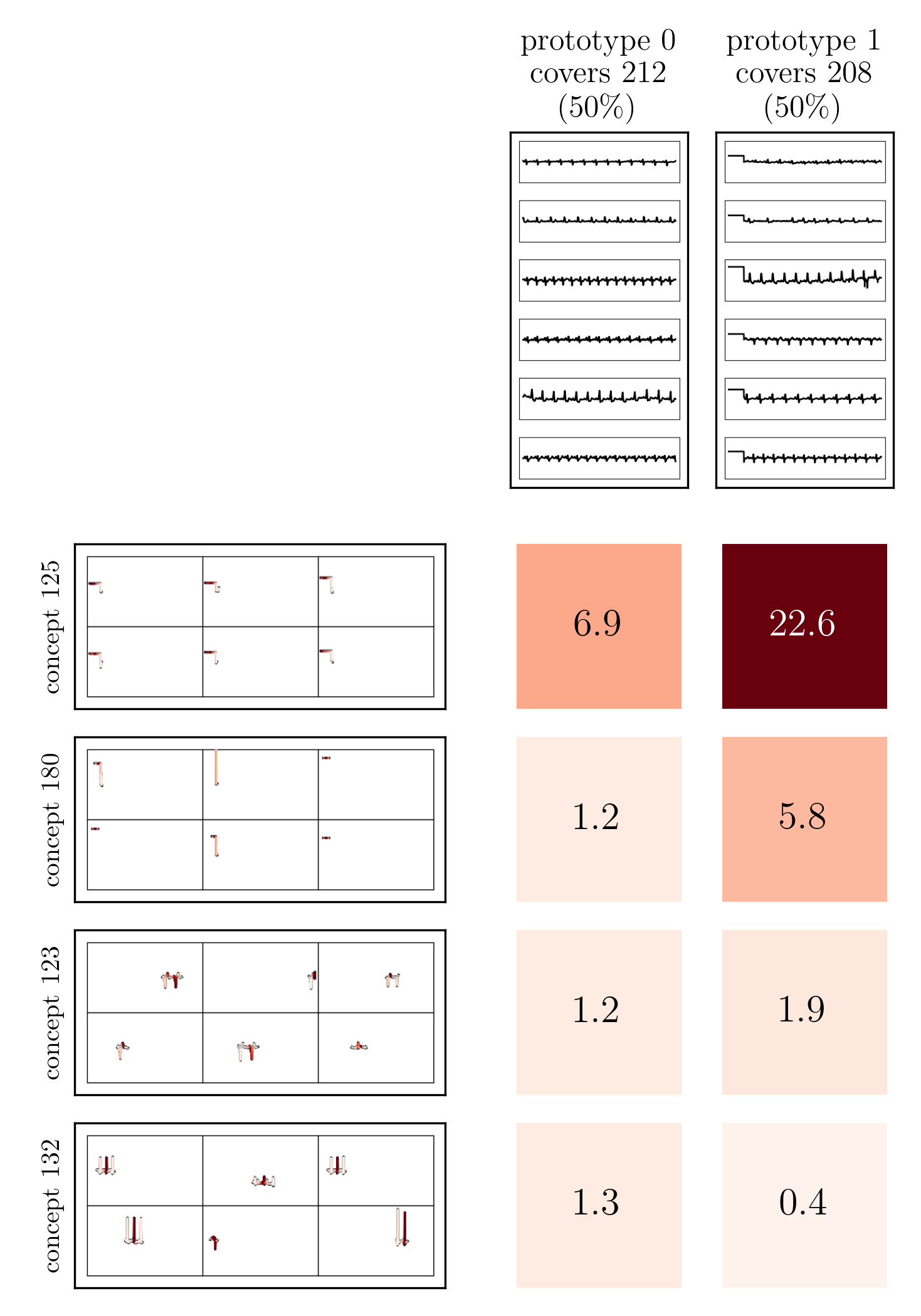}
    
    \caption{
    \gls{pcx} visualizations using latent relevances after the last residual block of the XResNet50 model trained on PTB-XL with the controlled \texttt{static noise} artifact. Only samples from the attacked class ``\gls{lvh}'' are considered.
    Columns represent prototypes with the signal coming from the attacked lead of six representative samples, rows represent concepts (visualized via RelMax) and the values in the matrix indicate the average relevance of the concept for the prototype. 
    \texttt{Prototype 1} clearly focuses on the inserted \texttt{static noise} artifact, with high scores for the related concepts (neurons) \texttt{$\#$125} and \texttt{$\#$180}.
    }
    \label{app:fig:reveal_ptb_pcx}
\end{figure*}

\paragraph{ECG Data (XResNet1d50)}
First, we apply \gls{spray} (data perspective) and pairwise cosine similarities between max-pooled relevance scores (model perspective) to the \gls{ecg} data in PTB-XL with the controlled \texttt{static noise} artifact in Fig.~\ref{app:fig:reveal_ptb}.
Whereas the application of \gls{spray} reveals a coherent cluster of poisoned samples, they do not appear as \emph{outlier}, but instead as \emph{inlier} behavior.
This indicates that the prediction behavior is very dominant, potentially due to the high poisoning rate of $50\%$. 
However, the model perspective reveals an outlier cluster of concepts clearly focusing on the inserted artifact (see neurons \texttt{$\#$125} and \texttt{$\#$180}).
Moreover, we apply \gls{pcx}, which further allows the analysis of inlier behavior by revealing prediction sub-strategies for considered classes, as shown in Fig.~\ref{app:fig:reveal_ptb_pcx}.
We can clearly identify \texttt{prototype 1} as sub-strategy using the inserted \texttt{static noise} artifact, with high relevance scores for the related concepts detected by neuron \texttt{$\#$125} and \texttt{$\#$180}.
Note, that the prototype covers $50\%$ of the test data, which is exactly the inserted poisoning rate.

\begin{figure*}[t!]
    \centering
    \includegraphics[width=.35\textwidth]{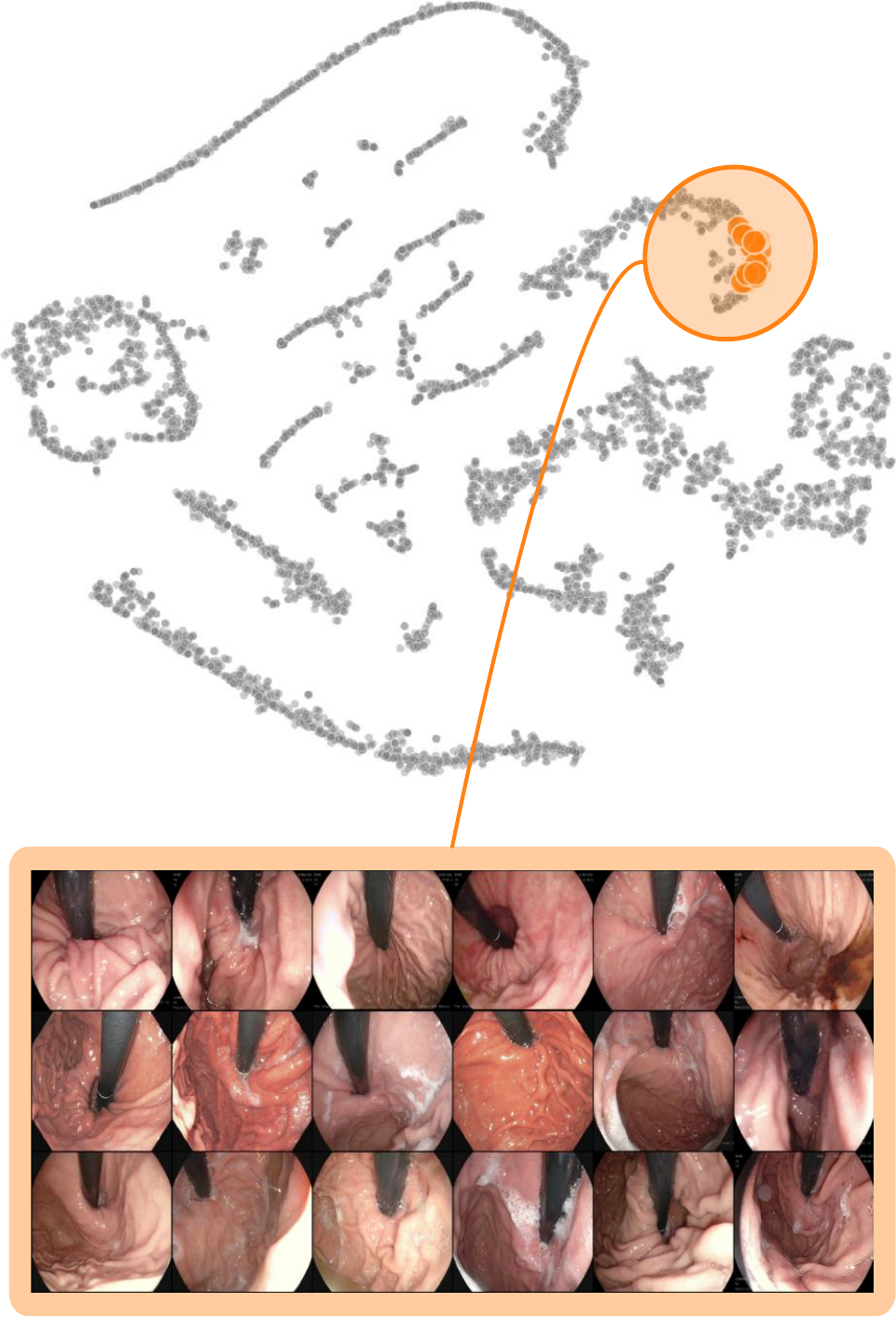}
    \quad\quad\quad\quad
    \includegraphics[width=.35\textwidth]{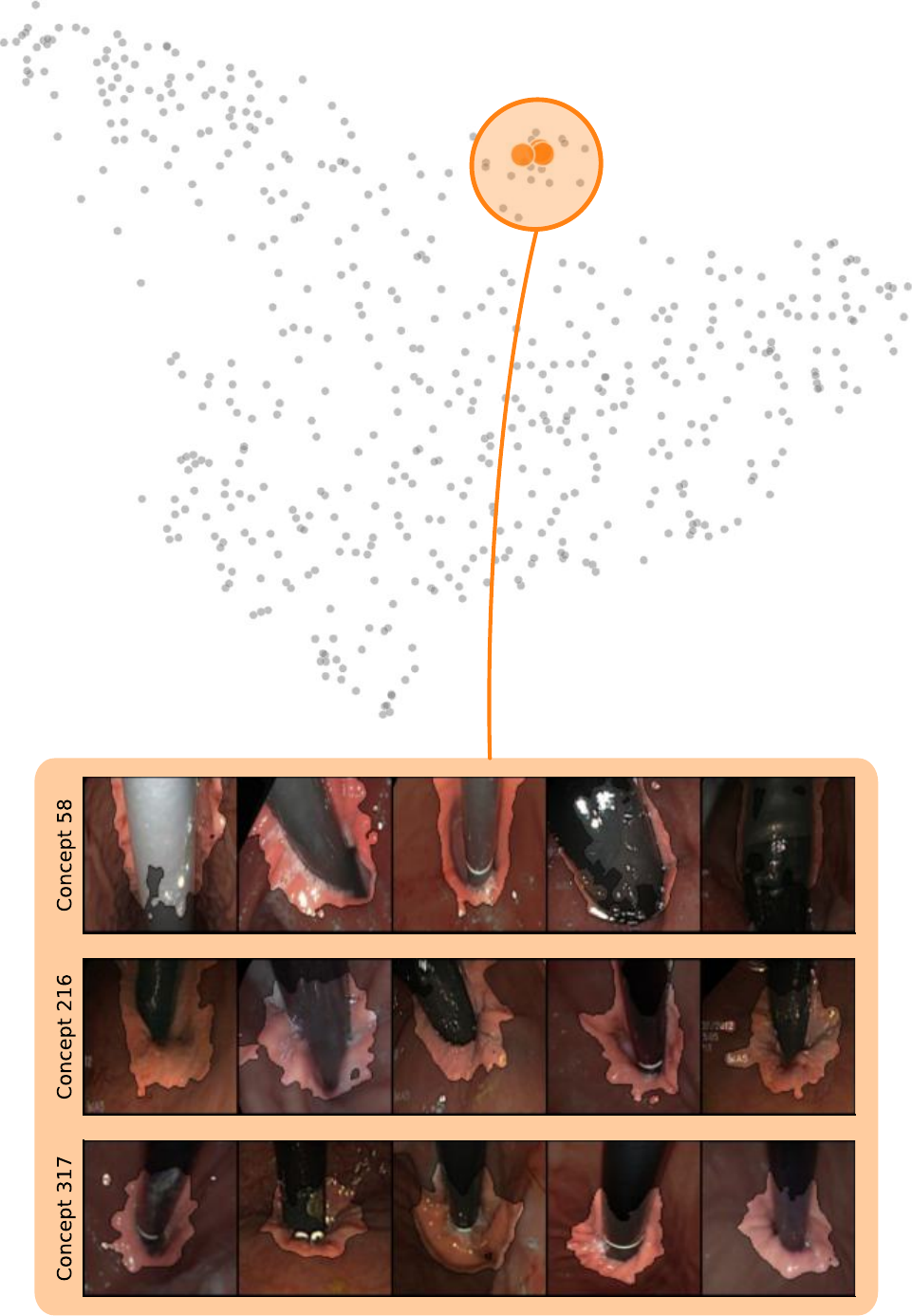}
    \caption{
    Artifact identification on the last ($13^\text{th}$) conv layer of a VGG16 model trained on HyperKvasir with samples from the ``no-disease''-class using \gls{spray} on latent relevances for the data perspective (\emph{left}) and pair-wise cosine similarities for the model perspective (\emph{right}). 
    While not being a clear outlier, there is a coherent set of samples (data perspective) and concepts (model perspective) focussing on the \texttt{insertion tube} artifact.
    }
    \label{app:fig:reveal:kvasir_insertion_tube}
\end{figure*}

\begin{figure*}[t!]
    \centering
    \includegraphics[width=.5\textwidth]{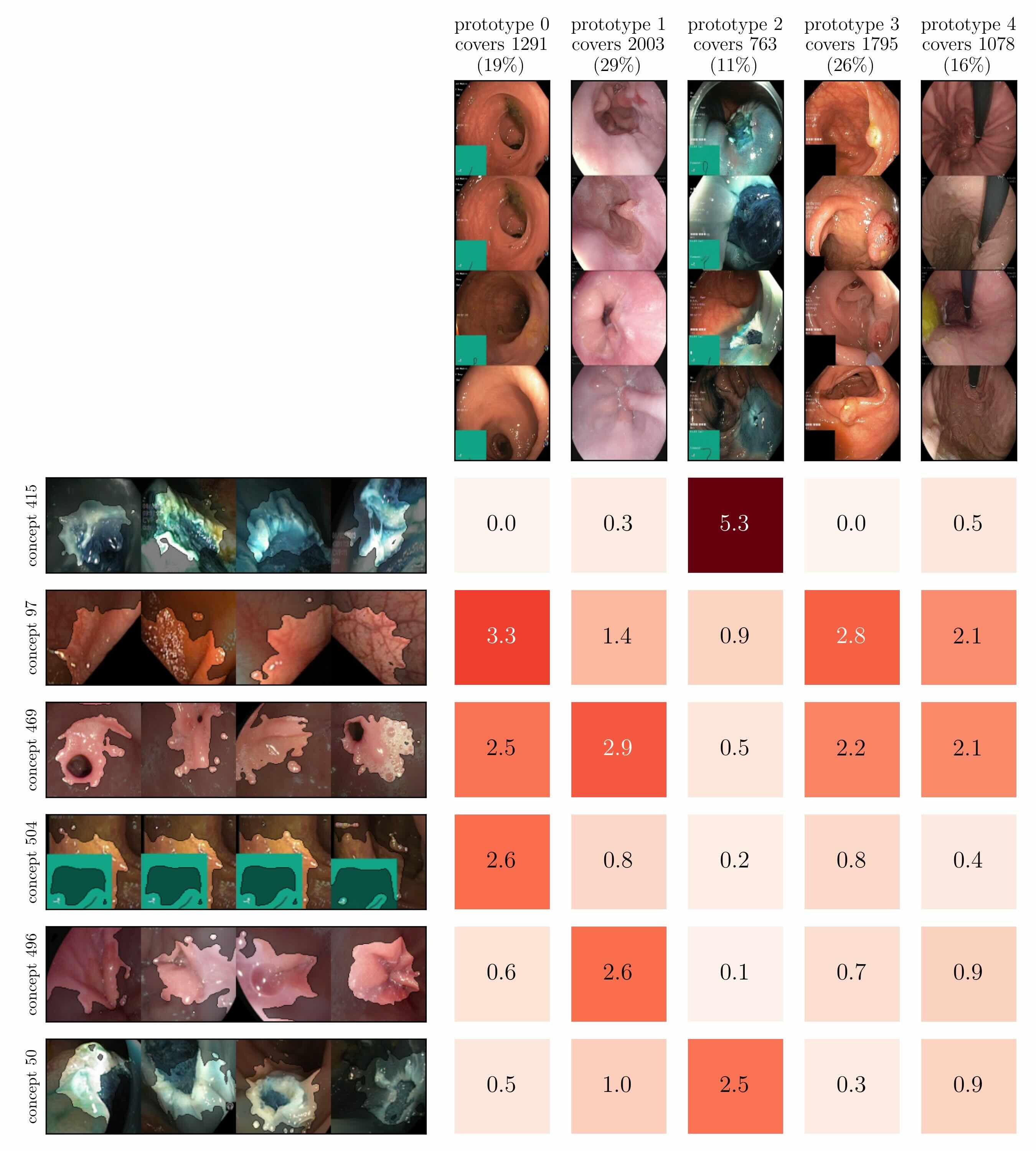}
    \quad
    \includegraphics[width=.45\textwidth]{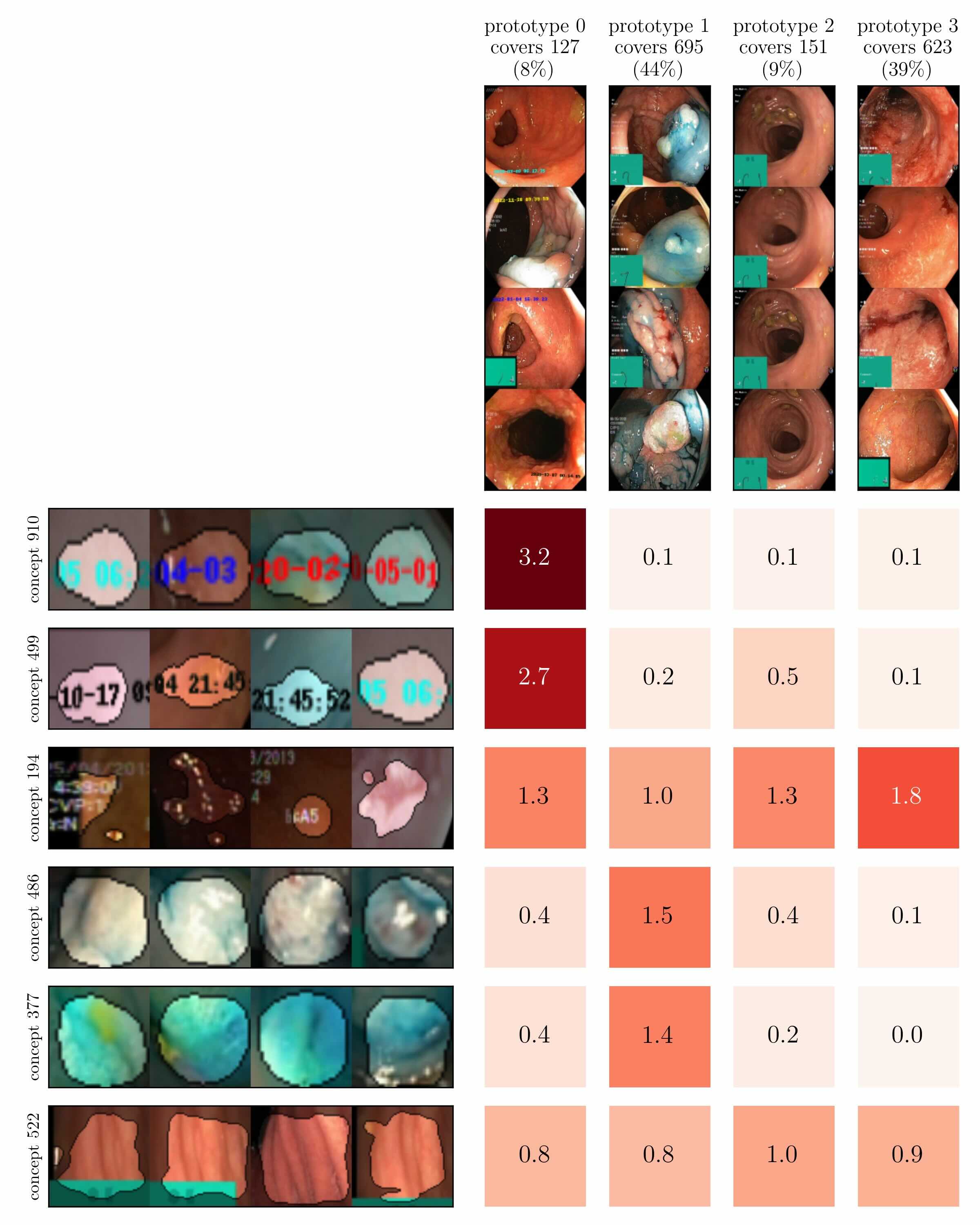}
    \caption{
    \gls{pcx} visualizations using latent relevances after the last ($13^\text{th}$) conv layer of a VGG16 model trained on HyperKvasir (\emph{left}) and after the $3^\text{rd}$ residual block of a ResNet50 model trained on HyperKvasir with the controlled \texttt{timestamp} artifact (\emph{right}). 
    Only samples from the class ``no disease'' are considered.
    Columns represent prototypes with four representative samples, rows represent concepts (visualized via RelMax) and the values in the matrix indicate the average relevance of the concept for the prototype. 
    For clean HyperKvasir (\emph{left}), \texttt{prototype 4} focuses on the \texttt{insertion tubes}. However, no artifact-related concept is associated with the prototype. 
    For the controlled HyperKvasir experiment (\emph{right}), \texttt{prototype 0} focuses on the \texttt{timestamp} artifact with high scores for the related concepts \texttt{$\#$910} and \texttt{$\#$499}.
    }
    \label{app:fig:reveal:pcx_kvasir_clean_and_attacked}
\end{figure*}

\paragraph{Vision Data (VGG16 and ResNet50)}
In Fig.~\ref{app:fig:reveal:kvasir_insertion_tube}, we apply bias detection approaches from the data perspective (\emph{left}) and model perspective (\emph{right}) to the VGG16 model trained on HyperKvasir using relevance scores for neurons on the last ($13^\text{th}$) conv layer. 
We detect a coherent, but not outlier cluster with samples containing the \texttt{insertion tube} from the data perspective, as well as a set of concepts focusing on the artifact from the model perspective.
The application of \gls{pcx}, however, reveals a sub-strategy for class ``non-disease'' with prototypical samples all containing the artifact, as shown in Fig.~\ref{app:fig:reveal:pcx_kvasir_clean_and_attacked} (\emph{left}).

\begin{figure*}[t!]
    \centering
    \includegraphics[width=.35\textwidth]{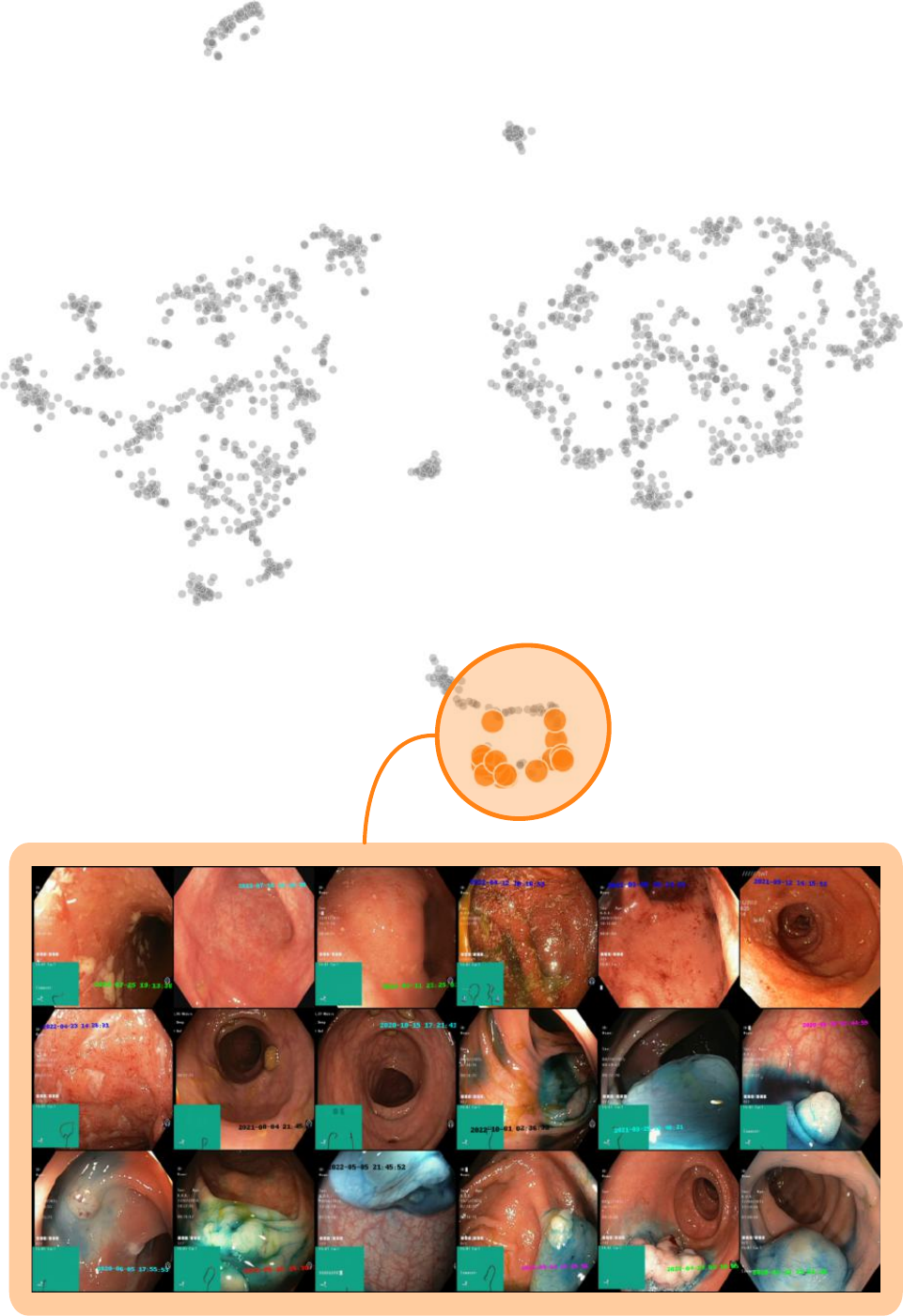}
    \quad\quad\quad\quad
    \includegraphics[width=.35\textwidth]{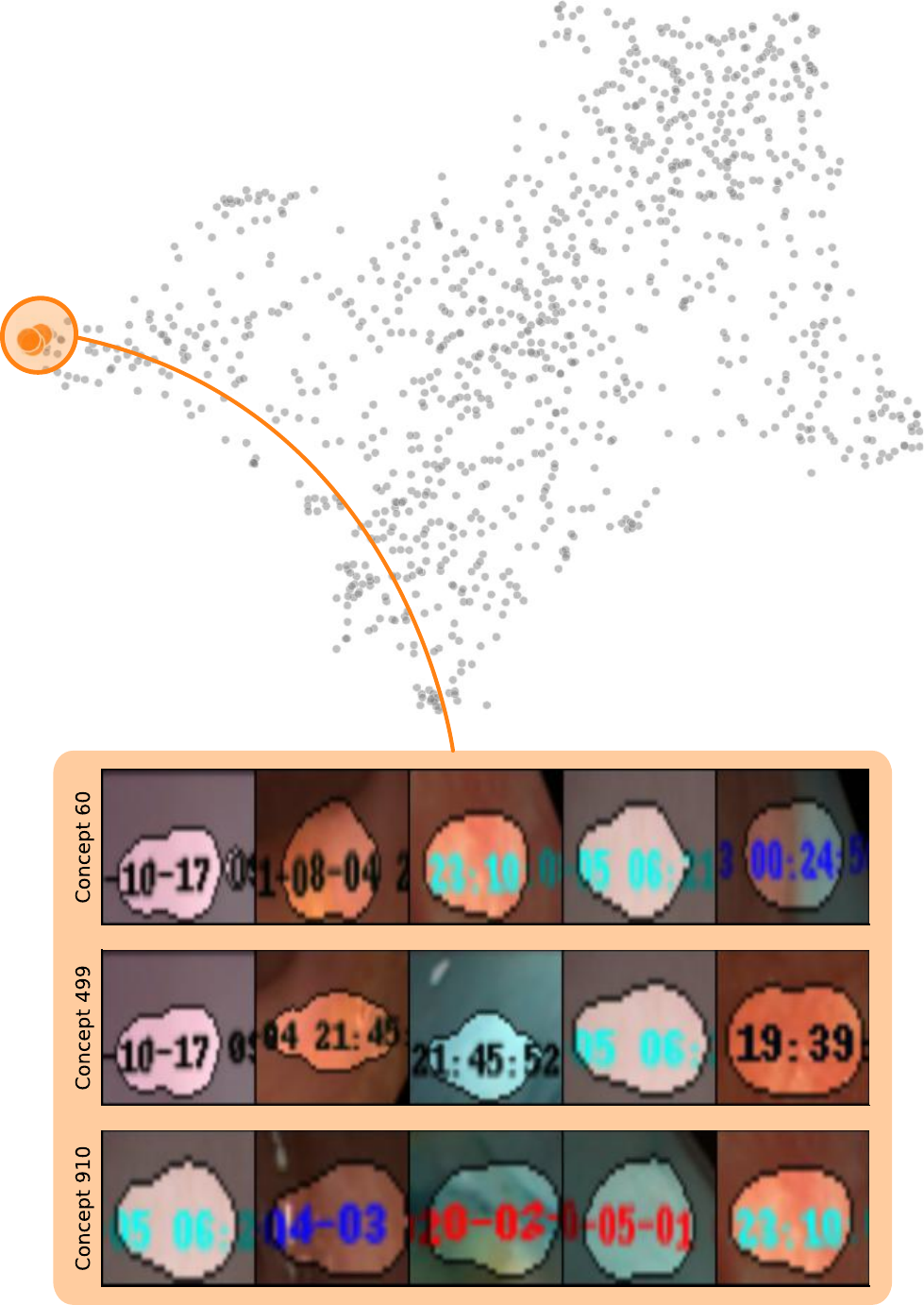}
    \caption{
    Bias identification after the $3^{\text{rd}}$ residual block of a ResNet50 model trained on HyperKvasir using the controlled \texttt{timestamp} artifact with samples from the ``disease''-class using \gls{spray} on latent relevances for the data perspective (\emph{left}) and pair-wise cosine similarities for the model perspective (\emph{right}). 
    We identify a clear outlier cluster with biased samples (data perspective) and a coherent set of concepts focusing on the considered artifact (model perspective).
    }
    \label{app:fig:reveal:kvasir_attacked}
\end{figure*}

Next, we apply our bias detection approaches to the ResNet50 model trained on HyperKvasir with the controlled \texttt{timestamp} artifact using relevance scores after the $3^\text{rd}$ residual block for samples from the attacked class (``disease'') in Fig.~\ref{app:fig:reveal:kvasir_attacked}.
\gls{spray} reveals a clear outlier cluster of poisoned samples containing the artifact (\emph{left}) and the concept clustering reveals a coherent set of concepts focusing on the inserted \texttt{timestamp} (\emph{right}, see neurons \texttt{$\#$60}, \texttt{$\#$499}, and \texttt{$\#$910}).
We further apply \gls{pcx} in Fig.~\ref{app:fig:reveal:pcx_kvasir_clean_and_attacked} (\emph{right}), clearly revealing \texttt{prototype 0} with related concepts \texttt{$\#$499} and \texttt{$\#$910} as prediction sub-strategy focusing on the inserted \texttt{timestamp} artifact.

\begin{figure*}[t!]
    \centering
    \includegraphics[width=.35\textwidth]{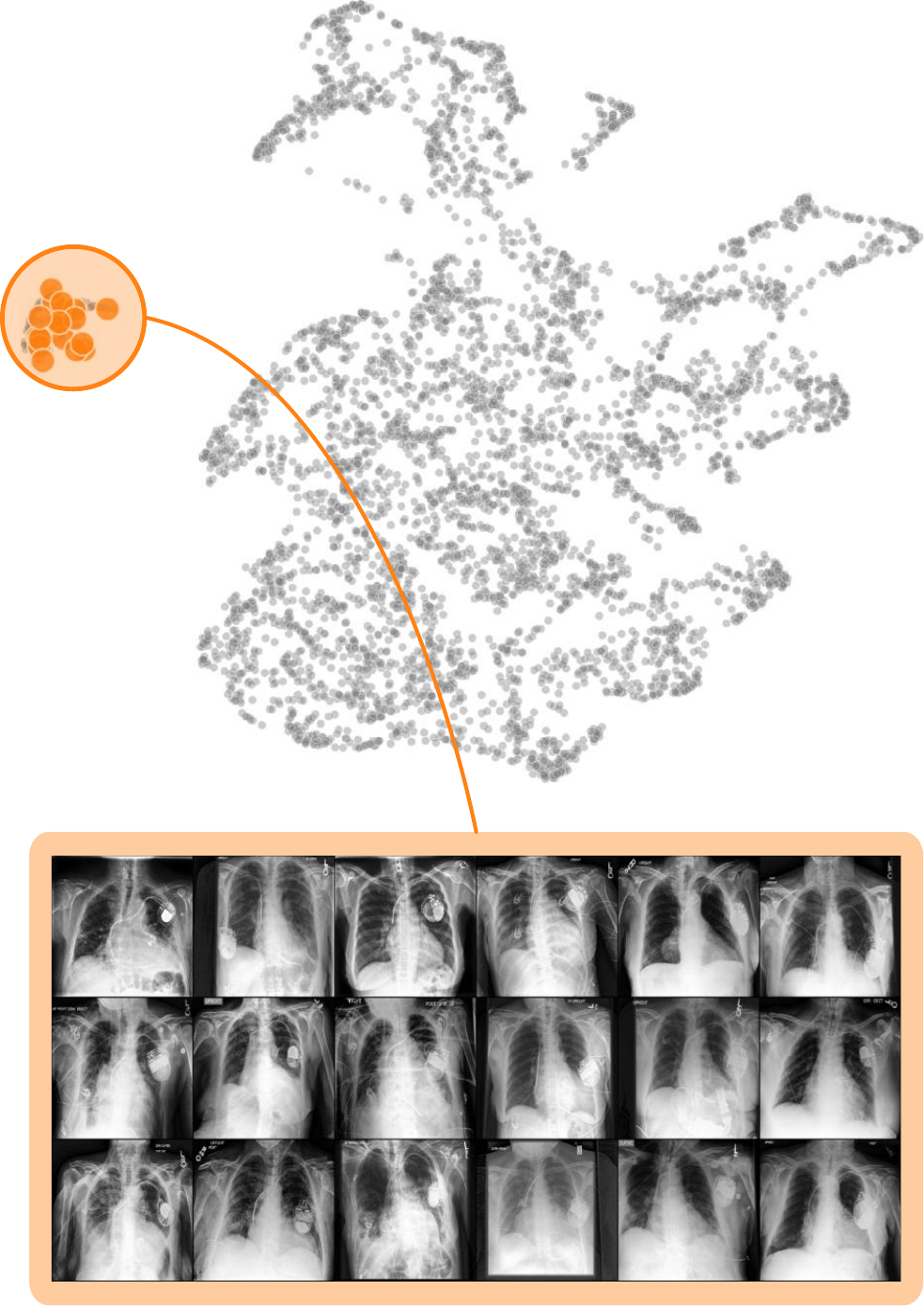}
    \quad\quad\quad\quad
    \includegraphics[width=.35\textwidth]{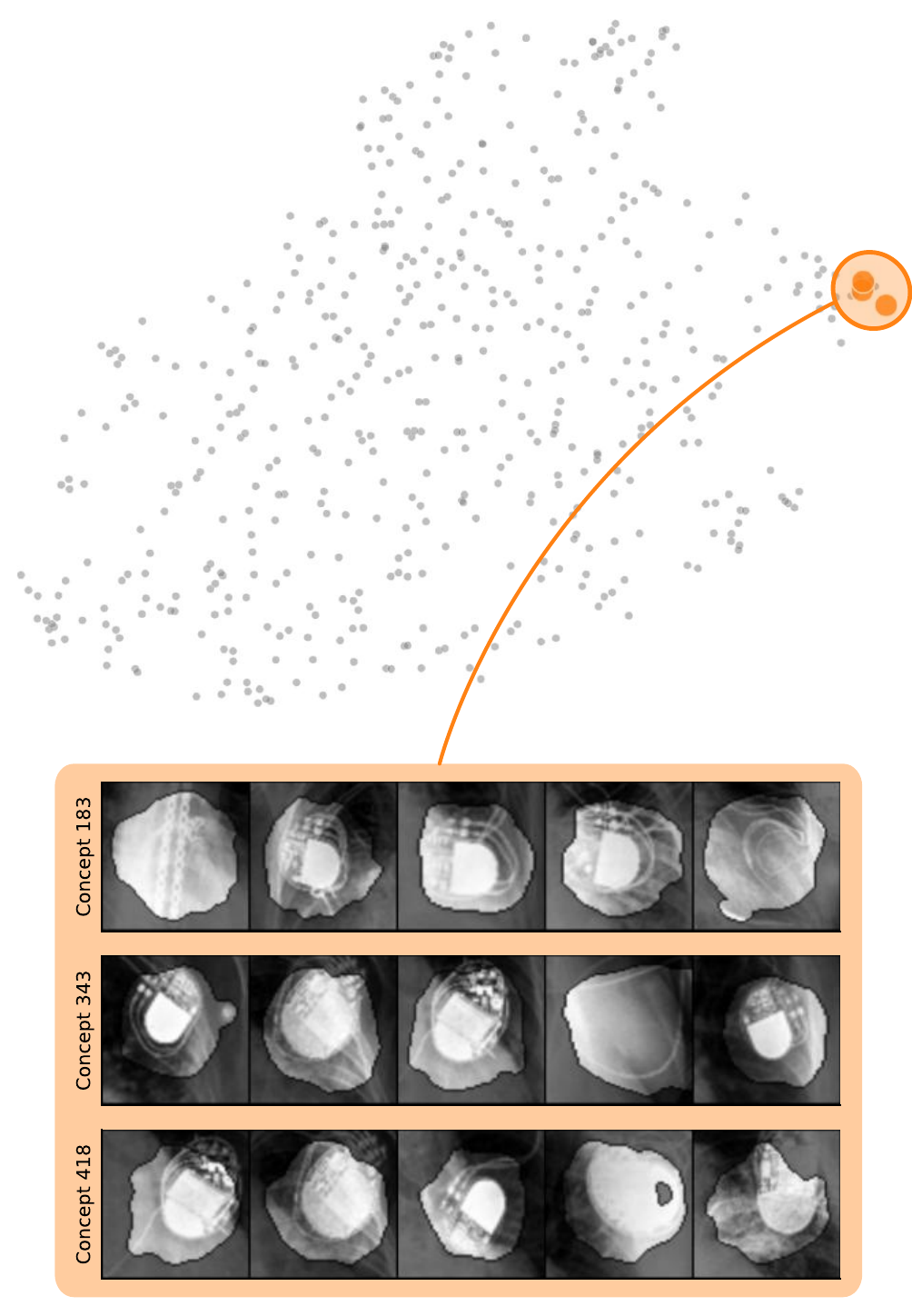}
    \caption{
    Bias identification after the $10^{\text{th}}$ conv layer of a VGG16 model trained on CheXpert with samples from the ``cardiomegaly''-class using \gls{spray} on latent relevances for the data perspective (\emph{left}) and pair-wise cosine similarities for the model perspective (\emph{right}). 
    We identify a clear outlier cluster with samples containing the \texttt{pacemaker} artifact (\emph{left}) and an outlier set of concepts focusing on the artifact (\emph{right}).
    }
    \label{app:fig:reveal:chexpert_pacemaker}
\end{figure*}

\begin{figure*}[t!]
    \centering
    \includegraphics[width=.45\textwidth]{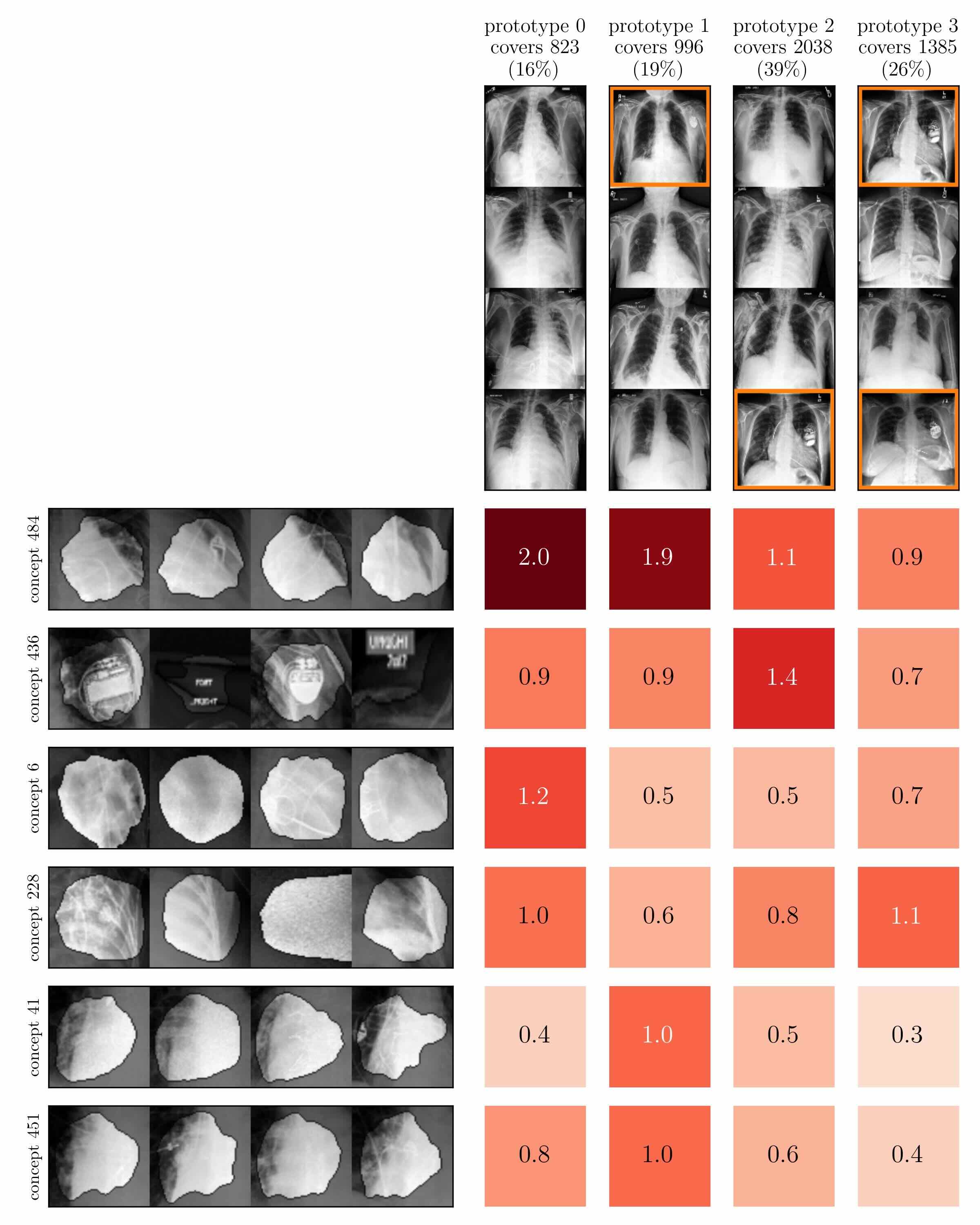}
    \quad
    \includegraphics[width=.45\textwidth]{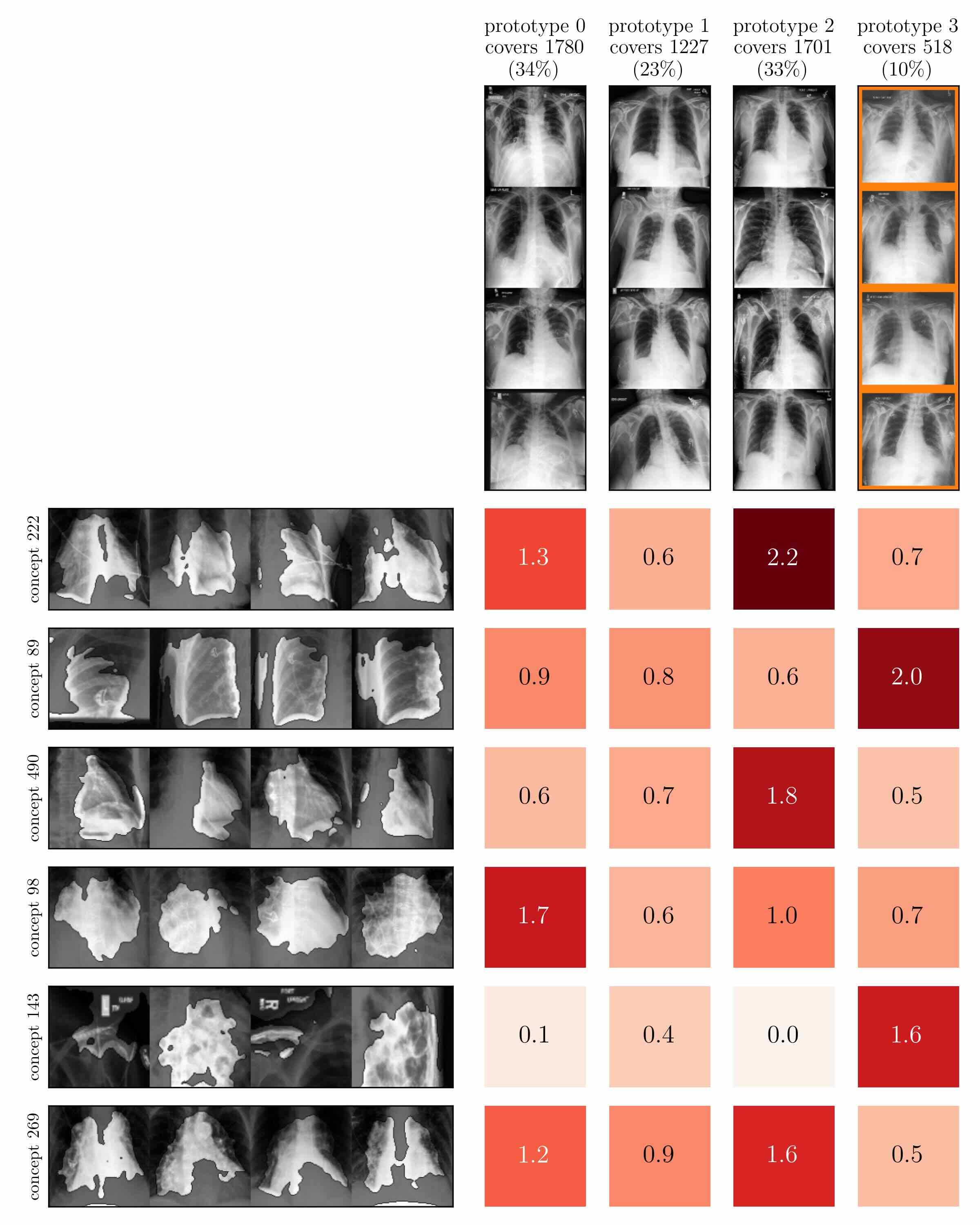}
    \caption{
    \gls{pcx} visualizations using latent relevances after the $12^\text{th}$ and $10^\text{th}$ conv layers of VGG16 models trained on the original CheXpert data (\emph{left}) and with controlled \texttt{brightness} artifact (\emph{right}). 
    Only samples from the class ``cardiomegaly'' are considered. 
    Columns represent prototypes with four representative samples, rows represent concepts (visualized via RelMax) and the values in the matrix indicate the average relevance of the concept for the prototype. 
    To ease readability, prototypical samples containing the expected artifacts \texttt{pacemaker} (\emph{right}) and \texttt{brightness} (\emph{left}) are highlighted with an orange box.
    For \texttt{pacemaker} (\emph{left}), the artifact appears in multiple prototypes, but concept \texttt{$\#$436}, relevant to most prototypes, appears to focus on the artifact.
    For the controlled \texttt{brightness} artifact (\emph{right}), \texttt{prototype 3} focuses on the considered artifact with high scores for the hard-to-interpret concepts (neurons) \texttt{$\#$89} and \texttt{$\#$143}.
    }
    \label{app:fig:reveal:pcx_chexpert}
\end{figure*}

For CheXpert, in Fig.~\ref{app:fig:reveal:chexpert_pacemaker} we reveal an outlier cluster of samples containing the \texttt{pacemaker} artifact from the data perspective (\emph{left}), and concepts focusing on the artifact from the model perspective (\emph{right}) using relevances for neurons on the $10^\text{th}$ conv layer. 
The application of \gls{pcx} for samples of the class ``cardiomegaly'' does not clearly reveal an impacted prototype, as shown in Fig.~\ref{app:fig:reveal:pcx_chexpert} (\emph{left}).
However, all prototypes have high relevance scores for neuron \texttt{$\#$436}, which appears to focus on artifacts such as the considered \texttt{pacemaker}.

\begin{figure*}[t!]
    \centering
    \includegraphics[width=.35\textwidth]{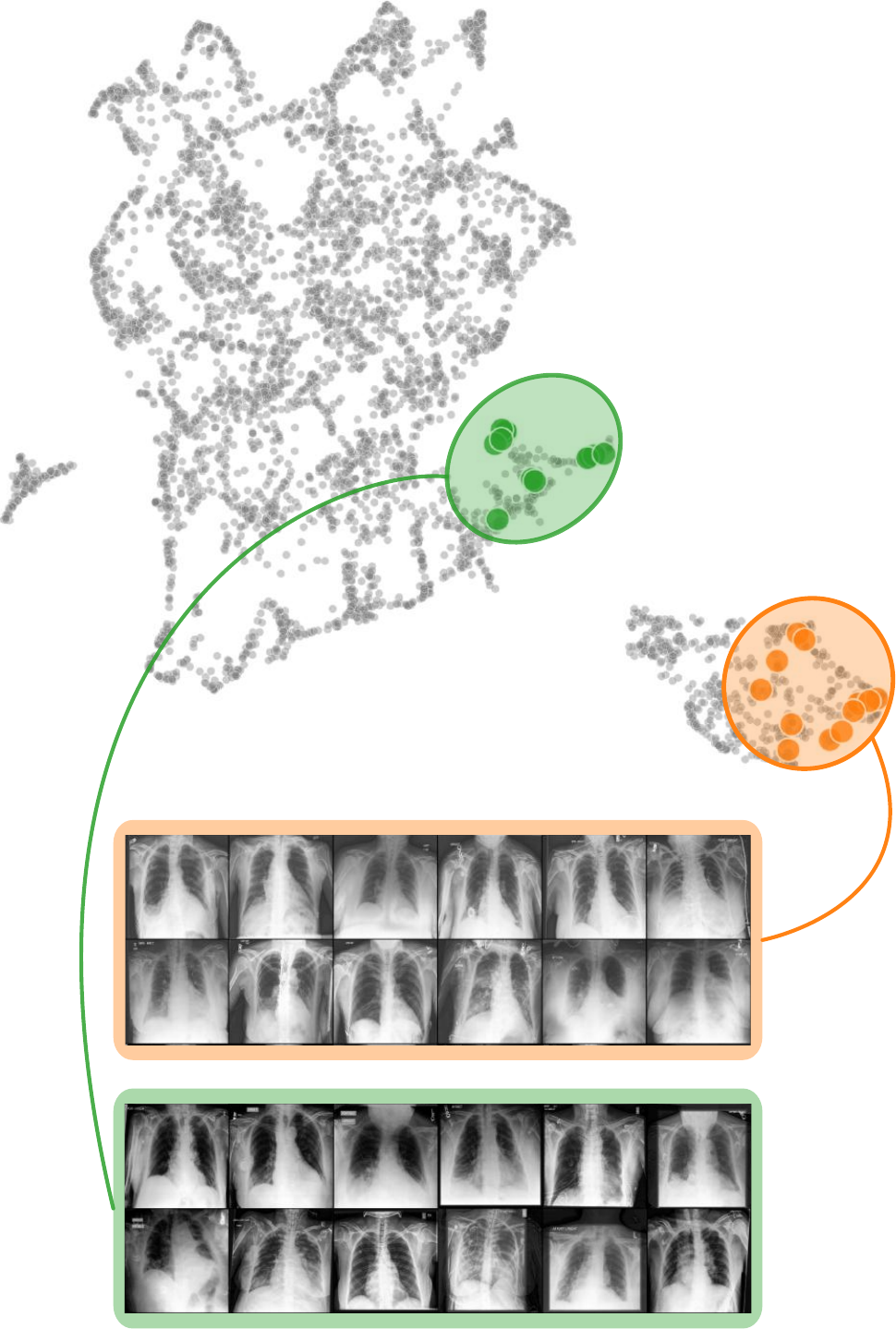}
    \quad\quad\quad\quad
    \includegraphics[width=.35\textwidth]{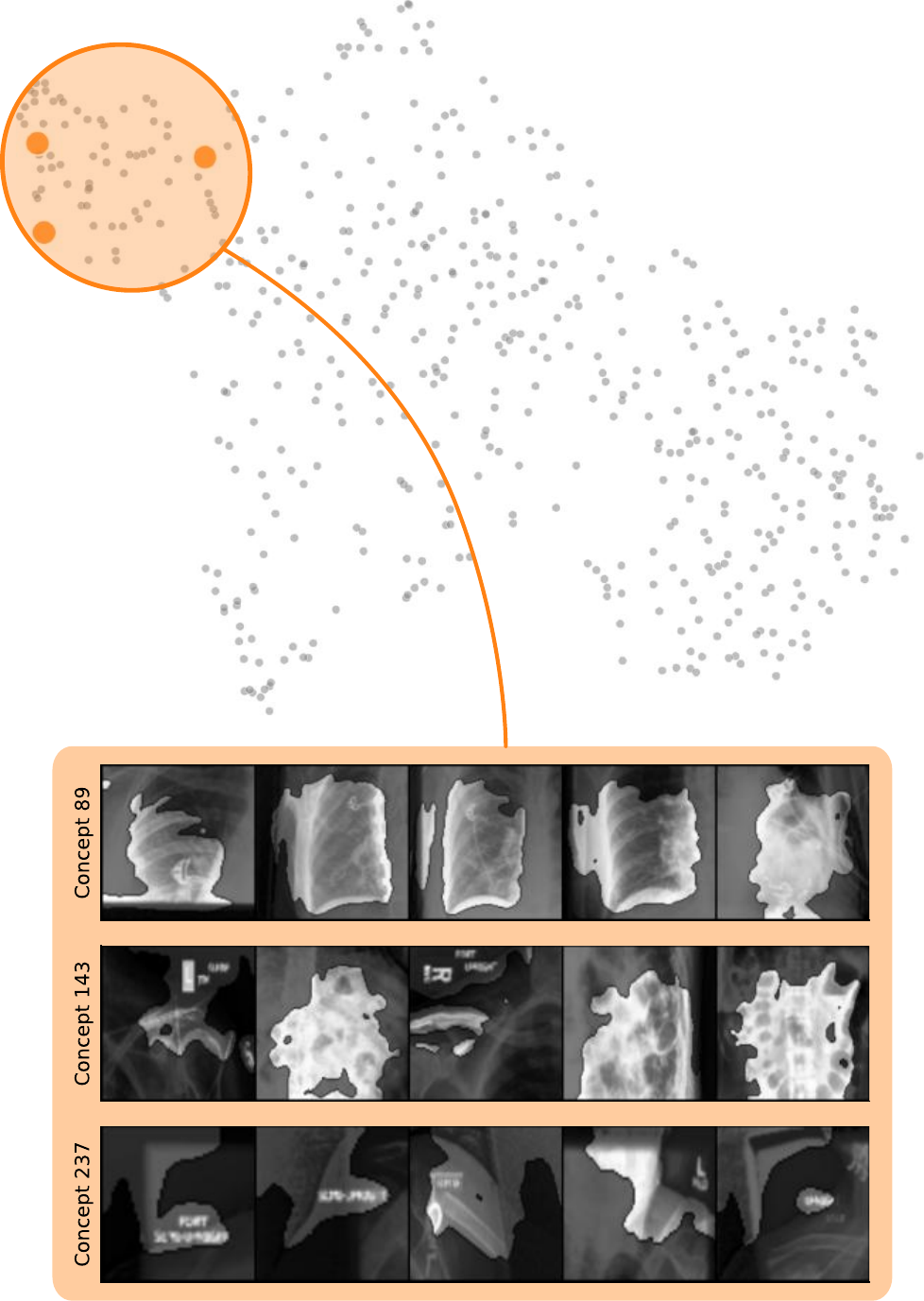}
    \caption{
    Bias identification after the $12^{\text{th}}$ conv layer of a VGG16 model trained on CheXpert using the controlled \texttt{brightness} artifact with samples from the ``cardiomegaly''-class using \gls{spray} on latent relevances for the data perspective (\emph{left}) and pair-wise cosine similarities for the model perspective (\emph{right}). 
    We identify a clear outlier cluster with samples with the \texttt{brightness} artifact ($\mycirc[myorange]$, \emph{left}). 
    For comparison, we further highlight a cluster of samples with clean samples ($\mycirc[mygreen]$).
    Note, that it is challenging to identify the artifact from the model perspective, as related concept visualizations do not reveal the \texttt{brightness}. 
    Since there is no outlier concept, we visualize concepts that are highly relevant for the prediction sub-strategy revealed via \gls{pcx} (see Fig.~\ref{app:fig:reveal:pcx_chexpert}).
    }
    \label{app:fig:reveal:chexpert_attacked}
\end{figure*}

Moreover, we run bias identification approaches for CheXpert with the controlled \texttt{brightness} artifact inserted into class ``cardiomegaly'' using relevance scores for neurons on the $12^\text{th}$ conv layer of the VGG16 model in Fig.~\ref{app:fig:reveal:chexpert_attacked}.
Whereas the data perspective reveals a clear outlier cluster of affected samples (\emph{left}), no outlier concepts can be detected and the interpretation of considered samples \wrt the \texttt{brightness} artifact is challenging (\emph{right}).
Note, that we selected concepts revealed via \gls{pcx} in Fig.~\ref{app:fig:reveal:pcx_chexpert} (\emph{right}), with \texttt{prototype 3} appearing to focus on samples containing the \texttt{brightness} artifact with high relevance scores for neurons \texttt{$\#$89} and \texttt{$\#$143}.

\begin{figure*}[t!]
    \centering
    \includegraphics[width=.35\textwidth]{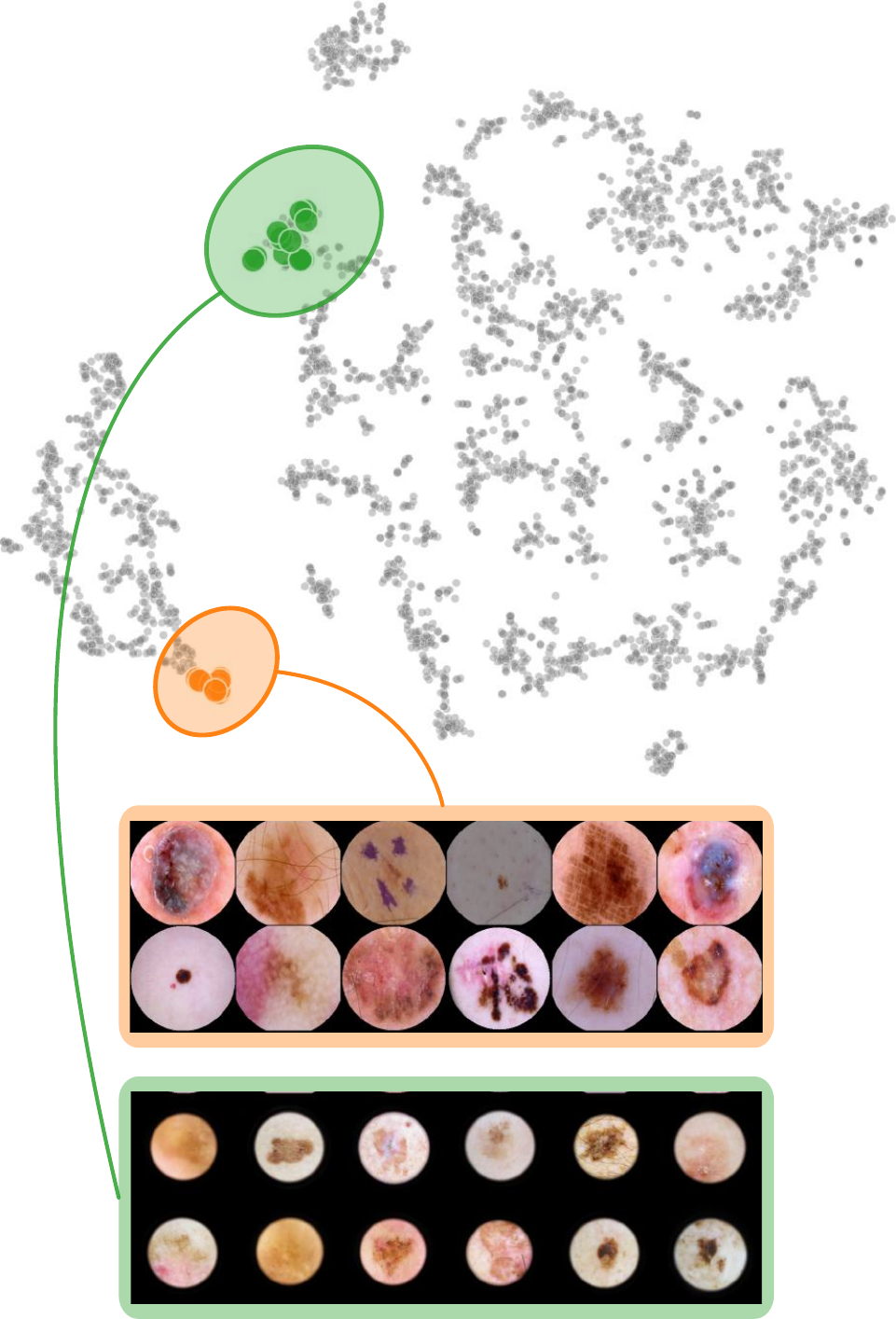}
    \quad\quad\quad\quad
    \includegraphics[width=.35\textwidth]{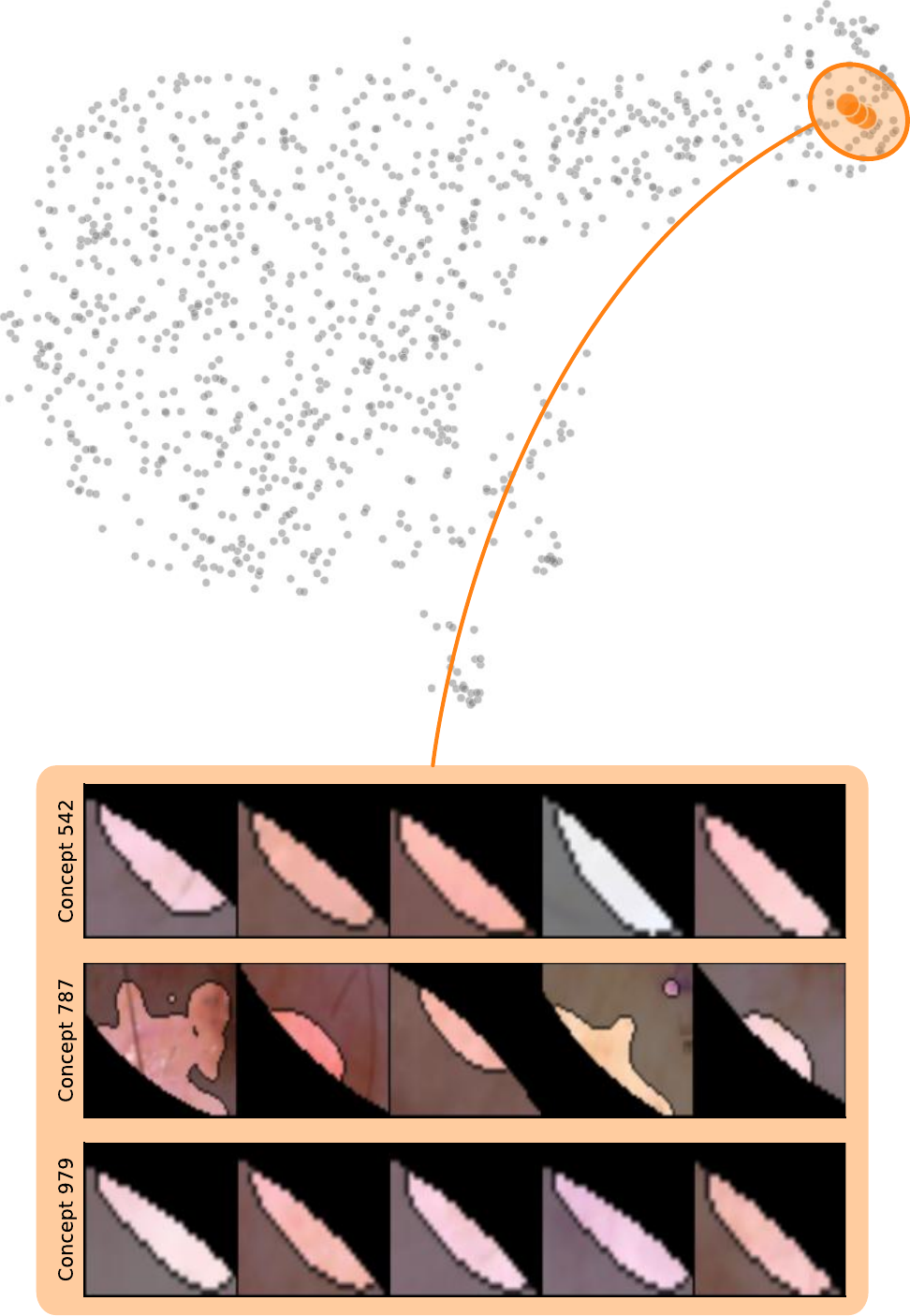}
    \caption{
    Bias identification after the $3^{\text{rd}}$ residual block of a ResNet50 model trained on ISIC2019 using the controlled \texttt{microscope} artifact with samples from the ``melanoma''-class using \gls{spray} on latent relevances for the data perspective (\emph{left}) and pair-wise cosine similarities for the model perspective (\emph{right}). 
    We identify multiple clusters with samples with different versions of the \texttt{microscope} artifact (\emph{left}) and a large outlier set of concepts focusing on the black circle (\emph{right}).
    }
    \label{app:fig:reveal:isic_attacked}
\end{figure*}

\begin{figure*}[t!]
    \centering
    \includegraphics[width=.45\textwidth]{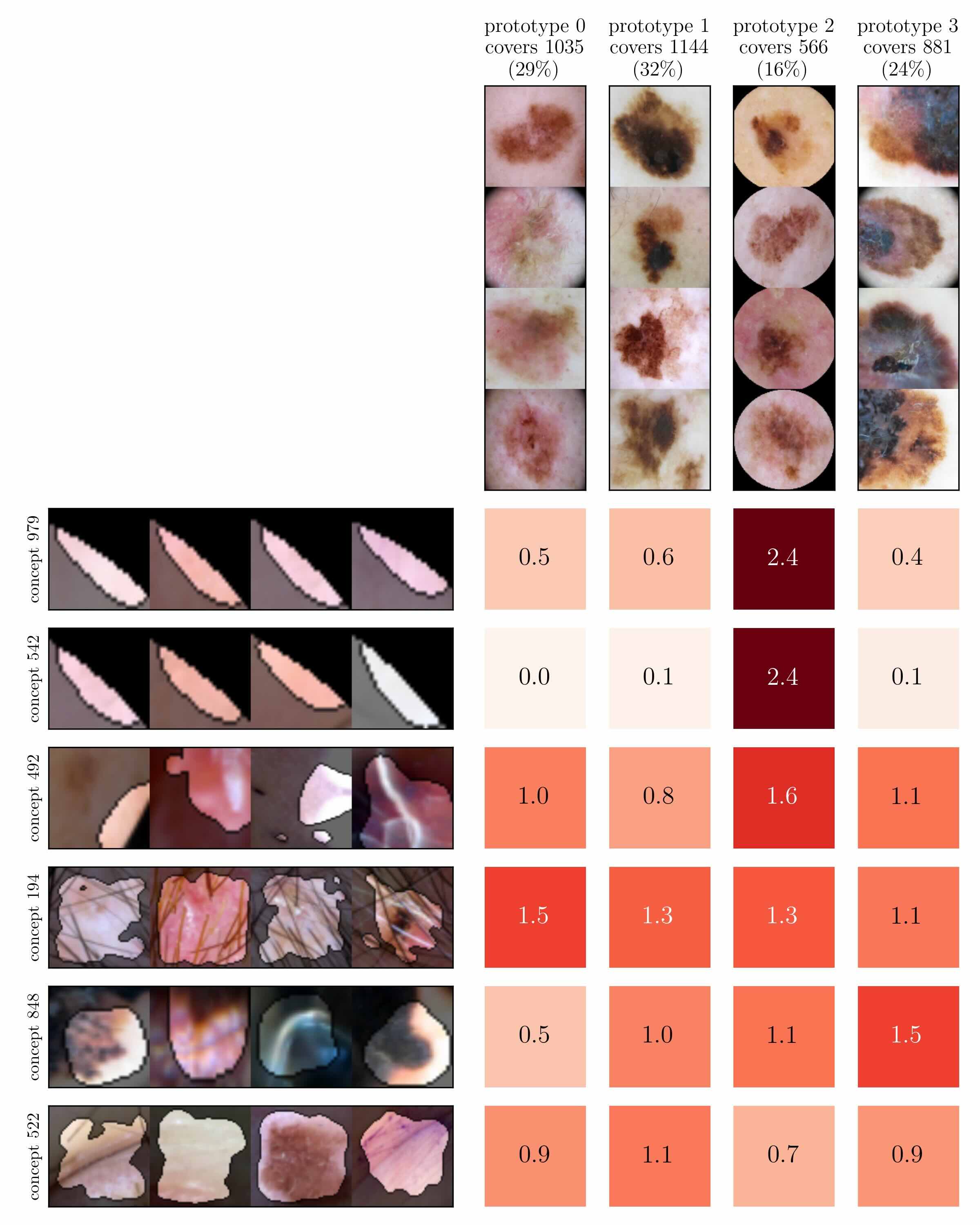}
    \quad
    \includegraphics[width=.45\textwidth]{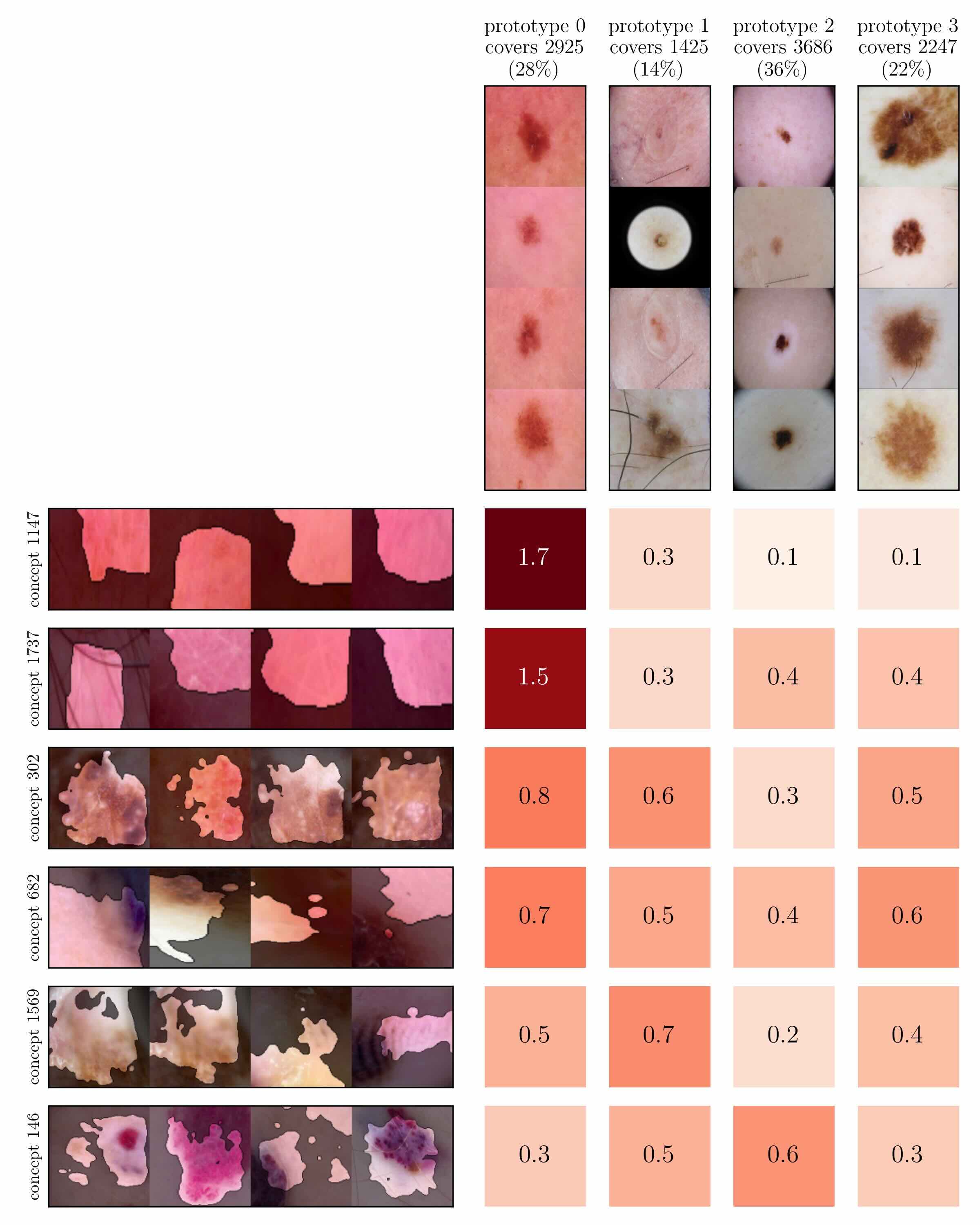}
    \caption{
    \gls{pcx} visualizations using latent relevances after the $3^\text{rd}$ residual block of a ResNet50 model trained on ISIC2019 with the controlled \texttt{microscope} artifact (\emph{left}, and after the $4^\text{th}$ residual block of a ResNet50 model trained on the clean ISIC2019 data (\emph{right}).
    For the former, only samples from the attacked class ``melanoma'' are considered, and for the latter we use samples from class ``melanocytic nevus''.
    Columns represent prototypes with four representative samples, rows represent concepts (visualized via RelMax) and the values in the matrix indicate the average relevance of the concept for the prototype. 
    For the attacked data (\emph{left}), \texttt{prototype 2} focuses on the inserted \texttt{microscope} artifact with high scores for the related concepts (neurons) \texttt{$\#$979} and \texttt{$\#$543}.
    For the clean ISIC2019 data (\emph{right}), interestingly, \texttt{prototype 0} focuses \texttt{red-colored skin} with high scores for the related concepts \texttt{$\#$1147} and \texttt{$\#$1737}.
    }
    \label{app:fig:reveal:pcx_isic_attacked_isic1}
\end{figure*}

For ISIC2019 with the controlled \texttt{microscope} artifact, we show bias identification results using relevance scores after the $3^\text{rd}$ residual block of the ResNet50 model in Fig.~\ref{app:fig:reveal:isic_attacked}. 
The data perspective (\emph{left}) reveals an outlier cluster of samples focusing on the inserted artifact ($\mycirc[myorange]$), and, interestingly, also a cluster focusing on a different type of black circles ($\mycirc[mygreen]$).
The model perspective (\emph{right}) reveals a large outlier set of concepts focusing on the black border caused by the \texttt{microscope} artifact.
Similarly, the application of \gls{pcx} in Fig.~\ref{app:fig:reveal:pcx_isic_attacked_isic1} (\emph{left}) leads to \texttt{prototype 2} as sub-strategy for the attacked class, clearly using the inserted artifact with high relevance scores for neurons \texttt{$\#$910} and \texttt{$\#$499}.

\begin{figure*}[t]
    \centering
    \includegraphics[width=.35\textwidth]{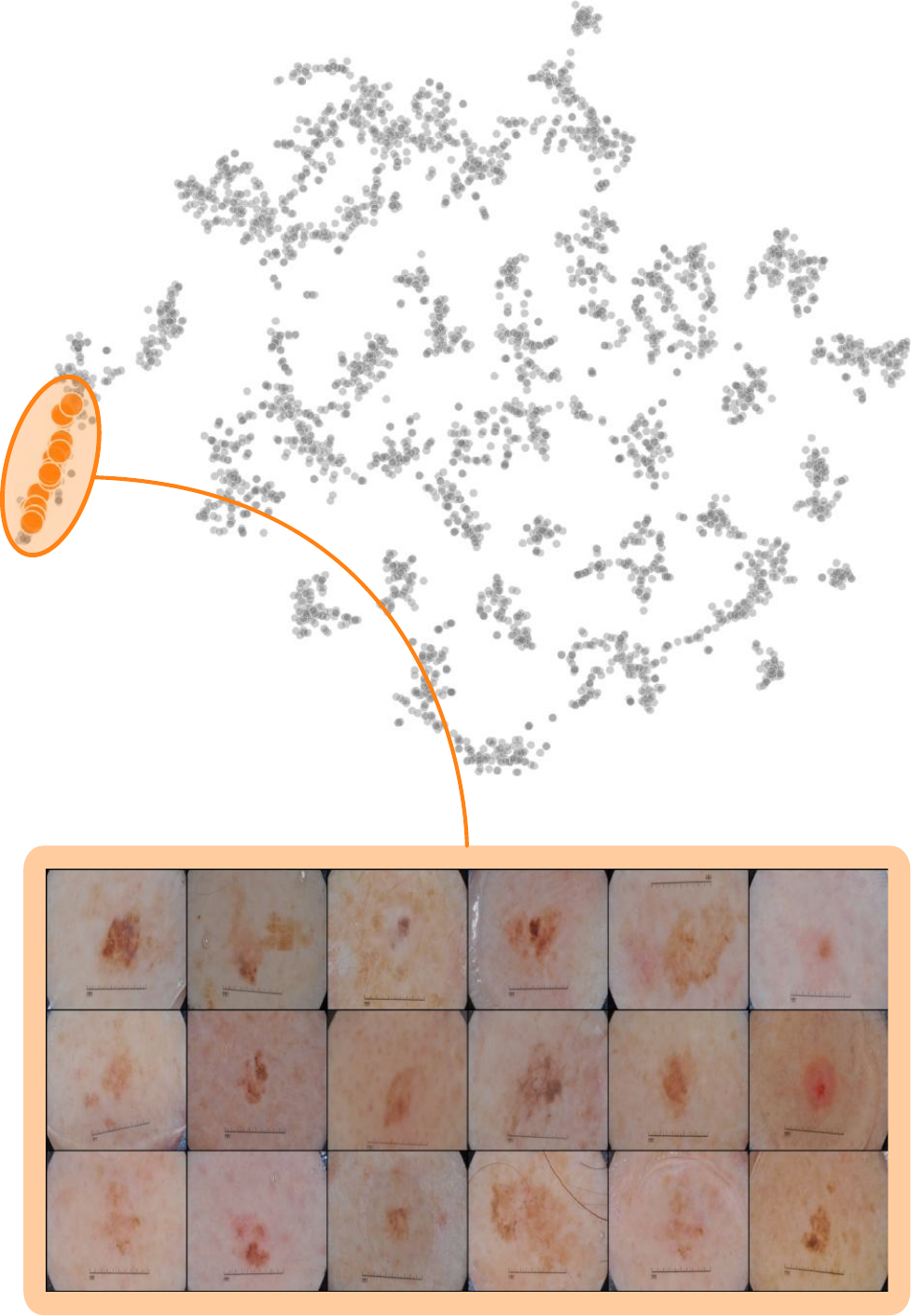}
    \quad\quad\quad\quad
    \includegraphics[width=.35\textwidth]{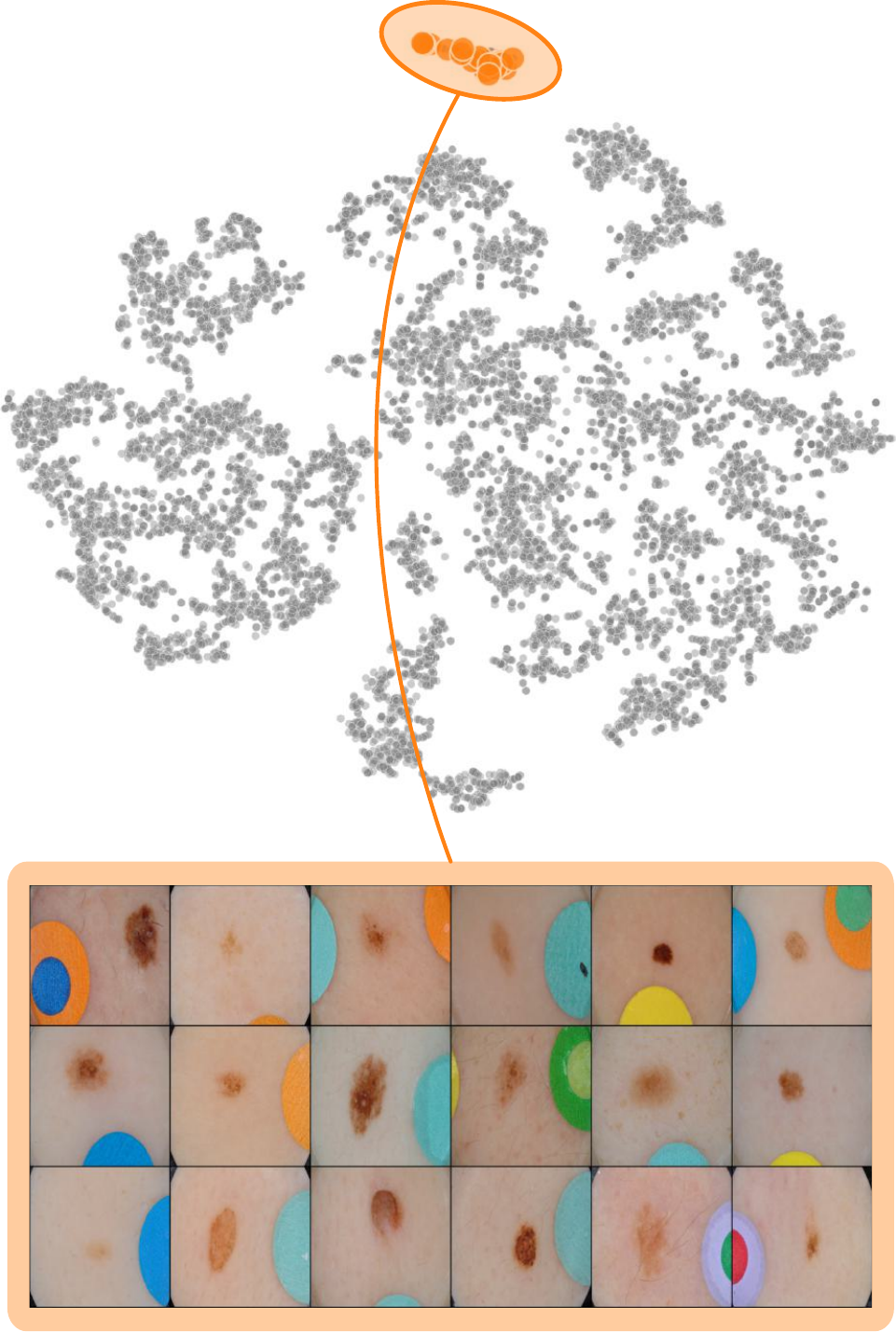}
    \caption{
    Bias identification from the data perspective using \gls{spray} with relevances after the $3^{\text{rd}}$ residual block of a ResNet50 model trained on ISIC2019 using samples from classes ``melanoma'' (\emph{left}) and ``melanocytic nevus'' (\emph{right}).
    We identify clear outlier clusters with samples containing the \texttt{ruler} (\emph{left}, ``melanoma'') and \texttt{band-aid} (\emph{right}, ``melanocytic nevus'') artifacts.
    }
    \label{app:fig:reveal:isic_spray_0_1}
\end{figure*}

\begin{figure*}[t]
    \centering
    \includegraphics[width=.35\textwidth]{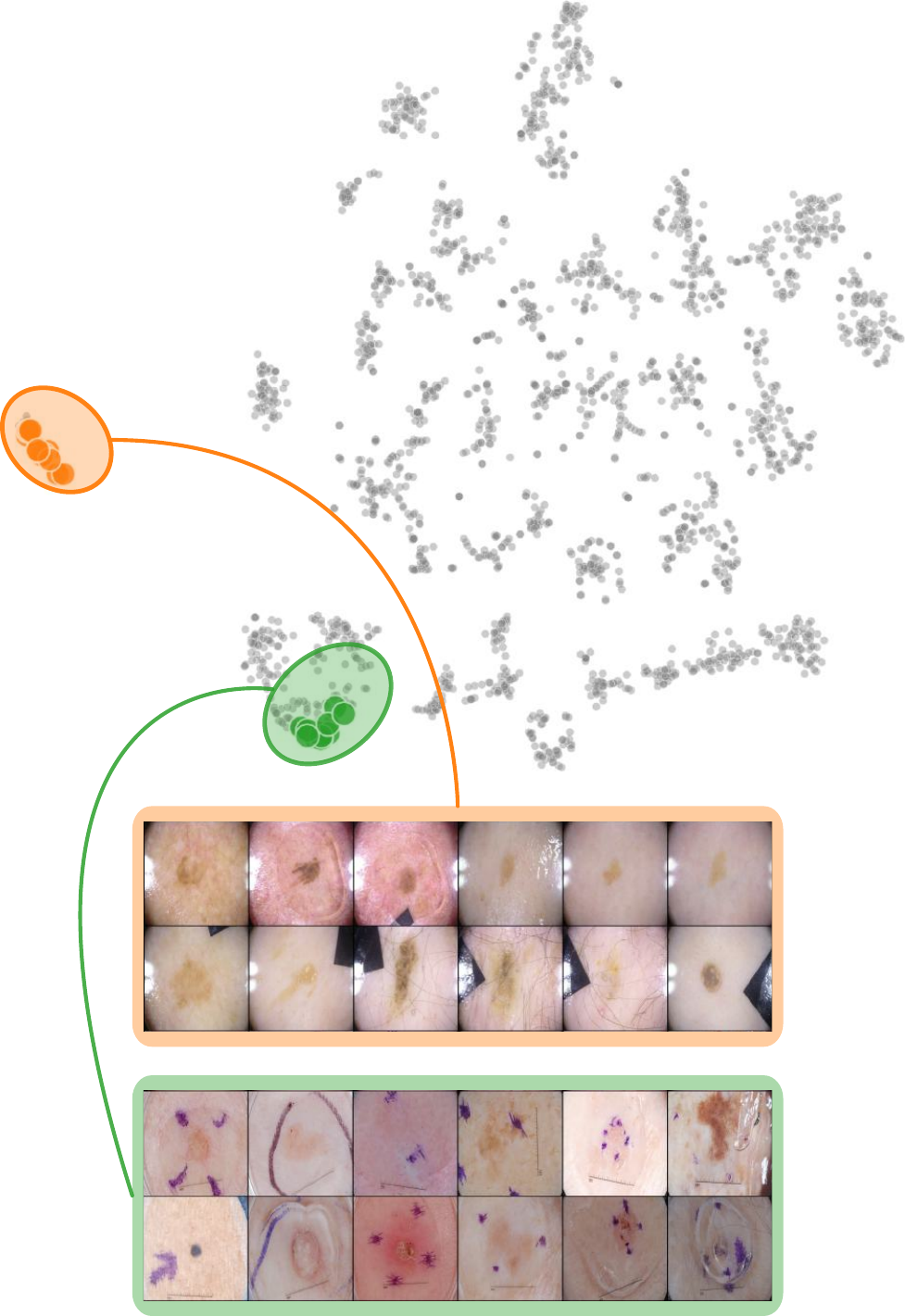}
    \caption{
    Bias identification from the data perspective using \gls{spray} with relevances after the $3^{\text{rd}}$ residual block of a ResNet50 for ISIC2019 using samples from the class ``benign keratosis''.
    We identify outlier clusters with samples containing the \texttt{reflection} ($\mycirc[myorange]$) and \texttt{skin marker} ($\mycirc[mygreen]$) artifacts.
    }
    \label{app:fig:reveal:isic_spray_4}
\end{figure*}

\begin{figure*}[t!]
    \centering
    \includegraphics[width=.35\textwidth]{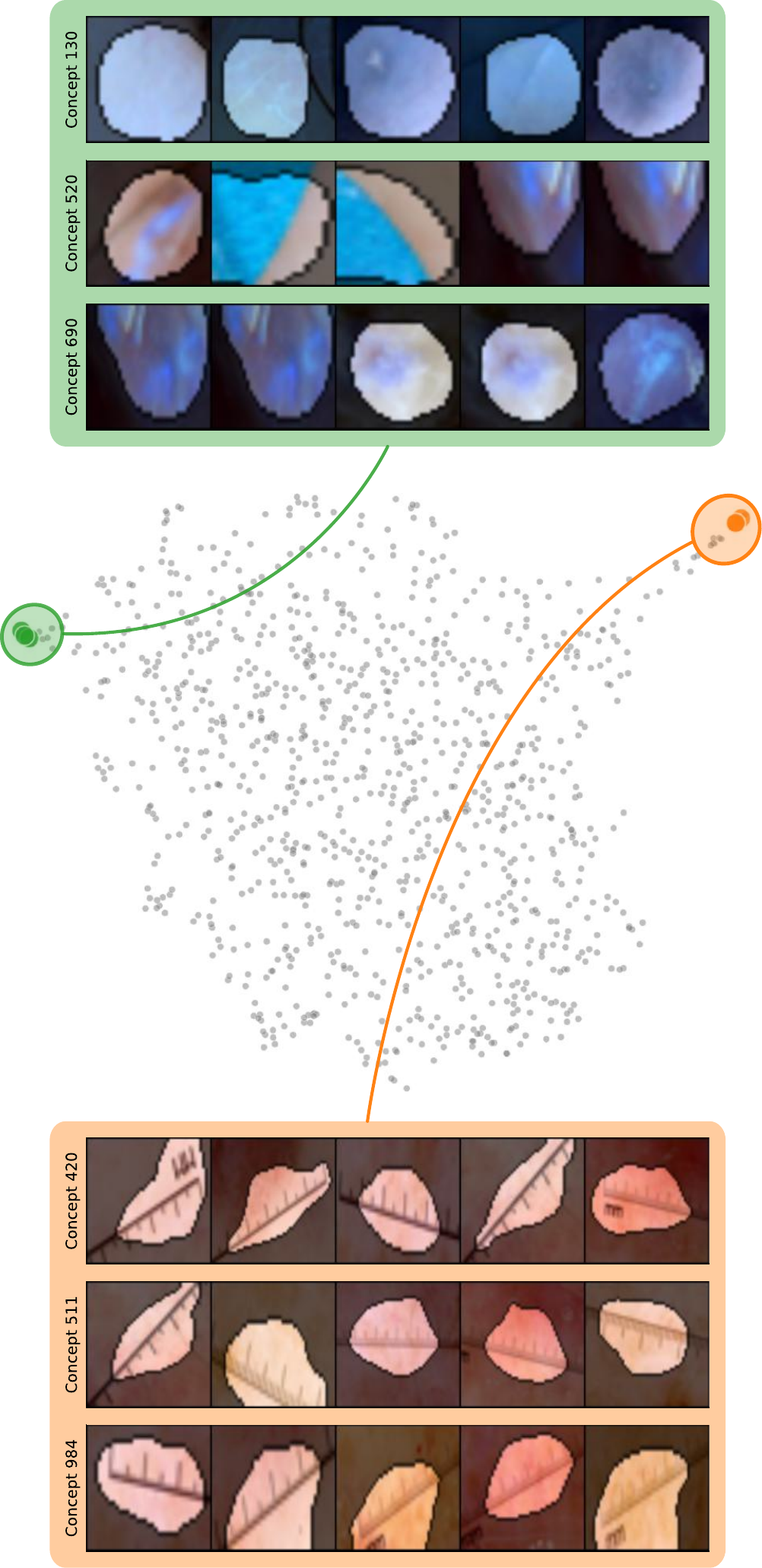}
    \quad\quad\quad\quad
    \includegraphics[width=.35\textwidth]{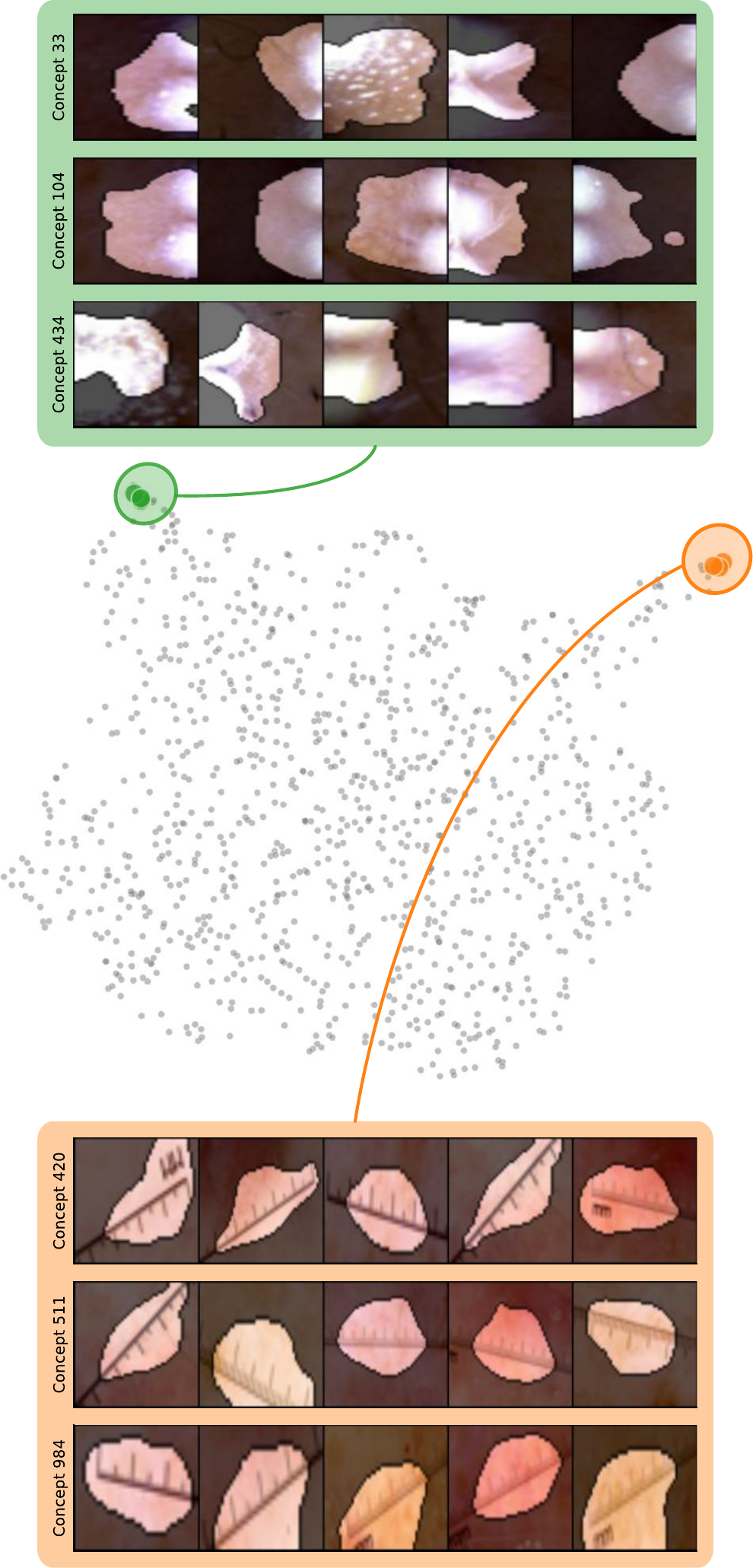}
    \caption{
    Bias identification from the model perspective using pair-wise cosine similarities between max-pooled relevances after the $3^{\text{rd}}$ residual block of a ResNet50 for ISIC2019 using samples from the classes ``melanoma'' (\emph{left}) and ``benign keratosis'' (\emph{right}).
    For ``melanoma'', we identify outlier concepts focusing on the \texttt{ruler} ($\mycirc[myorange]$) and \texttt{blueish tint} ($\mycirc[mygreen]$) artifacts.
    For ``benign keratosis'', We identify outlier concepts focussing on \texttt{reflections} ($\mycirc[mygreen]$), and, again \texttt{rulers} ($\mycirc[myorange]$).
    For the latter, predictions for both considered classes use exactly the same neurons.
    }
    \label{app:fig:reveal:isic_crp_0_4}
\end{figure*}

\begin{figure*}[t!]
    \centering
    \includegraphics[width=.6\textwidth]{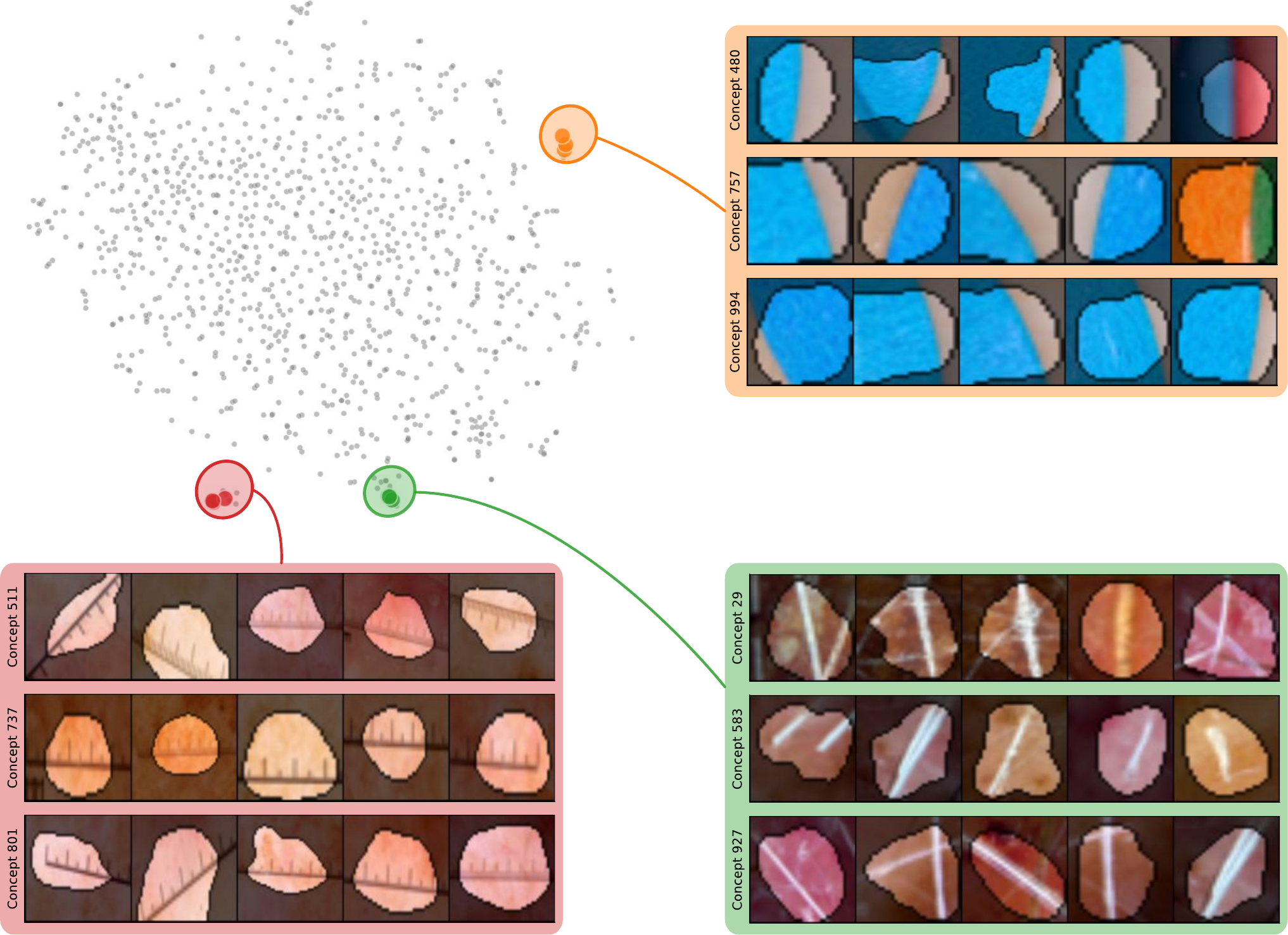}
    \caption{
    Bias identification from the model perspective using pair-wise cosine similarities between max-pooled relevances after the $3^{\text{rd}}$ residual block of a ResNet50 for ISIC2019 using samples from the class ``melanocytic nevus''.
    We identify outlier concepts focusing on the \texttt{ruler} ($\mycirc[myred]$), \texttt{white hair/lines} ($\mycirc[mygreen]$), and \texttt{band-aid} ($\mycirc[myorange]$) artifacts.
    }
    \label{app:fig:reveal:isic_crp_1}
\end{figure*}

To reveal real-world artifacts in ISIC2019, we apply shortcut identification approaches using relevance scores after the $3^\text{rd}$ residual block of the ResNet50 model using samples from classes ``melanoma'', ``melanocytic nevus'', and ``benign keratosis'', respectively. 
The application of \gls{spray} reveals the usage of the \texttt{ruler} artifact for the prediction of ``melanoma'' (see Fig.~\ref{app:fig:reveal:isic_spray_0_1}, \emph{left}), the \texttt{band-aid} artifact for the prediction of the benign class ``melanocytic nevus'' (see Fig.~\ref{app:fig:reveal:isic_spray_0_1}, \emph{right}), as well as the usage of the artifacts \texttt{reflection} ($\mycirc[myorange]$) and \texttt{skin marker} ($\mycirc[mygreen]$) for class ``benign keratosis'' (Fig.~\ref{app:fig:reveal:isic_spray_4}).
From the model perspective, we can reveal the usage of outlier neurons focusing on the concepts \texttt{blueish tint} ($\mycirc[mygreen]$) and \texttt{ruler} ($\mycirc[myorange]$) for class ``melanoma'', as shown in Fig.~\ref{app:fig:reveal:isic_crp_0_4} (\emph{left}).
For the class ``benign keratosis'', the model uses outlier neurons focusing on the concepts \texttt{reflection} ($\mycirc[mygreen]$) and \texttt{ruler}($\mycirc[myorange]$), as shown in Fig.~\ref{app:fig:reveal:isic_crp_0_4} (\emph{right}).
Lastly, the concept perspetive reveals neurons for several confounders for the class ``melanocytic nevus'', including \texttt{band-aids} ($\mycirc[myorange]$), \texttt{rulers} ($\mycirc[myred]$), and \texttt{white hair/lines} ($\mycirc[mygreen]$), as shown in Fig.~\ref{app:fig:reveal:isic_crp_1}.
Interestingly, the application of \gls{pcx} on samples from class ``melanocytic nevus'' using relevances after the $4^\text{th}$ residual block in Fig.~\ref{app:fig:reveal:pcx_isic_attacked_isic1} (\emph{right}) reveals \texttt{prototype 0}, which is a sub-strategy relying on the concept \texttt{red skin} and related neurons \texttt{$\#$1147} and \texttt{$\#$1737}.

\begin{figure*}[t!]
    \centering
    \includegraphics[width=.35\textwidth]{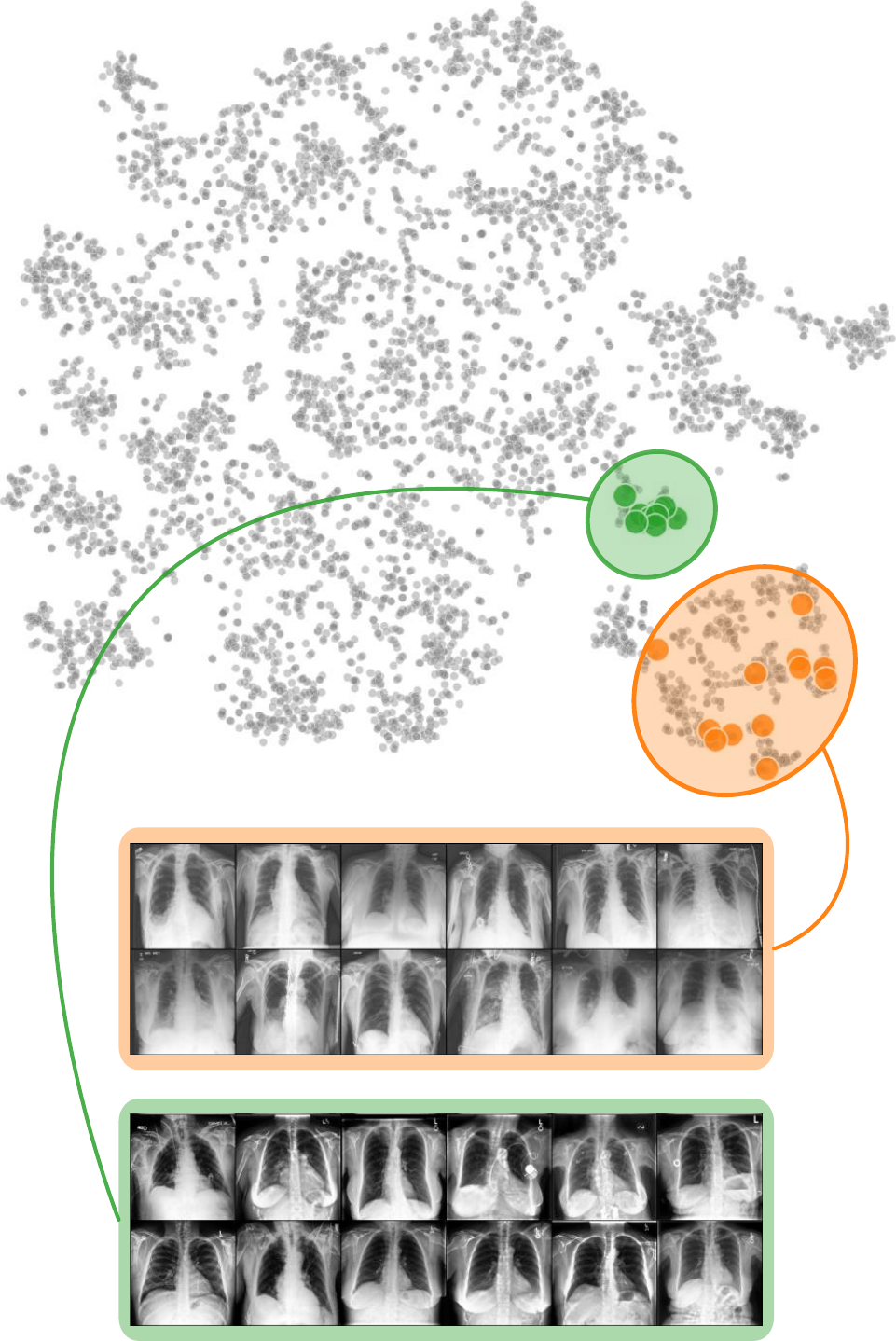}
    \quad\quad\quad
    \includegraphics[width=.35\textwidth]{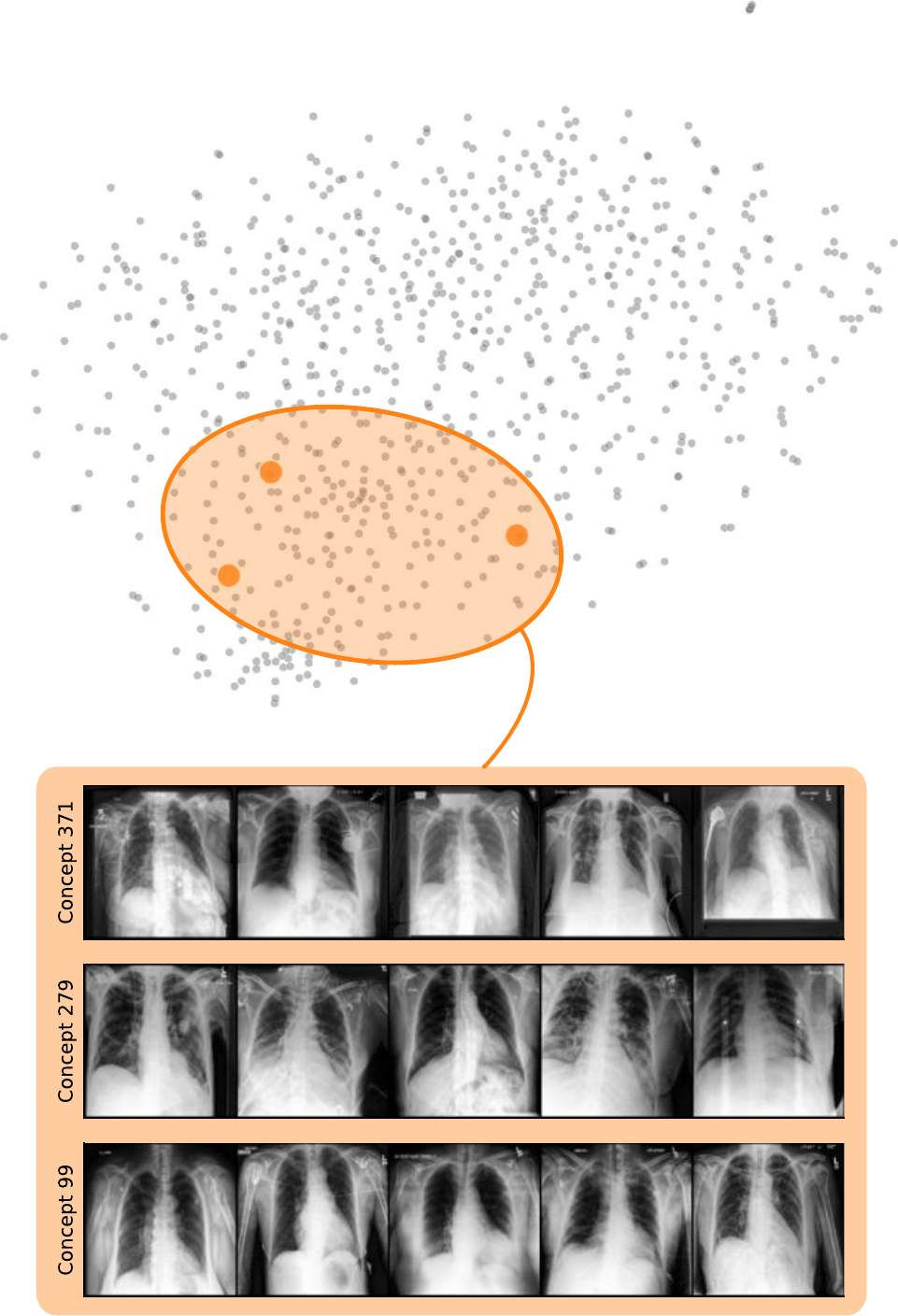}
    \caption{
    Artifact identification for a \gls{vit} model trained on CheXpert with the \texttt{brightness} artifact with samples from the ``cardiomegaly''-class using relevances of the \texttt{class}-token on the last layer.
    We apply \gls{spray} on latent relevances for the data perspective (\emph{left}) and pair-wise cosine similarities for the model perspective (\emph{right}). 
    Whereas the former reveals a clear cluster of samples with the \texttt{brightness} artifact, the latter does not lead to outlier concepts focusing ont he artifact. 
    Note that we highlight concepts for neurons $\#\texttt{371}$, $\#\texttt{279}$ and $\#\texttt{99}$, as they are revealed as relevant for impacted samples via \gls{pcx} in Fig.~\ref{app:fig:reveal:pcx_vit_chexpert_kvasir}.
    }
    \label{app:fig:reveal:vit_spray_crp}
\end{figure*}

\begin{figure*}[t!]
    \centering
    \includegraphics[width=.45\textwidth]{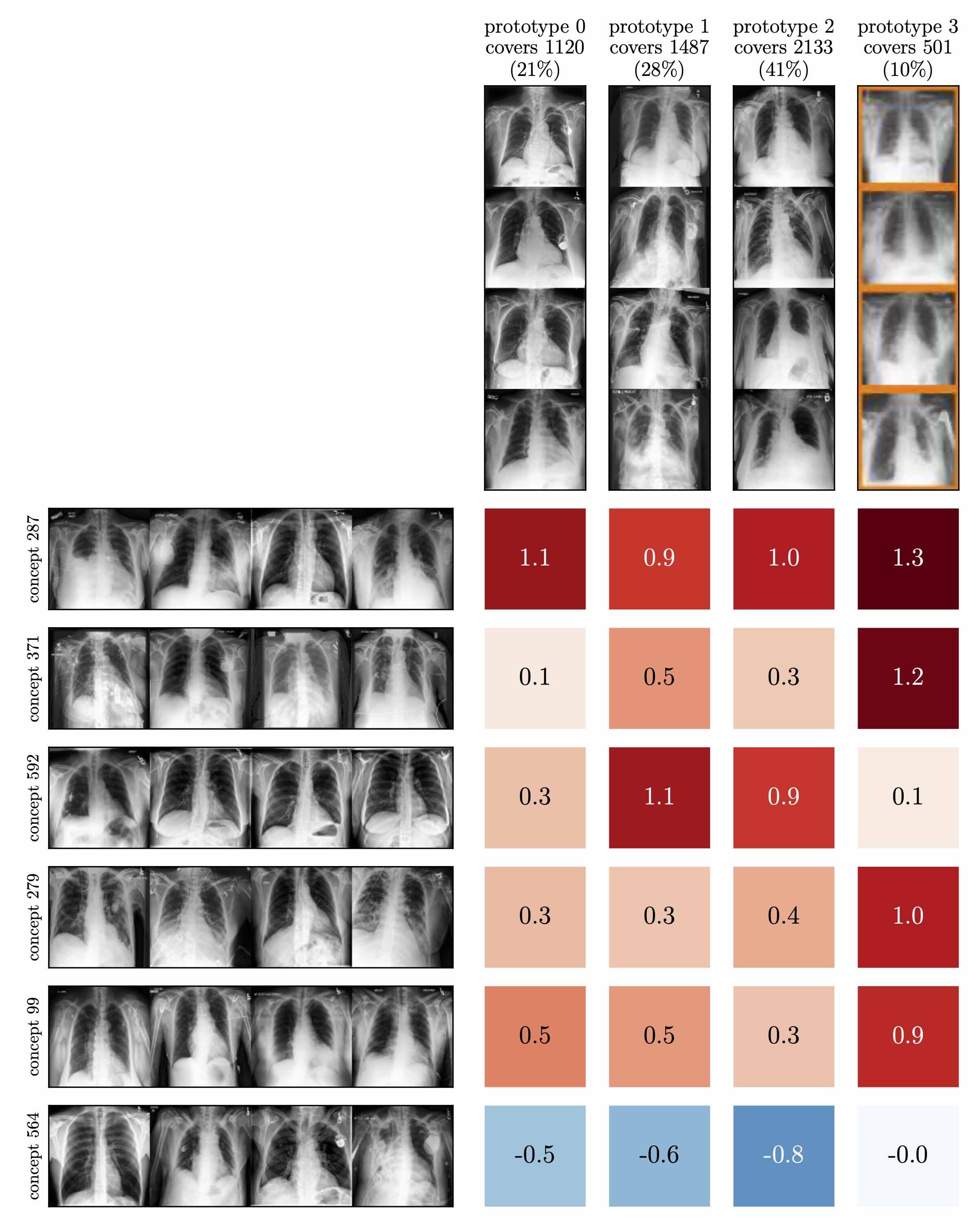}
    \quad
    \includegraphics[width=.5\textwidth]{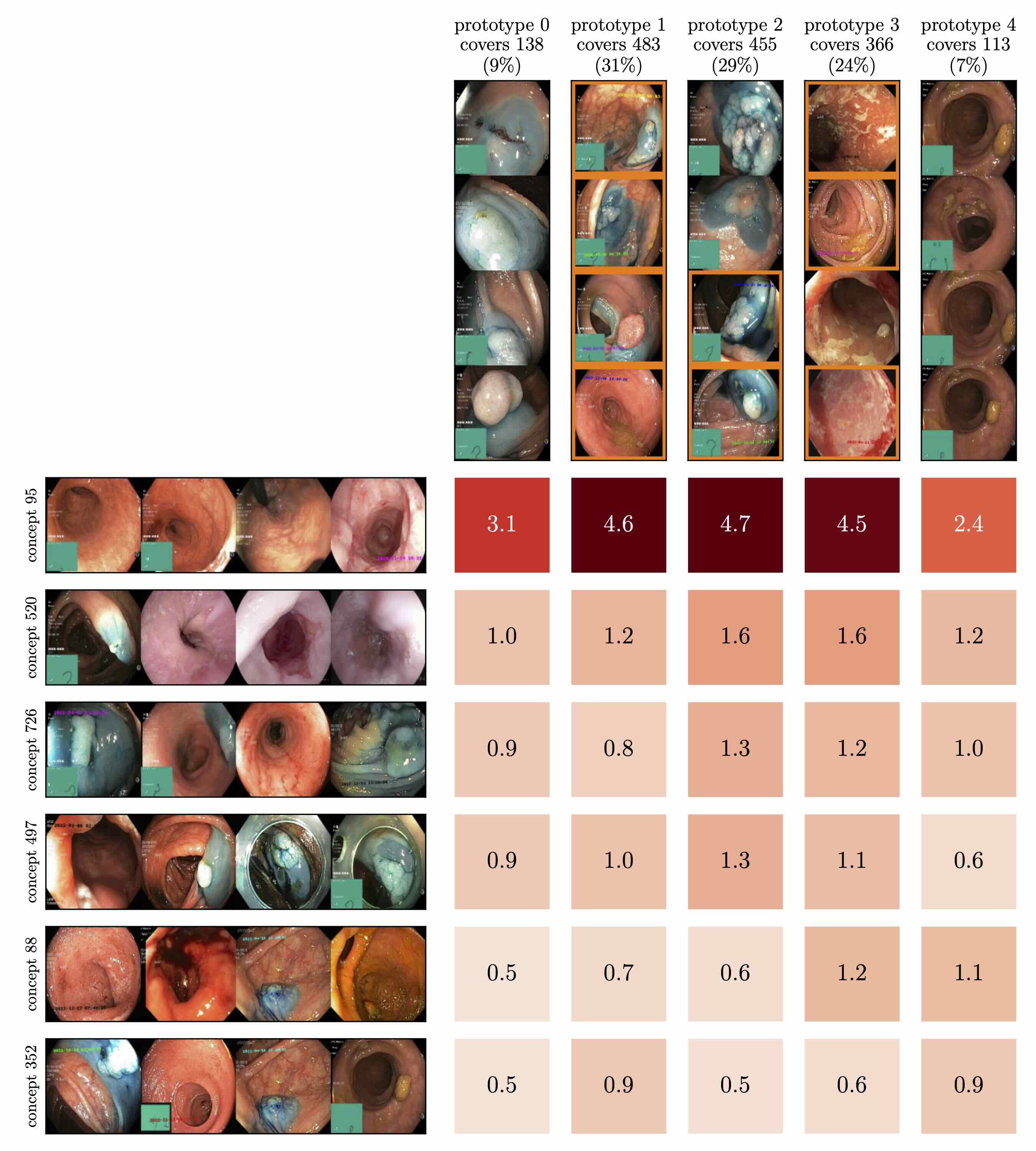}
    \caption{
    \gls{pcx} visualizations using latent relevances for the \texttt{class}-token on the last later of \gls{vit} trained on CheXpert with the controlled \texttt{brightness} artifact (\emph{left}, and on HyperKvasir with the controlled \texttt{timestamp} artifact (\emph{right}).
    Columns represent prototypes with four representative samples, rows represent concepts (visualized via RelMax) and the values in the matrix indicate the average relevance of the concept for the prototype. 
    For the CheXpert(\emph{left}), \texttt{prototype 3} focuses on the \texttt{brightness} artifact, with high scores for the related concepts (neurons) $\#\texttt{371}$, $\#\texttt{279}$ and $\#\texttt{99}$.
    For HyperKvasir (\emph{right}), no prototype primarily focuses on the artifact and clinically relevant features dominate.
    }
    \label{app:fig:reveal:pcx_vit_chexpert_kvasir}
\end{figure*}

\paragraph{Vision Data (ViT)}
The interpretation of neurons on the last layer of \glspl{vit} poses a particular challenge, as they are not preceded by a ReLU non-linearity. 
As such, both positive and negative activations are possible and the amplitude of the activations cannot be interpreted as the degree of existence of a certain concept. 
However, we apply bias identification approaches from the data perspective (\gls{spray}), model perspective (pairwise cosine similarities between neurons) and the combination thereof (\gls{pcx})
on relevance scores for the \texttt{class}-token on the last layer.
Fig.~\ref{app:fig:reveal:vit_spray_crp} shows that the data perspective (\emph{left}) can clearly reveal samples with the \texttt{brightness} artifact in CheXpert, but the model perspective (\emph{right}) does not detect outlier concepts focussing on the artifact. 
We highlight concepts revealed via the application of \gls{pcx} in Fig.~\ref{app:fig:reveal:pcx_vit_chexpert_kvasir} (\emph{left}), leading to \texttt{prototype 3} with a clear focus on the artifact on high average relevance scores for concepts $\#\texttt{371}$, $\#\texttt{279}$ and $\#\texttt{99}$.
For HyperKvasir, \gls{pcx} does not reveal a prototype mainly focusing on the \texttt{brightness}, but instead prototypes that distinguish by clinically valid features (Fig.~\ref{app:fig:reveal:pcx_vit_chexpert_kvasir}, \emph{right}).

\subsubsection{Biased Sample Retrieval}
\label{app:sec:exp_data_annotation}
\begin{figure*}[t!]
    \centering
    \includegraphics[width=.36\textwidth]{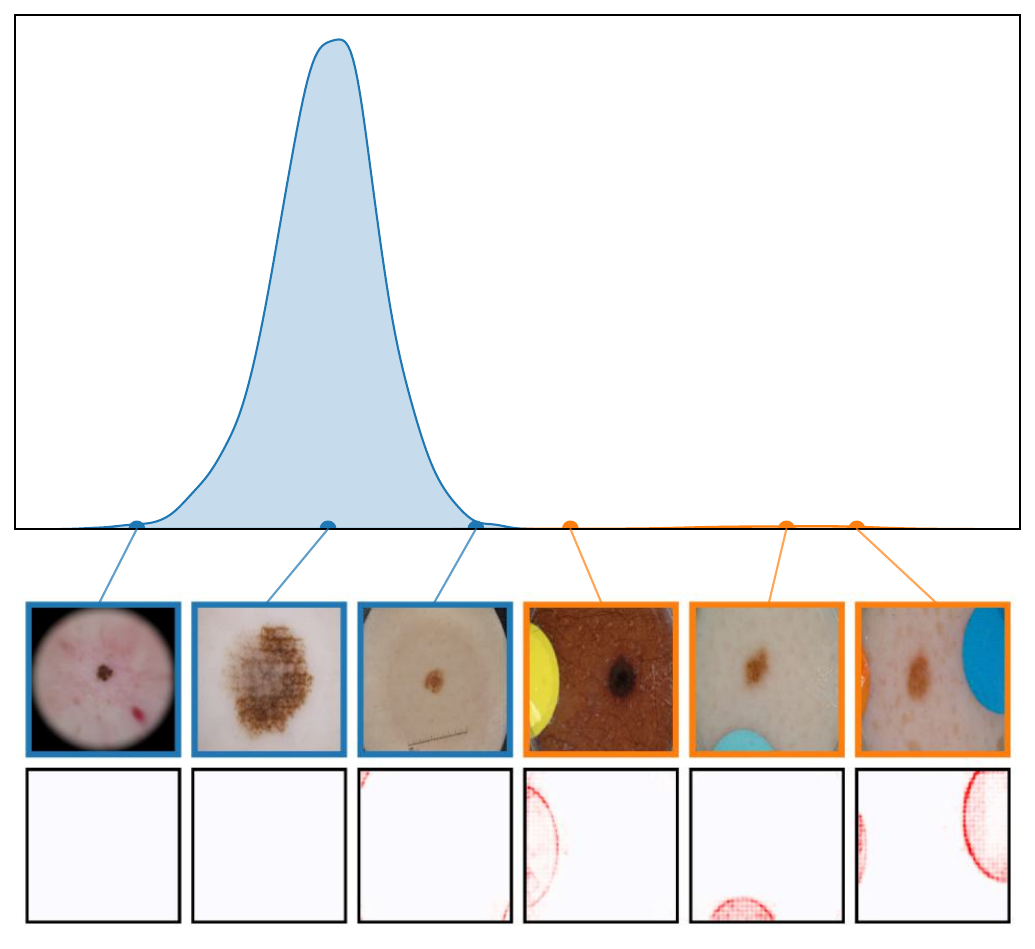}
    \quad\quad\quad\quad
    \includegraphics[width=.36\textwidth]{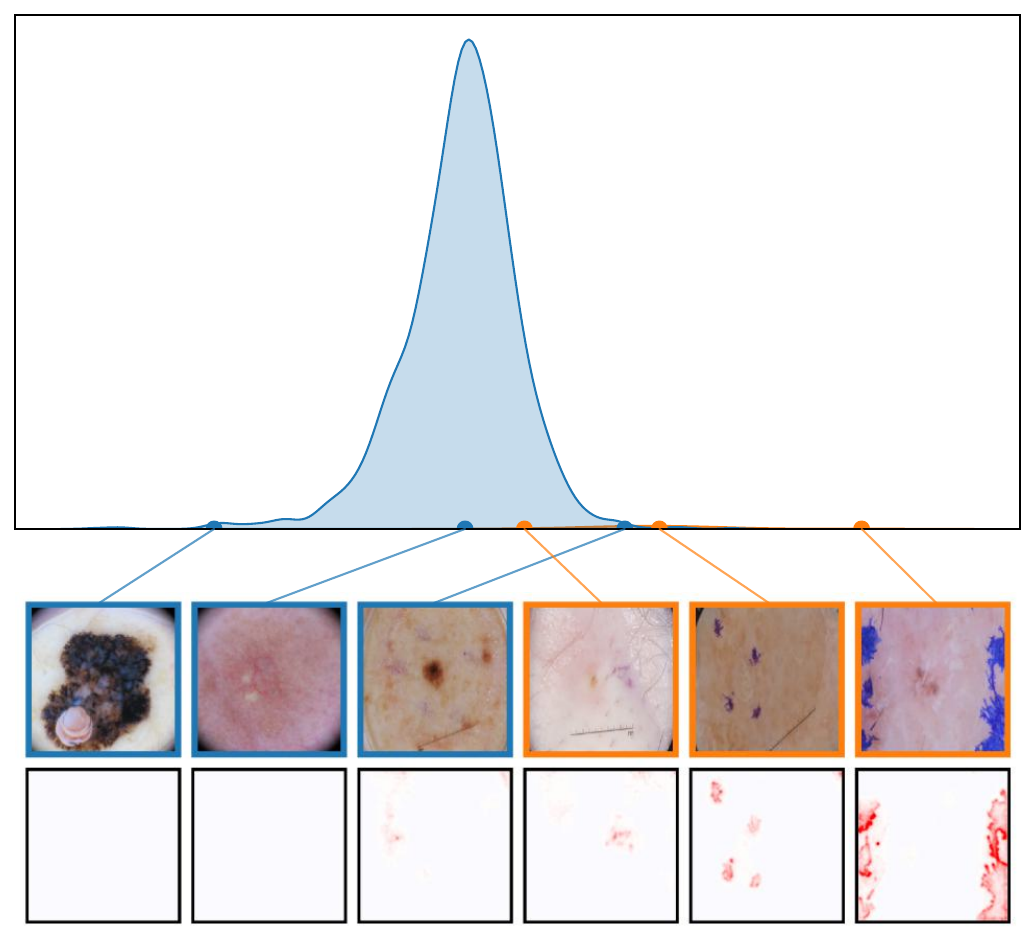}
    \caption{
    \emph{Top:} Distribution of bias scores for real artifacts \texttt{band-aid} (\emph{left}) and \texttt{skin-marker} (\emph{right}) in the ISIC2019, split into unlabeled samples (\emph{blue}) and samples labeled (\emph{orange}) as artifact.
    \emph{Bottom:} Samples at the 1-, 50-, and 99-percentile of each sample set and the artifact localization using the CAV. 
    For \texttt{skin-marker}, the unlabeled sample in the 99-percentile is a false negative, i.e., although barely visible, it contains ink, and has not yet been detected in the data annotation process.
    }
    \label{fig:data_annotation_real_bandaid_skinmarker}
\end{figure*}

\begin{figure*}[t!]
    \centering
    \includegraphics[width=.36\textwidth]{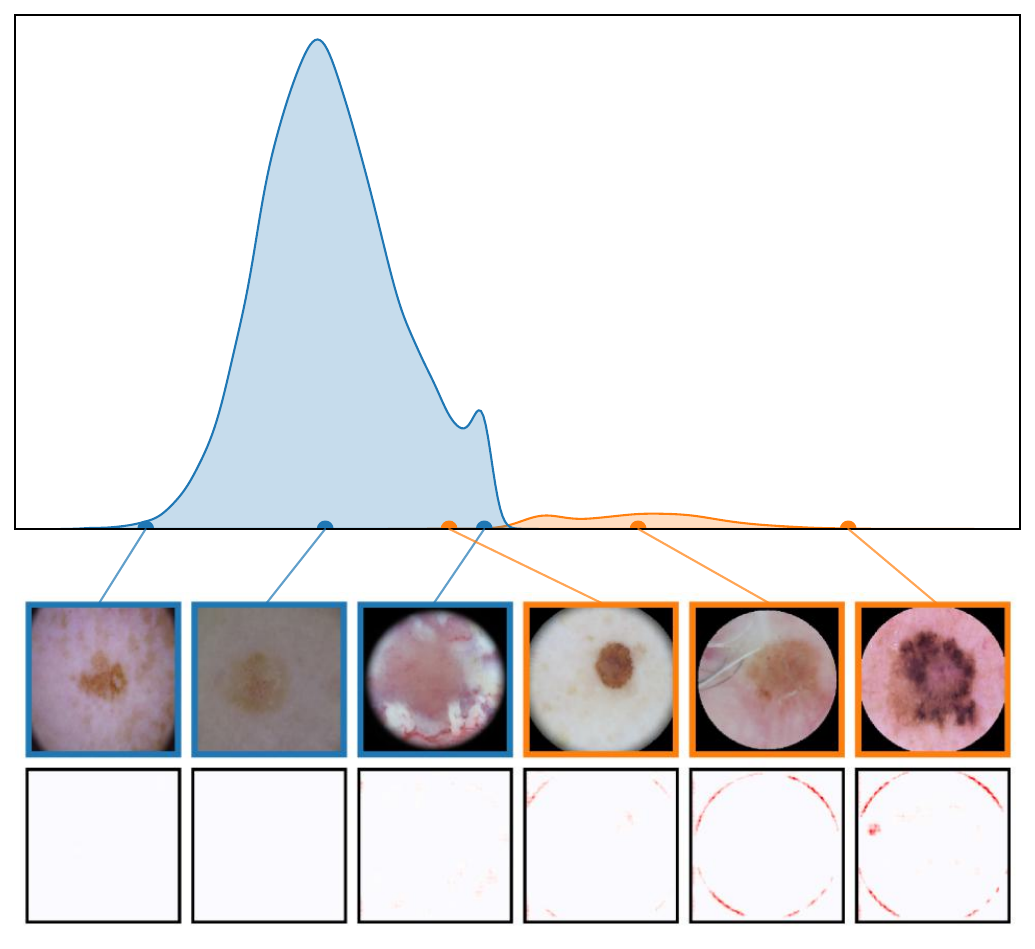}
    \quad\quad\quad\quad
    \includegraphics[width=.36\textwidth]{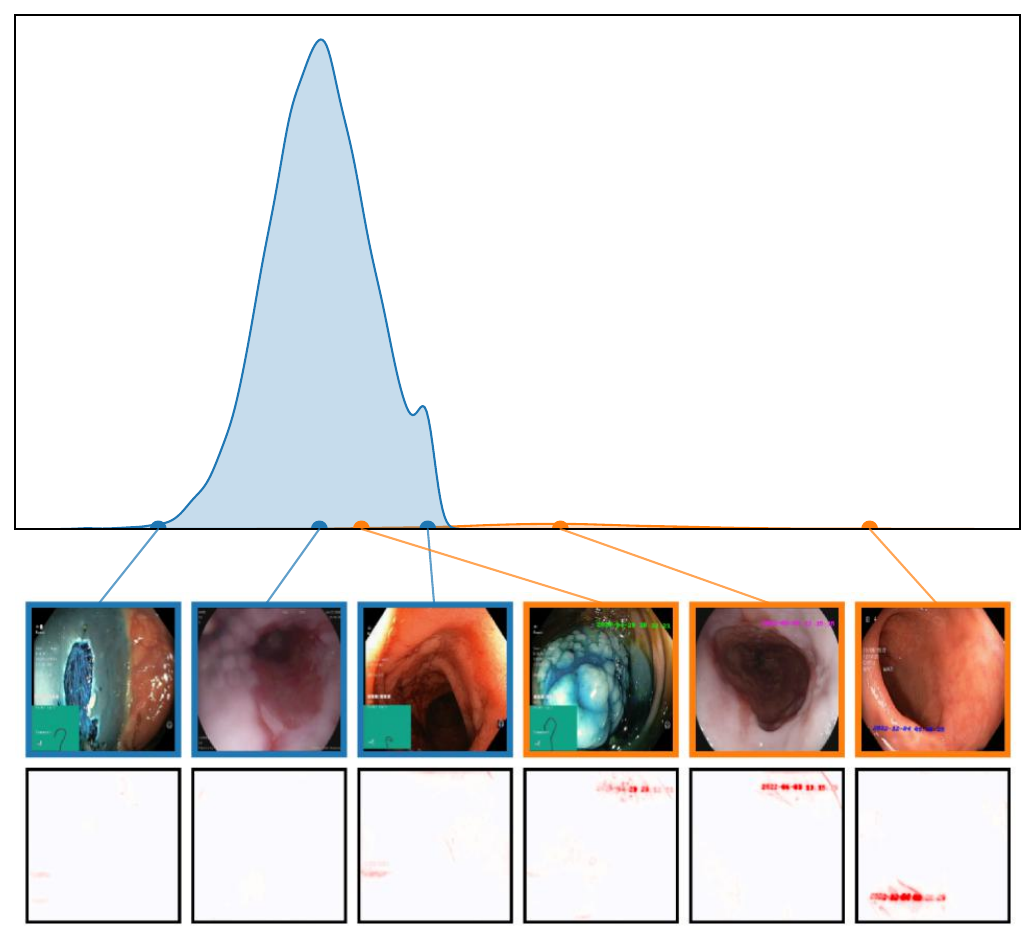}
    \caption{
    \emph{Top:} Distribution of bias scores for controlled artifacts \texttt{miscroscope} in ISIC2019 (\emph{left}) and \texttt{timestamp} in HyperKvasir (\emph{right}), split into clean (\emph{blue}) and manipulated (\emph{orange}) samples.
    \emph{Bottom:} Samples at the 1-, 50-, and 99-percentile of each sample set and the artifact localization using the CAV. 
    The samples in the 99-percentile of the clean sets contain natural artifacts similar to the artificially inserted ones.
    }
    \label{fig:data_annotation_controlled_isic_kvasir}
\end{figure*}

\begin{figure*}[t!]
    \centering
    \includegraphics[width=.36\textwidth]{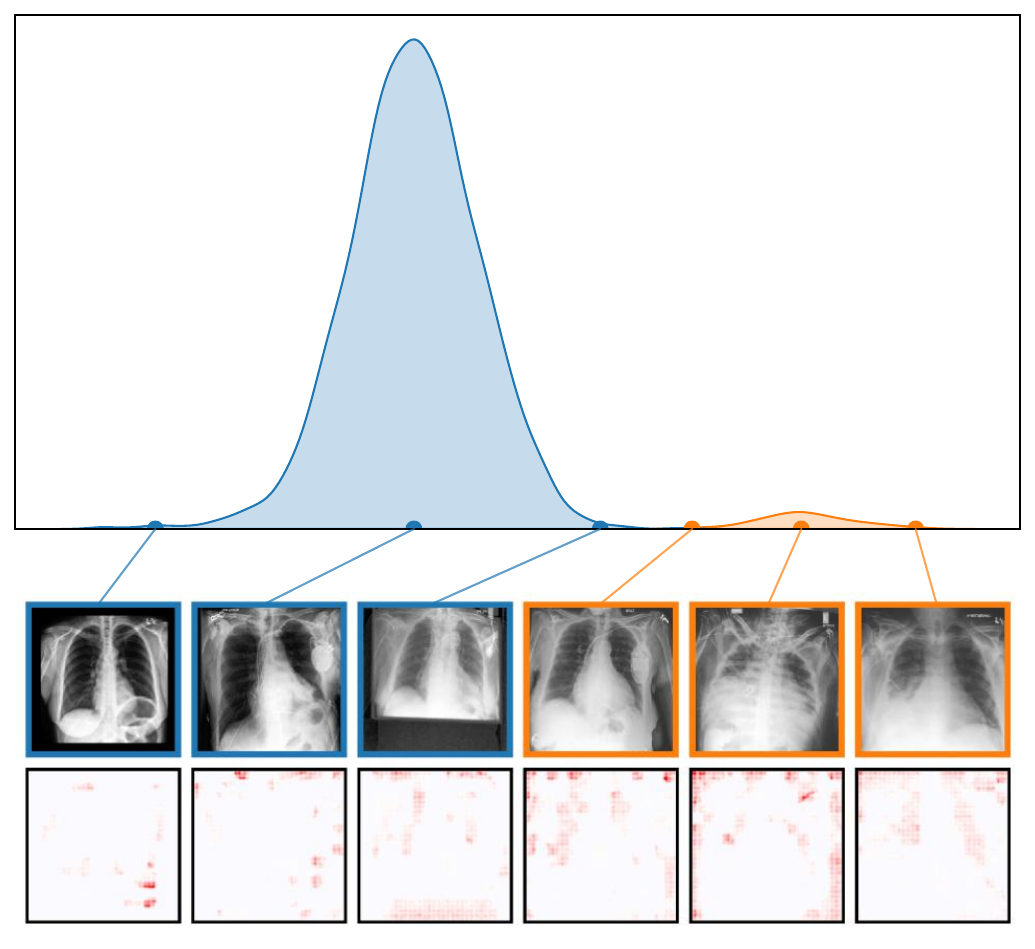}
    \quad\quad\quad\quad
    \includegraphics[width=.36\textwidth]{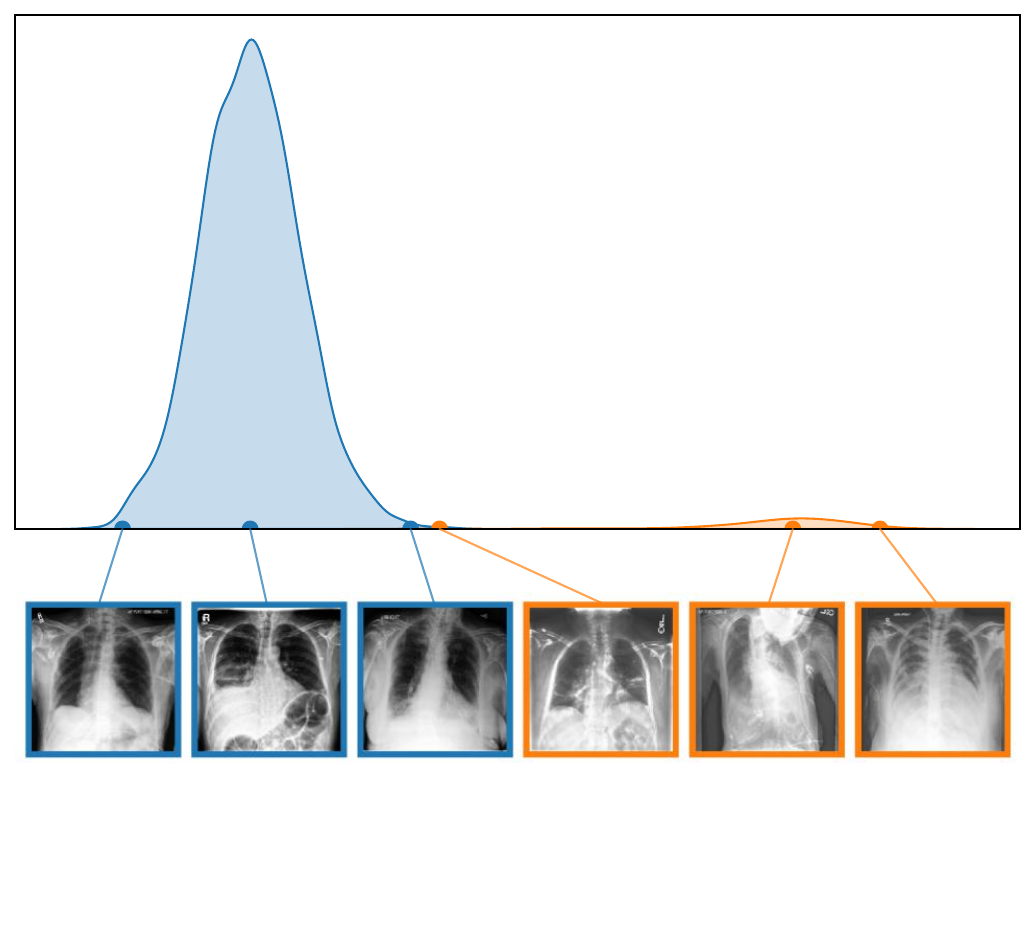}
    \caption{
    \emph{Top:} Distribution of bias scores for the controlled \texttt{brightness} artifact in CheXpert for ResNet50 (\emph{left}) and \gls{vit} (\emph{right}), split into clean (\emph{blue}) and manipulated (\emph{orange}) samples.
    \emph{Bottom:} Samples at the 1-, 50-, and 99-percentile of each sample set and the artifact localization using the CAV.
    }
    \label{fig:data_annotation_controlled_chexpert}
\end{figure*}

In addition to our experiments in Sec.~\ref{sec:exp_data_annotation}, we report quantitative biased sample retrieval results via \gls{auc} and \gls{ap} for the real-world artifacts \texttt{band-aid}, \texttt{ruler} (both ISIC2019) and \texttt{pacemaker} (CheXpert), as well as for the controlled artifacts \texttt{microscope} (ISIC2019), \texttt{timestamp} (HyperKvasir), and \texttt{brightness} (CheXpert) in Tabs.~\ref{tab:data_annotation_cav}~and~\ref{tab:data_annotation_single_neuron} for CAV-based and individual-neuron-based bias scores, respectively.
For the latter, the best performing neuron is selected on the validation set and results are reported for an unseen test set.
We further show the distribution of CAV-based bias scores for clean and biased samples for the artifacts \texttt{band-aid} and \texttt{skin marker} (both ISIC2019) in Fig.~\ref{fig:data_annotation_real_bandaid_skinmarker}, for the controlled artifacts \texttt{microscope} (ISIC2019) and \texttt{timestamp} (HyperKvasir) in Fig.~\ref{fig:data_annotation_controlled_isic_kvasir}, and lastly for \texttt{brightness} (CheXpert) in Fig.~\ref{fig:data_annotation_controlled_chexpert}.
The following models and layers are used for the distribution plots: last Conv layer of ResNet50 for \texttt{band-aid}, $11^{th}$ Conv layer of VGG16 for \texttt{skin marker} in ISIC2019, $3^\text{rd}$ residual block of ResNet50 for \texttt{microscope} for the controlled ISIC2019 dataset, $10^{th}$ Conv layer of VGG16 for \texttt{timestamp} in the controlled HyperKvasir dataset, as well as the $3^\text{rd}$ residual block of ResNet50 and the last linear layer of \gls{vit} for the controlled \texttt{brightness} artifact in CheXpert.

\begin{table*}[t]\centering
    \caption{
    Quantitative results for CAV-based biased sample retrieval \wrt the real-world artifacts \texttt{band-aid}, \texttt{ruler} (both ISIC2019), and \texttt{pacemaker} (CheXpert), as well as the controlled artifacts \texttt{microscope} (ISIC2019), \texttt{timestamp} (HyperKvasir), and \texttt{brightness} (CheXpert). 
    We report \gls{auc} and \gls{ap} to evaluate the ranking capabilities of CAV-based bias scores using activations on different layers of VGG16, ResNet50, and ViT models.
    Higher scores are better. 
    }
        \begin{tabular}{
    l@{\hspace{0.75em}}|
    c@{\hspace{1em}}|
    *{3}{c@{\hspace{0.25em}}c@{\hspace{0.25em}}|}
    c@{\hspace{0.25em}}c@{\hspace{0.25em}}|
    *{2}{c@{\hspace{0.25em}}}|
    *{2}{c@{\hspace{0.25em}}}
    }
        \toprule
         &  & \multicolumn{6}{c|}{ISIC2019} & \multicolumn{2}{c|}{HypKvasir} & \multicolumn{4}{c}{CheXpert} \\ 
        & & \multicolumn{2}{c|}{\texttt{band-aid}} & \multicolumn{2}{c|}{\texttt{ruler}} & \multicolumn{2}{c|}{\texttt{microscope}} & \multicolumn{2}{c|}{\texttt{timestamp}} & \multicolumn{2}{c|}{\texttt{pacemaker}} & \multicolumn{2}{c}{\texttt{brightness}} \\ 
        & layer & AUC & AP & AUC & AP & AUC & AP & AUC & AP & AUC & AP & AUC & AP \\ 
        \midrule
        \multirow{4}{*}{\rotatebox[origin=c]{90}{VGG16}} &     Conv 4 &                        1.0 &                      0.99 &                    0.99 &                   0.85 &                                             0.98 &                                            0.89 &                                    0.97 &                                   0.85 &                          0.70 &                       0.16 &                                          1.0 &                                        0.97 \\
           &     Conv 7 &                        1.0 &                      1.00 &                    1.00 &                   0.92 &                                             0.97 &                                            0.86 &                                    0.93 &                                   0.63 &                          0.91 &                       0.62 &                                          1.0 &                                        0.98 \\
           &    Conv 10 &                        1.0 &                      1.00 &                    1.00 &                   0.93 &                                             0.98 &                                            0.86 &                                    0.98 &                                   0.74 &                          0.98 &                       0.87 &                                          1.0 &                                        1.00 \\
           &    Conv 13 &                        1.0 &                      0.98 &                    1.00 &                   0.88 &                                             0.96 &                                            0.84 &                                    0.98 &                                   0.43 &                          0.96 &                       0.82 &                                          1.0 &                                        1.00 \\
        
         \midrule
        \multirow{4}{*}{\rotatebox[origin=c]{90}{ResNet50}} &    Block 1 &                        1.0 &                       1.0 &                     1.0 &                   0.87 &                                             0.99 &                                            0.92 &                                    0.98 &                                   0.87 &                          0.76 &                       0.26 &                                          1.0 &                                         1.0 \\
           &    Block 2 &                        1.0 &                       1.0 &                     1.0 &                   0.90 &                                             0.99 &                                            0.91 &                                    0.99 &                                   0.87 &                          0.95 &                       0.75 &                                          1.0 &                                         1.0 \\
           &    Block 3 &                        1.0 &                       1.0 &                     1.0 &                   0.92 &                                             0.98 &                                            0.86 &                                    1.00 &                                   0.96 &                          0.97 &                       0.83 &                                          1.0 &                                         1.0 \\
           &    Block 4 &                        1.0 &                       1.0 &                     1.0 &                   0.88 &                                             0.99 &                                            0.91 &                                    1.00 &                                   0.94 &                          0.95 &                       0.77 &                                          1.0 &                                         1.0 \\
             \midrule
           \rotatebox[origin=c]{90}{ViT} & Last layer &                       0.97 &                      0.96 &                     1.0 &                   0.88 &                                             0.97 &                                            0.83 &                                    0.99 &                                    0.9 &                          0.86 &                       0.47 &                                          1.0 &                                         1.0 \\
        \bottomrule
    \end{tabular}
    \label{tab:data_annotation_cav}
    \end{table*}
\begin{table*}[t]\centering
    \caption{
    Quantitative results for neuron-based biased sample retrieval \wrt the real-world artifacts \texttt{band-aid}, \texttt{ruler} (both ISIC2019), and \texttt{pacemaker} (CheXpert), as well as the controlled artifacts \texttt{microscope} (ISIC2019), \texttt{timestamp} (HyperKvasir), and \texttt{brightness} (CheXpert). 
    We report \gls{auc} and \gls{ap} to evaluate the ranking capabilities of individual-neuron-based bias scores using activations on different layers of VGG16, ResNet50, and ViT models.
    Out of all neurons, the best performing neurons are selected on the validation set and results are reported for the unseen test set.
    Higher scores are better. 
    }
        \begin{tabular}{
    l@{\hspace{0.75em}}|
    c@{\hspace{1em}}|
    *{3}{c@{\hspace{0.25em}}c@{\hspace{0.25em}}|}
    c@{\hspace{0.25em}}c@{\hspace{0.25em}}|
    *{2}{c@{\hspace{0.25em}}}|
    *{2}{c@{\hspace{0.25em}}}
    }
        \toprule
         &  & \multicolumn{6}{c|}{ISIC2019} & \multicolumn{2}{c|}{HypKvasir} & \multicolumn{4}{c}{CheXpert} \\ 
        & & \multicolumn{2}{c|}{\texttt{band-aid}} & \multicolumn{2}{c|}{\texttt{ruler}} & \multicolumn{2}{c|}{\texttt{microscope}} & \multicolumn{2}{c|}{\texttt{timestamp}} & \multicolumn{2}{c|}{\texttt{pacemaker}} & \multicolumn{2}{c}{\texttt{brightness}} \\ 
        & layer & AUC & AP & AUC & AP & AUC & AP & AUC & AP & AUC & AP & AUC & AP \\ 
        \midrule
\multirow{4}{*}{\rotatebox[origin=c]{90}{VGG16}} &     Conv 4 &                          0.98 &                         0.44 &                       0.91 &                      0.31 &                                                0.98 &                                               0.86 &                                       0.84 &                                      0.29 &                           0.64 &                          0.08 &                                            0.56 &                                           0.04 \\
           &     Conv 7 &                          0.96 &                         0.31 &                       0.95 &                      0.40 &                                                0.97 &                                               0.83 &                                       0.89 &                                      0.63 &                           0.73 &                          0.25 &                                            0.68 &                                           0.04 \\
           &    Conv 10 &                          0.94 &                         0.29 &                       0.97 &                      0.68 &                                                0.98 &                                               0.88 &                                       0.94 &                                      0.52 &                           0.91 &                          0.52 &                                            0.82 &                                           0.12 \\
           &    Conv 13 &                          0.97 &                         0.32 &                       0.98 &                      0.56 &                                                0.96 &                                               0.78 &                                       0.97 &                                      0.83 &                           0.79 &                          0.41 &                                            0.97 &                                           0.47 \\
        
         \midrule
        \multirow{4}{*}{\rotatebox[origin=c]{90}{ResNet50}} &    Block 1 &                          0.99 &                         0.85 &                       0.90 &                      0.24 &                           0.60 &                          0.17 &                                                0.98 &                                               0.82 &                                            0.80 &                                           0.13 &                                       0.90 &                                      0.73 \\
           &    Block 2 &                          1.00 &                         0.89 &                       0.94 &                      0.52 &                           0.77 &                          0.29 &                                                0.98 &                                               0.85 &                                            0.88 &                                           0.22 &                                       0.89 &                                      0.59 \\
           &    Block 3 &                          1.00 &                         0.90 &                       0.99 &                      0.70 &                           0.76 &                          0.26 &                                                0.97 &                                               0.83 &                                            0.94 &                                           0.20 &                                       1.00 &                                      0.92 \\
           &    Block 4 &                          1.00 &                         0.86 &                       0.99 &                      0.65 &                           0.87 &                          0.54 &                                                0.98 &                                               0.88 &                                            0.99 &                                           0.95 &                                       0.96 &                                      0.90 \\
           \midrule
\rotatebox[origin=c]{90}{ViT} & Last layer &                          0.73 &                         0.08 &                       0.88 &                      0.24 &                                                 0.9 &                                                0.5 &                                       0.95 &                                      0.79 &                           0.67 &                          0.15 &                                            0.96 &                                           0.77 \\
        \bottomrule
    \end{tabular}
    \label{tab:data_annotation_single_neuron}
    \end{table*}

\subsubsection{Spatial Bias Localization}
\label{app:sec:exp_bias_localization}

\begin{figure*}[t!]
    \centering
        \includegraphics[width=.4\textwidth]{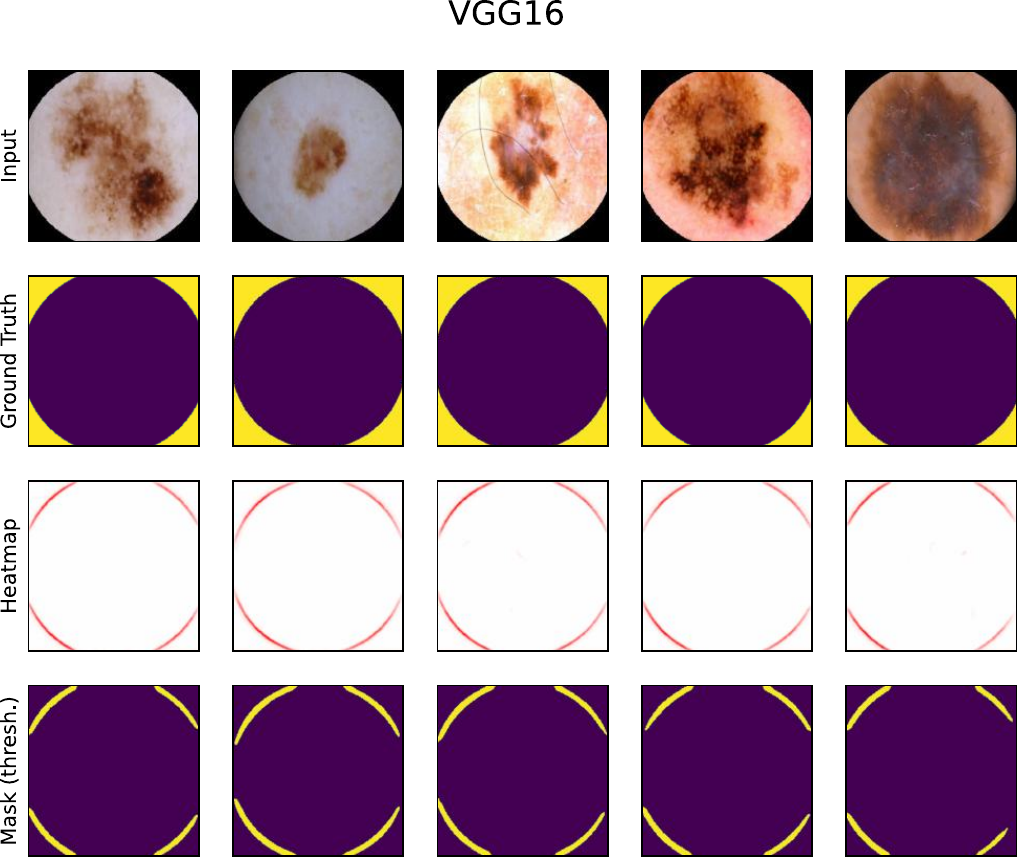}
        \quad\quad
        \includegraphics[width=.4\textwidth]{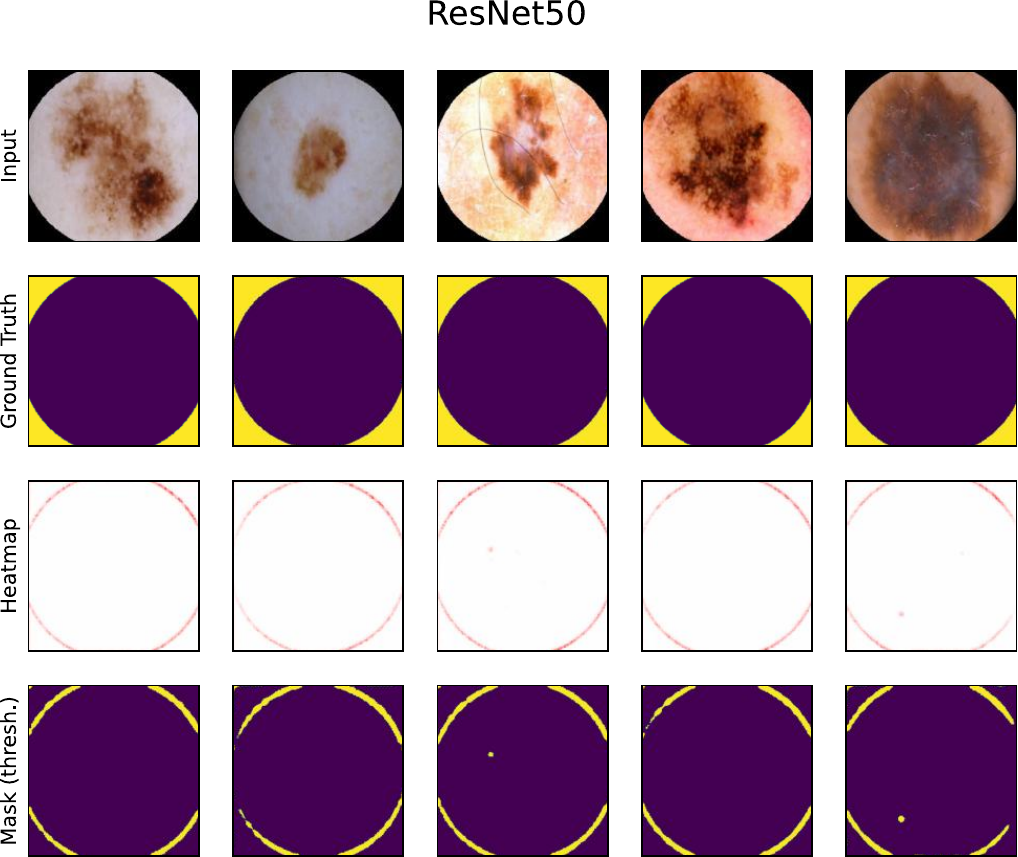}
    \caption{Examples for bias localization for the controlled \texttt{microscope} artifact in ISIC2019 using VGG16 (\emph{left}) and ResNet50 (\emph{right}) model architectures. We show the input samples (\emph{$1^\text{st}$ row}), ground truth masks (\emph{$2^\text{nd}$ row}), computed concept heatmaps using CAVs (\emph{$3^\text{rd}$ row}), and binarized masks using Otsu's method (\emph{$4^\text{th}$ row}).
    Both models focus only on the border of the circle, instead of the entire area.
    }
    \label{app:fig:exp_bias_localization_isic_microscope}
\end{figure*}

\begin{figure*}[t!]
    \centering
    \includegraphics[width=.4\textwidth]{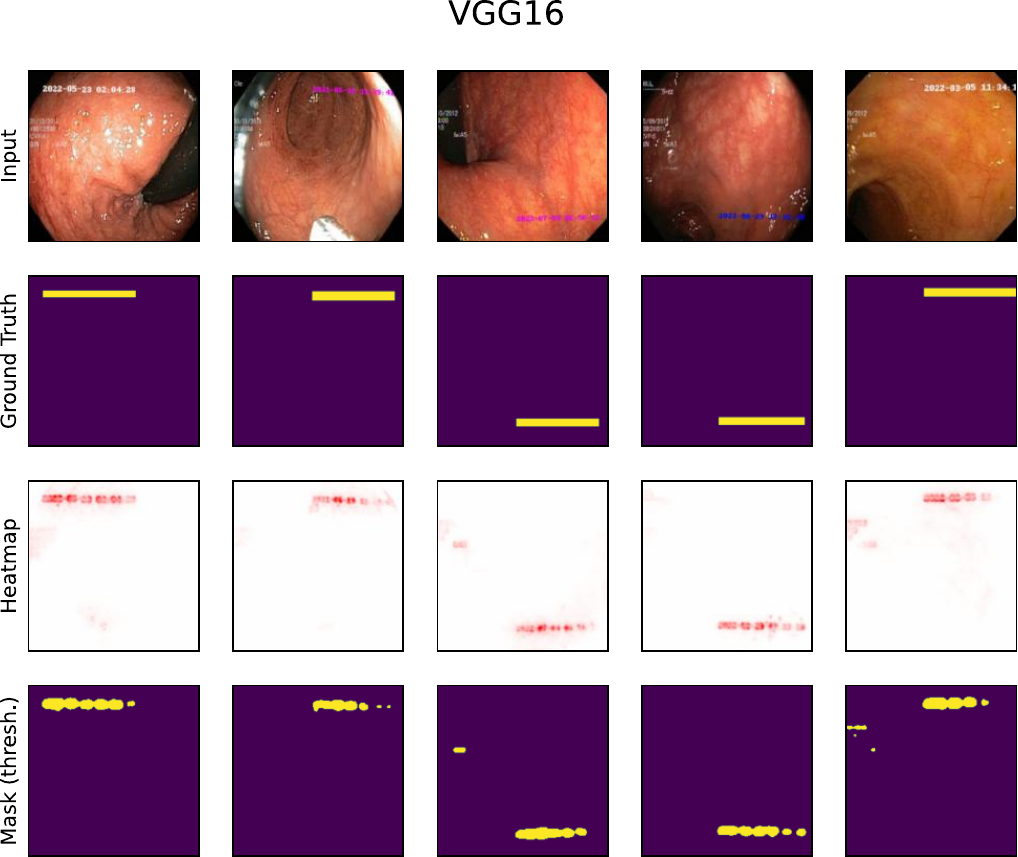}
    \quad\quad
    \includegraphics[width=.4\textwidth]{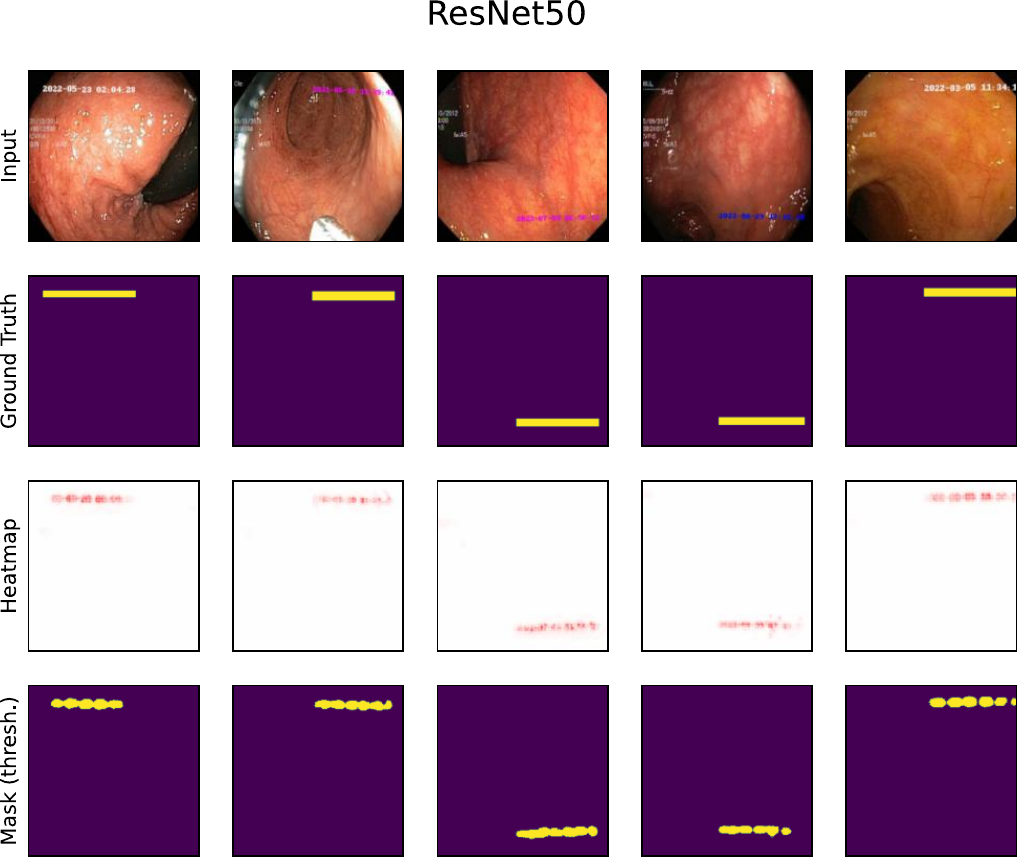}
    \caption{
    Examples for bias localization for the controlled \texttt{timestamp} artifact in HyperKvasir using VGG16 (\emph{left}) and ResNet50 (\emph{right}) model architectures. We show the input samples (\emph{$1^\text{st}$ row}), ground truth masks (\emph{$2^\text{nd}$ row}), computed concept heatmaps using CAVs (\emph{$3^\text{rd}$ row}), and binarized masks using Otsu's method (\emph{$4^\text{th}$ row}).
    }
    \label{app:fig:exp_bias_localization_kvasir_timestamp}
\end{figure*}

Complementing our experiments in Sec.~\ref{sec:exp_bias_localization}, we report our quantitative bias localization results, specifically the artifact relevance and \gls{iou}, for the controlled artifacts \texttt{microscope} (ISIC2019) and \texttt{timestamp} (HyperKvasir) in tabular form in Tab.~\ref{tab:bias_localization}.
Moreover, we show exemplary concept heatmaps and binary localization masks using CAVs on the best performing layer of VGG16 and ResNet50 in Figs.~\ref{app:fig:exp_bias_localization_isic_microscope}~and~\ref{app:fig:exp_bias_localization_kvasir_timestamp} for \texttt{microscope} and \texttt{timestamp}, respectively. 
For the \texttt{microscope} artifact, both models only detect the border of the circle instead of the entire area.
This leads to low \gls{iou} scores, as the metric also measures what the computed mask does \emph{not} cover.

\begin{figure*}[t!]
    \centering
    \includegraphics[width=.4\textwidth]{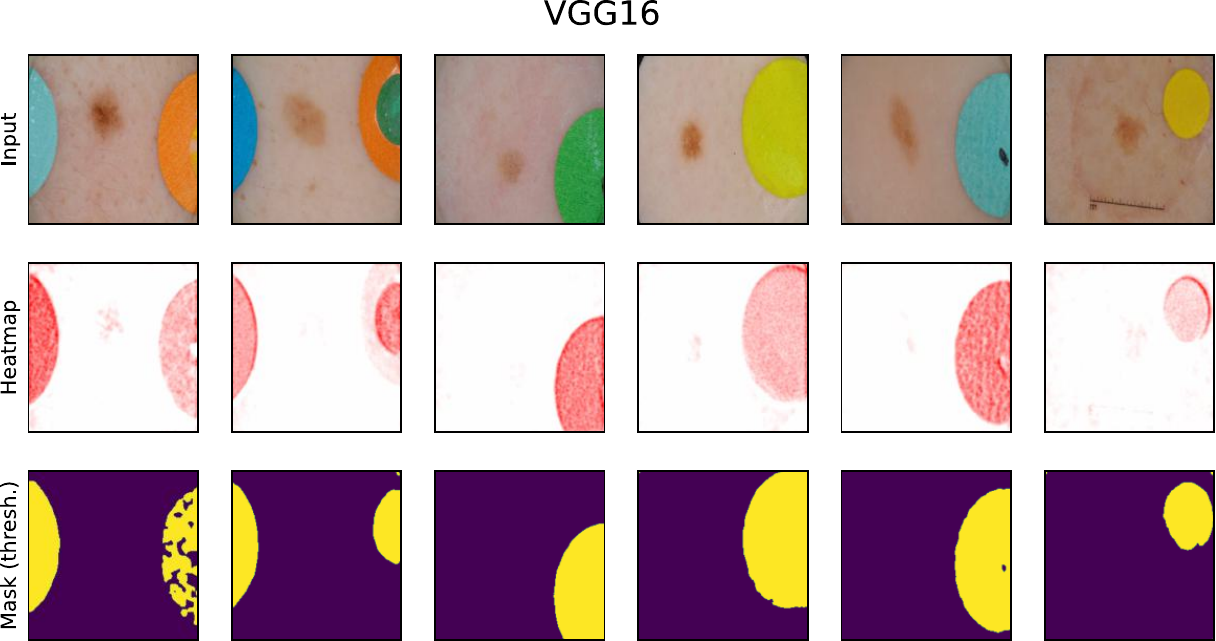}
    \quad\quad
    \includegraphics[width=.4\textwidth]{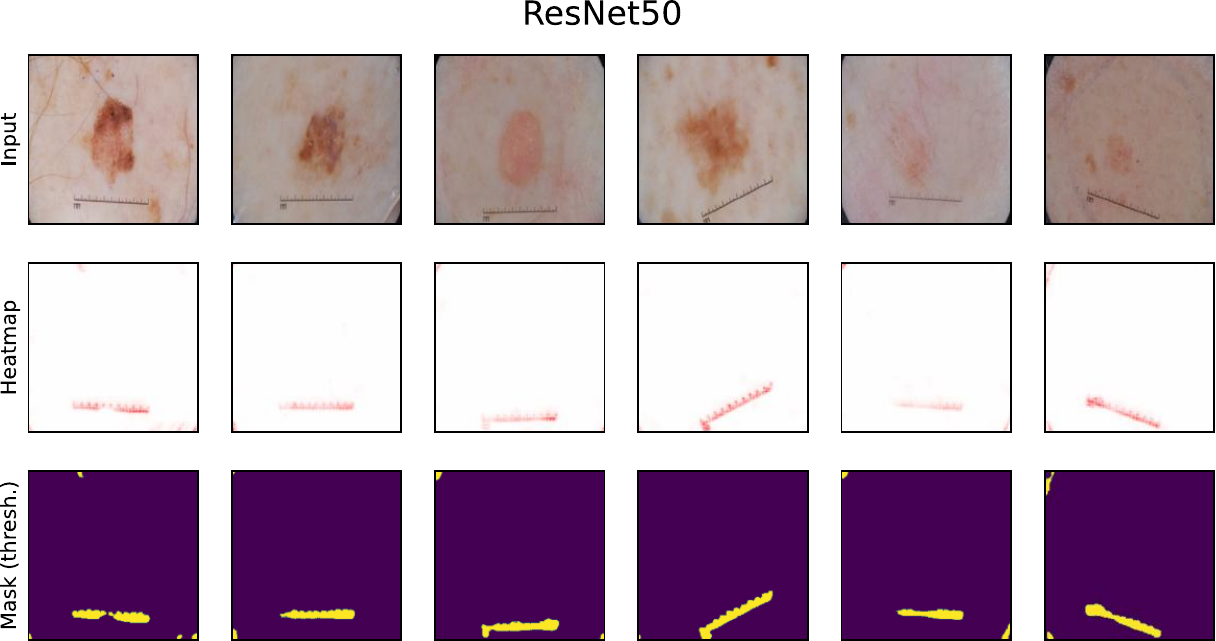}
    \caption{
    Examples for bias localization for the real-world artifacts \texttt{band-aid} (\emph{left}) and \texttt{ruler} (\emph{right}) in ISIC2019. 
    The artifact concept is modeled using CAVs with activations after the $3^\text{rd}$ conv layer of a VGG16 model for \texttt{band-aid} and after the $3^\text{rd}$ residual block of a ResNet50 model for \texttt{ruler}. 
    We show the input samples (\emph{$1^\text{st}$ row}), computed concept heatmaps (\emph{$2^\text{nd}$ row}), and binarized masks using Otsu's method (\emph{$3^\text{rd}$ row}).
    }
    \label{app:fig:exp_bias_localization_real_isic}
\end{figure*}

\begin{figure*}[t!]
    \centering
    \includegraphics[width=.4\textwidth]{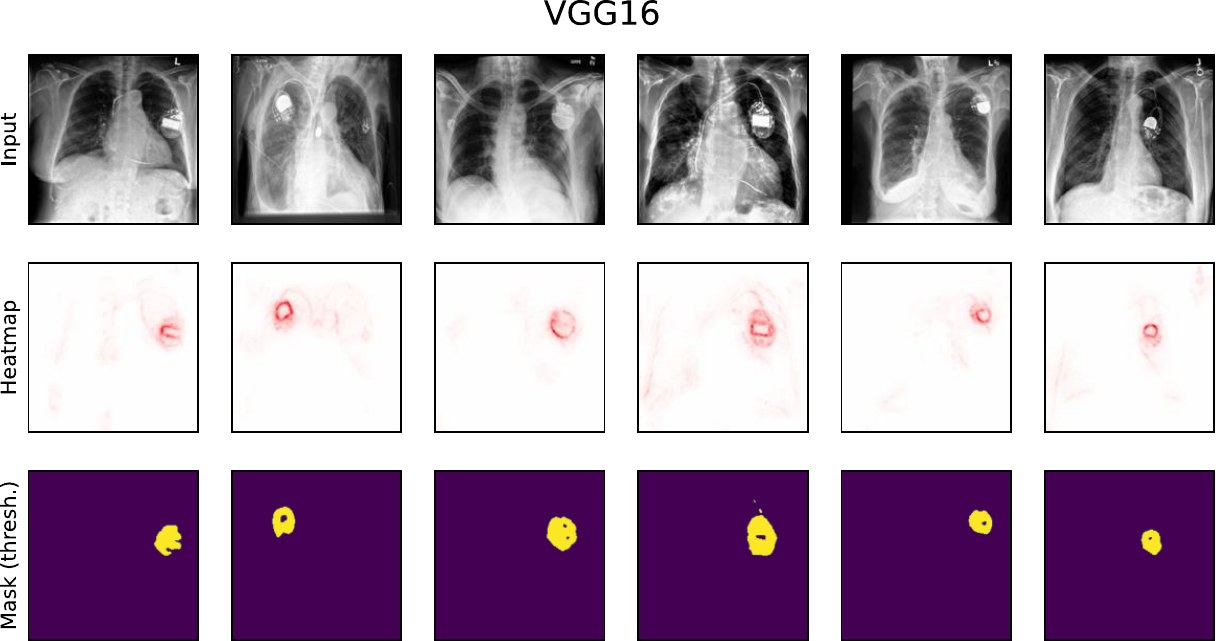}
    \quad\quad
    \includegraphics[width=.4\textwidth]{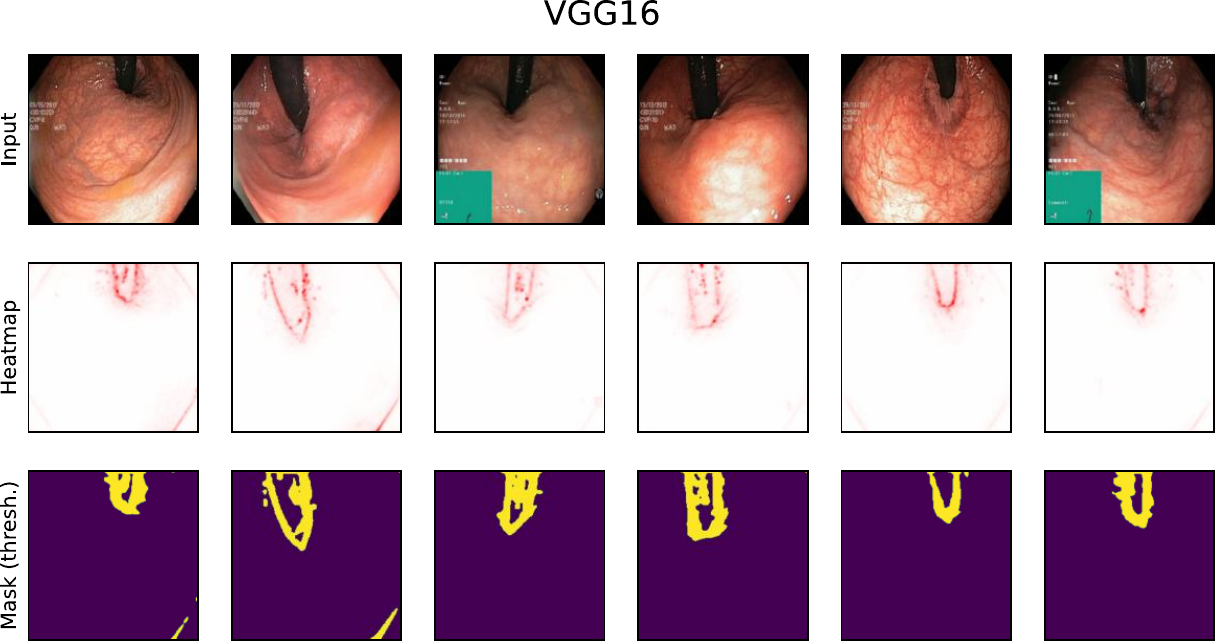}
    \caption{
    Examples for bias localization for the real-world artifacts \texttt{pacemaker} in CheXpert (\emph{left}) and \texttt{insertion tube} in HyperKvasir(\emph{right}). 
    The artifact concept is modeled using CAVs with activations after the $10^\text{th}$ and $13^\text{th}$ conv layers of VGG16 models, respectively. 
    We show input samples (\emph{$1^\text{st}$ row}), concept heatmaps (\emph{$2^\text{nd}$ row}), and binarized masks using Otsu's method (\emph{$3^\text{rd}$ row}).
    }
    \label{app:fig:exp_bias_localization_real_chexpert_kvasir}
\end{figure*}

\begin{figure*}[t!]
    \centering
    \includegraphics[width=.4\textwidth]{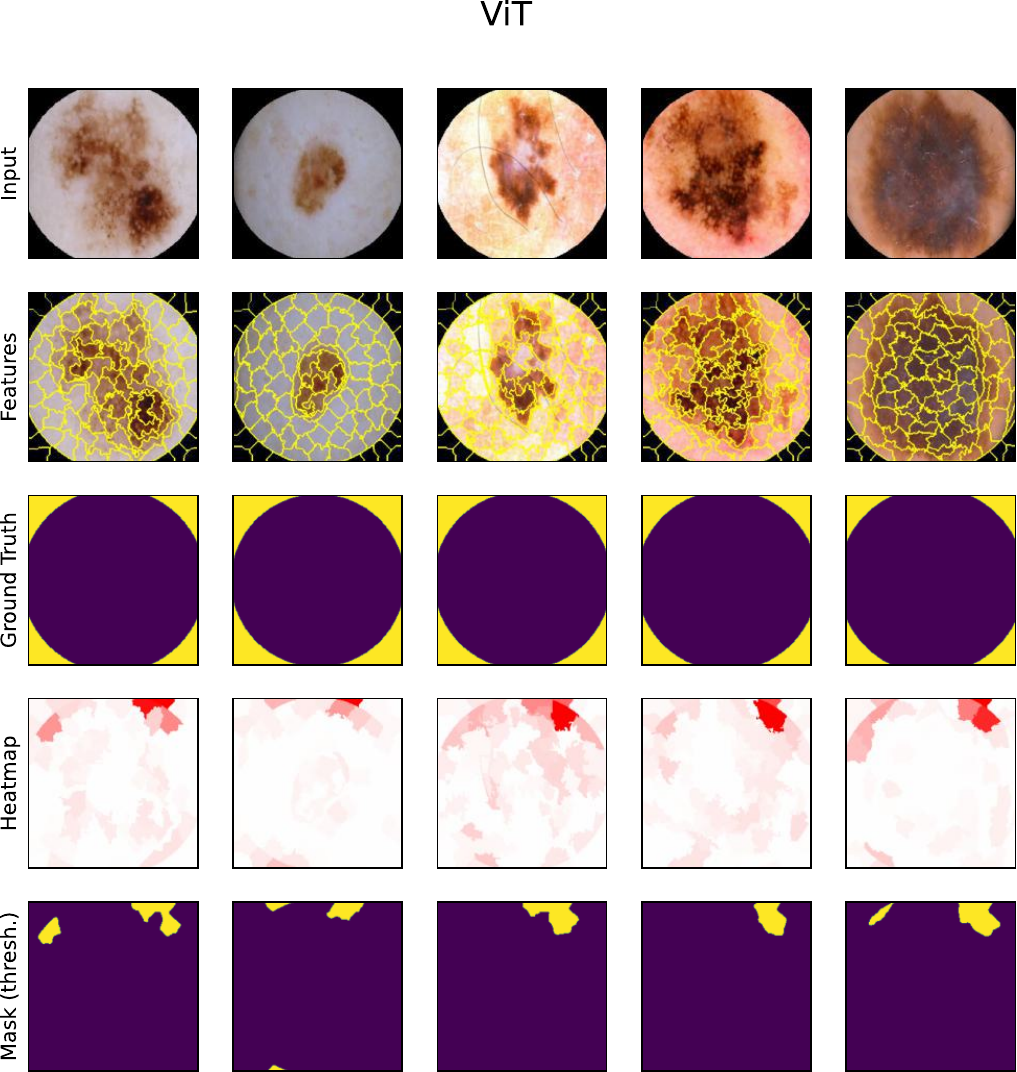}
    \quad\quad
    \includegraphics[width=.4\textwidth]{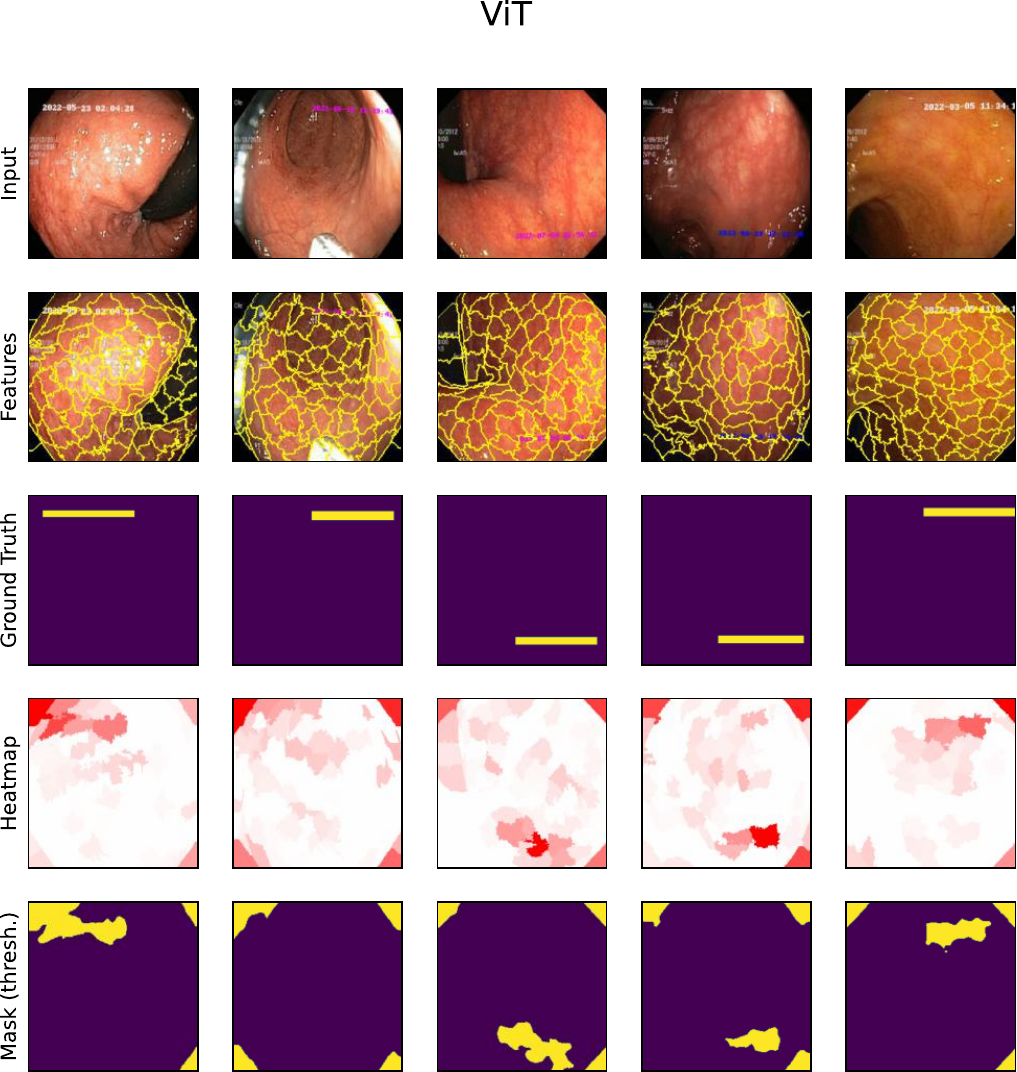}
    \caption{
    Examples for bias localization for the controlled artifacts \texttt{microscope} in ISIC2019 (\emph{left}) and \texttt{timestamp} in HyperKvasir(\emph{right}) using \gls{vit} models.
    The artifact concept is modeled using CAVs with activations in the last linear layer.
    \gls{shap}~\protect\citepapp{lundberg2017unified} 
    on superpixels computed via 
    \gls{slic}~\protect\citepapp{achanta2012slic}.
    We show input samples (\emph{$1^\text{st}$ row}), superpixels (\emph{$2^\text{nd}$ row}), ground truth localization (\emph{$3^\text{rd}$ row}), concept heatmaps (\emph{$4^\text{th}$ row}), and binarized masks using Otsu's method (\emph{$5^\text{th}$ row}).
    }
    \label{app:fig:exp_bias_localization_vit}
\end{figure*}

In addition, we show computed localizations for real-world artifacts \texttt{band-aid}, \texttt{ruler} (both ISIC2019), \texttt{pacemaker} (CheXpert), and \texttt{insertion tube} (HyperKvasir) in Figs.~\ref{app:fig:exp_bias_localization_real_isic}~and~\ref{app:fig:exp_bias_localization_real_chexpert_kvasir}.
These localizations can be valuable inputs to bias mitigation approaches, such as \gls{rrr}, as well as for the computation of metrics measuring the artifact reliance, \eg, the artifact relevance. 
We further show artifact localizations for the controlled artifacts \texttt{microscope} (ISIC2019) and \texttt{timestamp} (HyperKvasir) using \gls{vit} models in Fig.~\ref{app:fig:exp_bias_localization_vit}.
Due to the lack of reliable solutions for backpropagation-based local explanation methods for \gls{vit} architectures, we apply 
\gls{shap}~\citepapp{lundberg2017unified} 
on superpixels comuputed via 
\gls{slic}~\citepapp{achanta2012slic} 
to explain the output of the \gls{cav} to compute concept heatmaps.
Note, that this approach highly limits the flexibility of the bias localization due to the pre-defined image structure given by the pixel statistics forming superpixels. 

\begin{table*}[t]\centering
    \caption{
    Quantitative bias localization results for the controlled artifacts \texttt{microscope} (ISIC2019) and \texttt{timestamp} (HyperKvasir) with ground truth masks available. 
    We report artifact relevance and \gls{iou} for computed bias localizations using CAVs in different layers of VGG16 and ResNet50 models. 
    Higher scores are better. 
    }
        \begin{tabular}{c|c|cc|cc}
        \toprule
         &  & \multicolumn{2}{c|}{\texttt{microscope} (ISIC2019)} & \multicolumn{2}{c}{\texttt{timestamp} (HyperKvasir)}\\ 
       model & layer & \% relevance & IoU & \% relevance & IoU \\ 
        \midrule
         \multirow{4}{*}{\rotatebox[origin=c]{90}{VGG16}} &     Conv 4 &                                       0.50 &                0.14 &                                    0.18 &                0.26 \\
           &     Conv 7 &                                       0.45 &                0.11 &                                    0.38 &                0.58 \\
           &    Conv 10 &                                       0.45 &                0.11 &                                    0.53 &                0.60 \\
           &    Conv 13 &                                       0.32 &                0.08 &                                    0.44 &                0.52 \\
        \midrule
          \multirow{4}{*}{\rotatebox[origin=c]{90}{ResNet50}} &    Block 1 &                                       0.55 &                0.16 &                                    0.28 &                0.42 \\
           &    Block 2 &                                       0.49 &                0.12 &                                    0.63 &                0.68 \\
           &    Block 3 &                                       0.48 &                0.09 &                                    0.74 &                0.65 \\
           &    Block 4 &                                       0.40 &                0.04 &                                    0.69 &                0.58 \\
                \bottomrule
    \end{tabular}
    \label{tab:bias_localization}
    \end{table*}

\subsubsection{Bias Mitigation}
\label{app:sec:exp_bias_mitigation}
In the following, we provide additional bias mitigation results with our controlled artifacts, namely the \texttt{static noise} in ECG data (PTB-XL), the \texttt{timestamp} (HyperKvasir), and \texttt{miscroscope} (ISIC2019). 
Specifically, we compare the accuracy on a clean test set and a biased test set, where the artifact is artificially inserted into samples from all classes. 
Moreover, we measure the model's sensitivity towards the bias concept by computing (1) the percentage of relevance, measured via \gls{lrp} and (2) the TCAV score~\citep{kim2018interpretability}. 
The latter is based on the concept sensitivity measured via \gls{cav} $\bh$ and the latent feature use, computed as gradient of the prediction \wrt latent activations $\ba(\bx)$:

\begin{equation}
    \text{TCAV}_{\text{sens}}(\bx) = \boldsymbol\nabla_{\bm{a}} \bfunct(\bfeat(\bx)) \cdot \bh~.
\end{equation}

Given a subset containing the bias concept $\mathcal{X}_b=\{\bx_i \in \mathcal{X}\mid t_i=1\}$, the TCAV score is measured as the percentage of samples with positive sensitivity towards changes along the estimated concept direction $\bh$ and is computed as

\begin{equation}
    \text{TCAV} = 
    \frac{|\{\bx \in \mathcal{X}_b \mid \text{TCAV}_{\text{sens}}(\bx) > 0\}|}
    {|\mathcal{X}_b|}\,.
\end{equation}

Hence, a TCAV score of 0.5 indicates no concept reliance, while higher or lower scores are interpreted as negative and positive influence, respectively. 
We report $\Delta \text{TCAV}=|\text{TCAV}-0.5|$, where 0 indicates no sensitivity and higher values are interpreted as reliance on the artifact.

In Tab.~\ref{tab:mitigation_ecg}, we report results for PTB-XL, using the bias mitigation methods \gls{rrr}, \gls{rrclarc}, and \gls{pclarc}, in comparison with a Vanilla model without a bias-aware loss term.
All bias mitigation approaches, particularly \gls{clarc}-based methods, successfully increase the accuracy and decrease the false positive rate (for the attacked class LVH) on the biased test set. 
\gls{rrclarc} and \gls{pclarc} successfully reduce the artifact relevance, but have little impact on the TCAV score. 
For the former, instead of using a spatial mask localizing the artifact, we use temporal masks and compute the percentage of relevance put onto the attacked time span, \ie, the first second. 

Additional results for the vision datasets with controlled artifacts (HyperKvasir - \texttt{timestamp} $|$ ISIC2019 - \texttt{microscope} $|$ CheXpert - \texttt{brightness}) are shown in Tabs.~\ref{tab:mitigation_vgg}~and~\ref{tab:mitigation_resnet_full} for VGG16 and ResNet50 model architectures, respectively.
For the correction method \gls{rrr}, in addition to the mitigation results using ground truth masks (gt), we include results for model correction with automatically computed artifacts masks using relevance heatmaps (hm) and binarized masks using Otsu's method (bin).
For VGG16, \gls{rrr} outperforms other mitigation approaches for the correction of the easily localizable \texttt{timestamp} artifact in HyperKvasir. 
For ISIC2019 (see Tab.~\ref{tab:mitigation_vgg}), whereas \gls{pclarc} outperforms other approaches in terms of artifact relevance and $\Delta\text{TCAV}$, it has a poor performance in terms of accuracy on the biased test set. 
This can be explained by the fact that post-hoc model-editing does not allow the model to learn alternative prediction strategies when biases are unlearnt. 
For CheXpert, \gls{rrclarc} achieves the best results for biased accuracy and $\Delta\text{TCAV}$.
Note, that we refrained from reporting results for \gls{rrr} on CheXpert, as we consider the \texttt{brightness} artifact as unlocalizable in input space. 
Interestingly, the application of \gls{pclarc} leads to a drastic decrease in accuracy on the clearn dataset, which can be explained by collateral damage~\citep{bareeva2024reactive}, \ie, the suppression of related, yet relevant concept directions. 
For ResNet50 (see Tab.~\ref{tab:mitigation_resnet_full}), \gls{rrclarc} outperforms other bias mitigation approaches across all considered datasets and metrics, and achieves accuracies on the biased test sets that are close to those on the clean test sets. 
Lastly, in Tab.~\ref{tab:mitigation_vit} we report bias mitigation results for \gls{vit} models. Note that instead of reporting $\Delta\text{TCAV}$, we report $\text{TCAV}_{\text{sens}}$ results directly to obtain more fine-grained results. 
Especially \gls{rrr} is capable of mitigating the considered biases to some extent int terms of biased accuracy and artifact relevance and \gls{rrclarc} performs best in terms of $\text{TCAV}_{\text{sens}}$.

\begin{figure*}[t!]
    \centering
    \includegraphics[width=.4\textwidth]{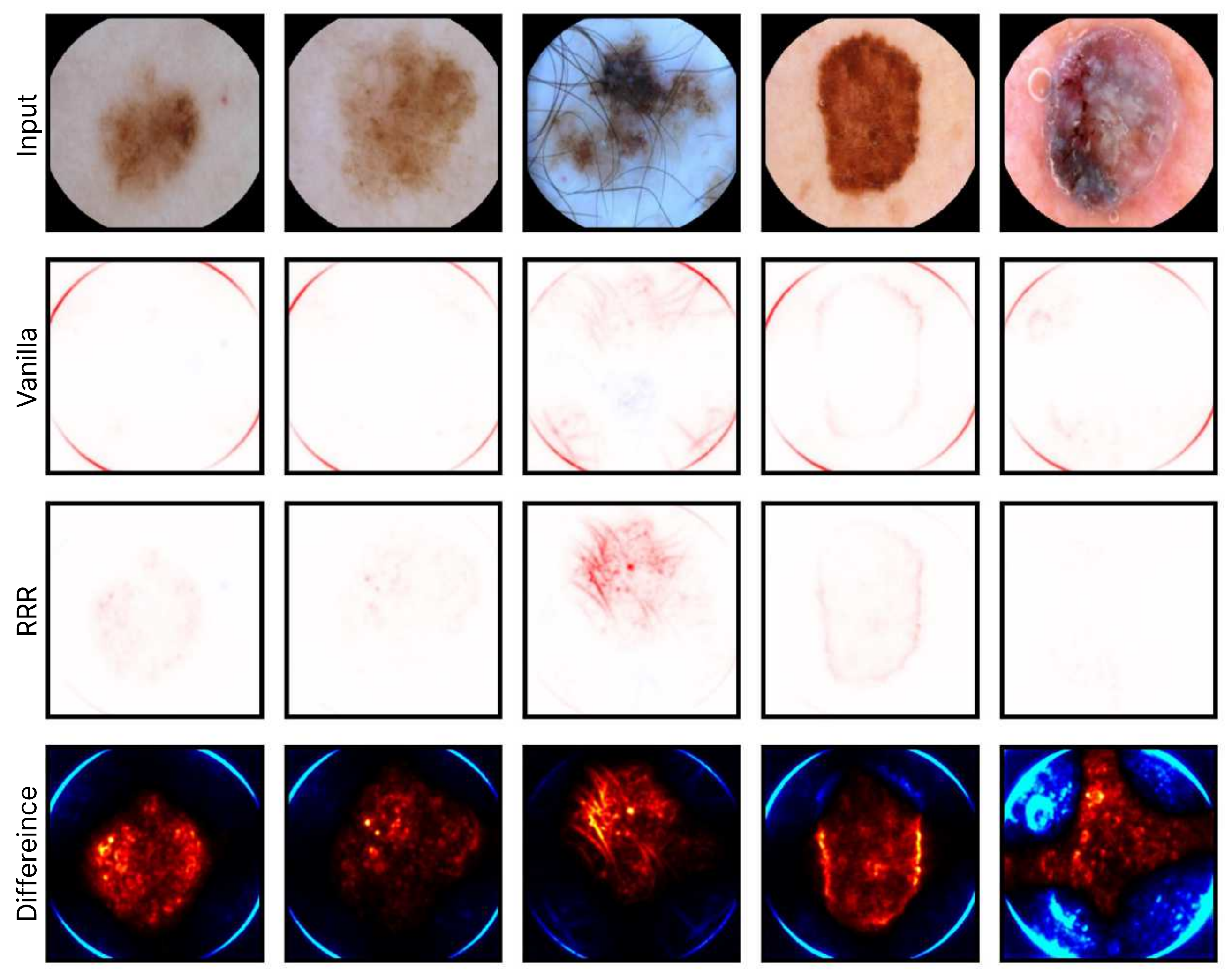}
    \quad\quad
    \includegraphics[width=.4\textwidth]{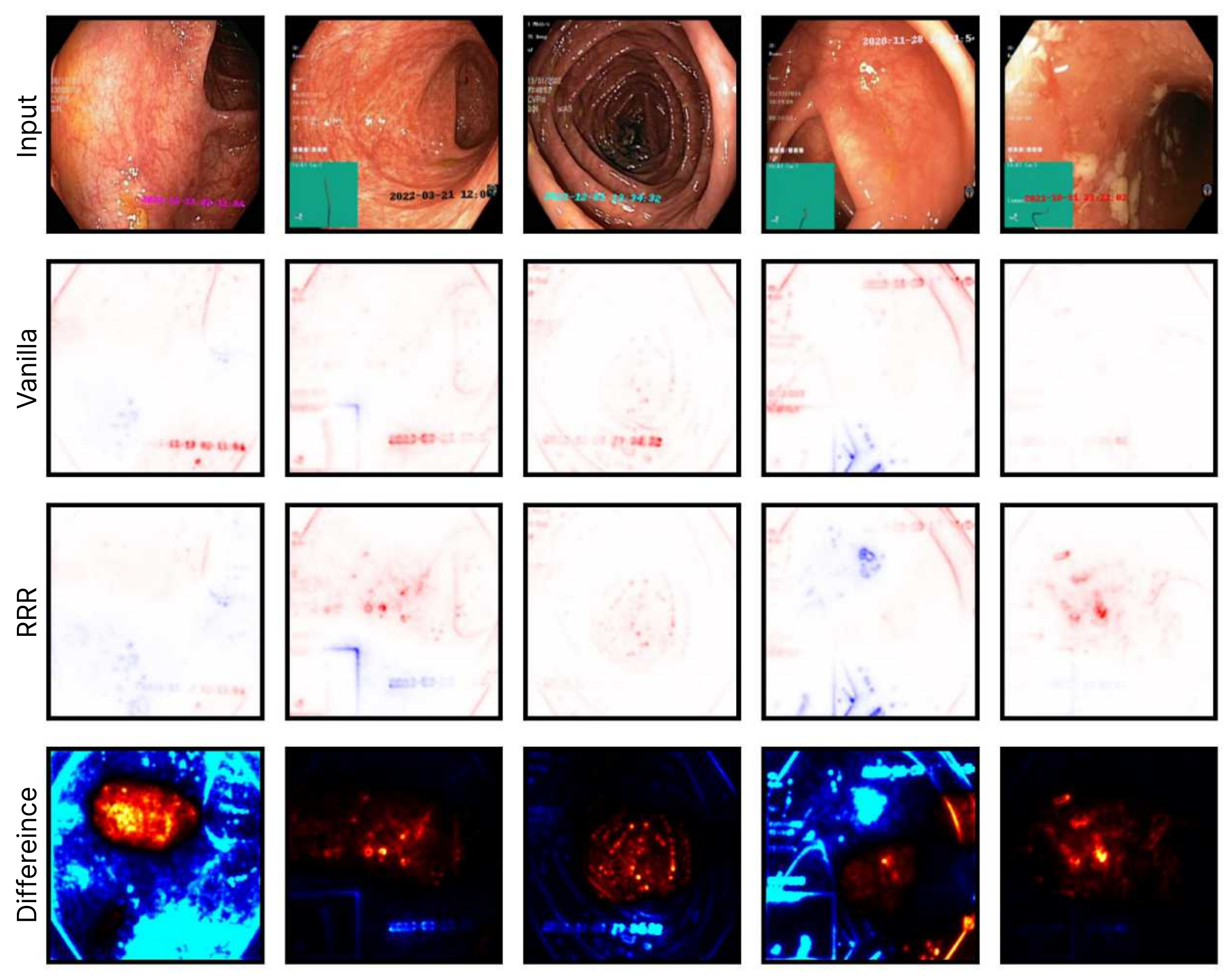}
    \caption{
    Local explanations, computed via \gls{lrp} and presented as relevance heatmaps, wihout (\emph{Vanilla}) and with (\emph{\gls{rrr}}) bias mitigation for the controlled versions of the ISIC2019 (\emph{left}) and HyperKvasir (\emph{right}) datasets. 
    Red indicates a positive and blue a negative impact towards the predicted class. 
    Whereas the Vanilla model is highly dependant on the inserted artifacts, \ie, \texttt{microscope} for ISIC2019 and \texttt{timestamp} for HyperKvasir, the corrected model (RRR) shifts the relevance from the artifact to medically relevant features. 
    We further plot a difference heatmap between the Vanilla and corrected (\gls{rrr}) models, with red and blue indicating increased and decreased relevance, respectively.
    }
    \label{app:fig:mitigation_qualitative}
\end{figure*}

\paragraph{Qualitative Bias Mitigation Results}
\label{app:sec:mitigation_qualitative}
In addition to the quantitative bias mitigation results discussed in Sec.~\ref{app:sec:exp_bias_mitigation}, we show qualitative findings in Fig.~\ref{app:fig:mitigation_qualitative}.
Specifically, we show local explanations as relevance heatmaps computed with \gls{lrp} for models corrected via \gls{rrr} in comparison with the Vanilla model for VGG16 models trained on the controlled versions of the ISIC2019 and HyperKvasir datasets. 
We selected test samples from the attacked classes, \ie, ``melanoma'' for ISIC2019 containing the \texttt{microscope} artifact and ``disease'' from HyperKvasir with the artificial \texttt{timestamp}.
Whereas the Vanilla models focus on the artifacts, the corrected models shift their attention to medically relevant features. 
This trend is confirmed by the quantitative results in Tab.~\ref{tab:mitigation_vgg}, with decreased artifact relevance for \gls{rrr} (gt) runs in comparison with the Vanilla model.

\paragraph{Hyperparameters}
Bias mitigation approaches utilizing a bias-aware loss term, such as \gls{rrr} and \gls{rrclarc}, require a hyperparameter $\lambda$, balancing between the classification loss and the auxiliary loss term, as defined in Eq.~\ref{eq:loss_rr}.
High values can lead to collateral damage by ``over-correcting'' the targeted bias and low values might have no impact.
Therefore, we test values $\lambda \in \{1^1, 5 \cdot 1^1, 1^2, 5 \cdot 1^2, ..., 5 \cdot 1^9\}$ for \gls{rrr}, and  $\lambda \in \{1^1, 5 \cdot 1^1, 1^2, 5 \cdot 1^2, ..., 5 \cdot 1^{12}\}$ for \gls{rrclarc}.
We pick the best performing $\lambda$ values on the validation set and report the bias mitigation results on an unseen test set. 
The selected $\lambda$ values are shown in Tab.~\ref{tab:mitigation_hyperparameters}.
As learning rate, we use the learning rate used during training (see Tab.~\ref{app:tab:training_details}) divided by $10$.

\begin{table*}[t]\centering
    \caption{Bias mitigation results with \gls{rrr}, \gls{rrclarc}, and \gls{pclarc} (plain and reactive) for XResNet1d50 model for static noise as controlled spurious correlation in the PTB-XL dataset.  We report accuracy and false positive rate (class: LVH) on a clean and biased test set, artifact relevance and $\Delta\text{TCAV}$ and arrows indicate whether high ($\uparrow$) or low ($\downarrow$) are better.}
\begin{tabular}{@{}
l@{\hspace{1em}}|
c@{\hspace{1.5em}}c@{\hspace{1.5em}}|
c@{\hspace{1.5em}}c@{\hspace{1.5em}}|
c@{\hspace{1.5em}}|
c@{\hspace{1.5em}}
}
\toprule
&\multicolumn{2}{c}{Accuracy $\uparrow$} & \multicolumn{2}{c}{FPR (LVH) $\downarrow$} && \\
Method &
clean & biased  & clean & biased & Art. rel.  $\downarrow$& $\Delta\text{TCAV}$ 
$\downarrow$\\ 
\midrule
    
\emph{Vanilla} &            ${0.96}$ &             ${0.94}$ &              ${0.00}$ &                 ${0.43}$ &              ${0.48}$ & ${0.06}$ \\
   RRR &            ${0.96}$ &             ${0.95}$ &              ${0.00}$ &                 ${0.32}$ &              ${0.57}$ & $\hspace{-1px}\mathbf{0.06}\hspace{-1px}$ \\
      RR-ClArC &            ${0.96}$ &             $\hspace{-1px}\mathbf{0.96}\hspace{-1px}$ &              ${0.00}$ &                 ${0.09}$ &              $\hspace{-1px}\mathbf{0.26}\hspace{-1px}$ & ${0.09}$ \\
       P-ClArC &            ${0.96}$ &             $\hspace{-1px}\mathbf{0.96}\hspace{-1px}$ &              ${0.00}$ &                 $\hspace{-1px}\mathbf{0.02}\hspace{-1px}$ &              ${0.28}$ & ${0.10}$ \\
        rP-ClArC &            ${0.96}$ &             $\hspace{-1px}\mathbf{0.96}\hspace{-1px}$ &              ${0.00}$ &                 $\hspace{-1px}\mathbf{0.02}\hspace{-1px}$ &              ${0.28}$ & ${0.10}$ \\       

    \bottomrule
\end{tabular}
    \label{tab:mitigation_ecg}
    \end{table*}

\begin{table*}[t]\centering
    \caption{Bias mitigation results with \gls{rrr} using computed heatmaps (hm), binarized versions (bin), and ground truth masks (gt), \gls{rrclarc}, and \gls{pclarc} (plain and reactive) for VGG16 models for controlled spurious correlations, specifically ISIC2019 (\texttt{microscope}) $|$ HyperKvasir (\texttt{timestamp}) $|$ CheXpert (\texttt{brightness}).  We report accuracy on a clean and biased test set, artifact relevance and $\Delta\text{TCAV}$ and arrows indicate whether high ($\uparrow$) or low ($\downarrow$) are better.}
\begin{tabular}{@{}
l@{\hspace{1em}}
c@{\hspace{1em}}
c@{\hspace{1em}}
c@{\hspace{1.5em}}
c@{\hspace{1.5em}}
}
        \toprule
Method &
Accuracy (clean) $\uparrow$ & Accuracy (biased) $\uparrow$ & Art. relevance  $\downarrow$& $\Delta\text{TCAV}$ 
$\downarrow$\\ 
\midrule
\emph{Vanilla} & ${0.85}$ $\,|\,$ ${0.97}$ $\,|\,$ ${0.83}$ & ${0.25}$ $\,|\,$ ${0.67}$ $\,|\,$ ${0.53}$ &  ${0.45}$ $\,|\,$ ${0.26}$ $\,|\,$  -  & ${0.34}$ $\,|\,$ ${0.34}$ $\,|\,$ ${0.31}$ \\
 RRR (hm) &  ${0.82}$ $\,|\,$ ${0.96}$ $\,|\,$ \hspace{6px}-\hspace{6px} &  ${0.70}$ $\,|\,$ ${0.94}$ $\,|\,$ \hspace{6px}-\hspace{6px} & ${0.09}$ $\,|\,$ ${0.07}$ $\,|\,$ - &  ${0.18}$ $\,|\,$ ${0.24}$ $\,|\,$ \hspace{6px}-\hspace{6px} \\
RRR (bin) &  ${0.81}$ $\,|\,$ ${0.97}$ $\,|\,$ \hspace{6px}-\hspace{6px} &  $\hspace{-1px}\mathbf{0.73}\hspace{-1px}$ $\,|\,$ ${0.95}$ $\,|\,$ \hspace{6px}-\hspace{6px} & ${0.08}$ $\,|\,$ ${0.07}$ $\,|\,$ - &  $\hspace{-1px}\mathbf{0.17}\hspace{-1px}$ $\,|\,$ ${0.26}$ $\,|\,$ \hspace{6px}-\hspace{6px} \\
 RRR (gt) &  ${0.82}$ $\,|\,$ ${0.97}$ $\,|\,$ \hspace{6px}-\hspace{6px} &  $\hspace{-1px}\mathbf{0.73}\hspace{-1px}$ $\,|\,$ $\hspace{-1px}\mathbf{0.96}\hspace{-1px}$ $\,|\,$ \hspace{6px}-\hspace{6px} & $\hspace{-1px}\mathbf{0.05}\hspace{-1px}$ $\,|\,$ $\hspace{-1px}\mathbf{0.06}\hspace{-1px}$ $\,|\,$ - &  ${0.20}$ $\,|\,$ ${0.30}$ $\,|\,$ \hspace{6px}-\hspace{6px} \\
       RR-ClArC & ${0.82}$ $\,|\,$ ${0.96}$ $\,|\,$ ${0.83}$ & ${0.64}$ $\,|\,$ ${0.95}$ $\,|\,$ $\hspace{-1px}\mathbf{0.82}\hspace{-1px}$ &  ${0.28}$ $\,|\,$ ${0.13}$ $\,|\,$  -  & ${0.22}$ $\,|\,$ ${0.12}$ $\,|\,$ $\hspace{-1px}\mathbf{0.03}\hspace{-1px}$ \\
        P-ClArC & ${0.42}$ $\,|\,$ ${0.78}$ $\,|\,$ ${0.77}$ & ${0.58}$ $\,|\,$ ${0.88}$ $\,|\,$ ${0.78}$ &  ${0.21}$ $\,|\,$ ${0.07}$ $\,|\,$  -  & ${0.27}$ $\,|\,$ $\hspace{-1px}\mathbf{0.05}\hspace{-1px}$ $\,|\,$ ${0.13}$ \\
rP-ClArC & ${0.84}$ $\,|\,$ ${0.97}$ $\,|\,$ ${0.83}$ & ${0.59}$ $\,|\,$ ${0.89}$ $\,|\,$ ${0.78}$ &  ${0.21}$ $\,|\,$ ${0.07}$ $\,|\,$  -  & ${0.27}$ $\,|\,$ $\hspace{-1px}\mathbf{0.05}\hspace{-1px}$ $\,|\,$ ${0.13}$ \\
    \bottomrule
\end{tabular}
    \label{tab:mitigation_vgg}
    \end{table*}

\begin{table*}[t]\centering
    \caption{Bias mitigation results with \gls{rrr} using computed heatmaps (hm), binarized versions (bin), and ground truth masks (gt), \gls{rrclarc}, and \gls{pclarc} (plain and reactive) for ResNet50 models for controlled spurious correlations, specifically ISIC2019 (\texttt{microscope}) $|$ HyperKvasir (\texttt{timestamp}) $|$ CheXpert (\texttt{brightness}).  We report accuracy on a clean and biased test set, artifact relevance and $\Delta\text{TCAV}$ and arrows indicate whether high ($\uparrow$) or low ($\downarrow$) are better.}
\begin{tabular}{@{}
l@{\hspace{1em}}
c@{\hspace{1em}}
c@{\hspace{1em}}
c@{\hspace{1.5em}}
c@{\hspace{1.5em}}
}
        \toprule
Method &
Accuracy (clean) $\uparrow$ & Accuracy (biased) $\uparrow$ & Art. relevance  $\downarrow$& $\Delta\text{TCAV}$ 
$\downarrow$\\ 
\midrule
    
\emph{Vanilla} & ${0.87}$ $\,|\,$ ${0.97}$ $\,|\,$ ${0.81}$ & ${0.28}$ $\,|\,$ ${0.62}$ $\,|\,$ ${0.44}$ &  ${0.55}$ $\,|\,$ ${0.51}$ $\,|\,$  -  & ${0.17}$ $\,|\,$ ${0.30}$ $\,|\,$ ${0.33}$ \\
 RRR (hm) &  ${0.87}$ $\,|\,$ ${0.97}$ $\,|\,$ \hspace{6px}-\hspace{6px} &  ${0.44}$ $\,|\,$ ${0.77}$ $\,|\,$ \hspace{6px}-\hspace{6px} & ${0.55}$ $\,|\,$ ${0.48}$ $\,|\,$ - &  ${0.16}$ $\,|\,$ ${0.29}$ $\,|\,$ \hspace{6px}-\hspace{6px} \\
RRR (bin) &  ${0.87}$ $\,|\,$ ${0.97}$ $\,|\,$ \hspace{6px}-\hspace{6px} &  ${0.43}$ $\,|\,$ ${0.76}$ $\,|\,$ \hspace{6px}-\hspace{6px} & ${0.54}$ $\,|\,$ ${0.46}$ $\,|\,$ - &  ${0.16}$ $\,|\,$ ${0.29}$ $\,|\,$ \hspace{6px}-\hspace{6px} \\
 RRR (gt) &  ${0.84}$ $\,|\,$ ${0.97}$ $\,|\,$ \hspace{6px}-\hspace{6px} &  ${0.51}$ $\,|\,$ ${0.82}$ $\,|\,$ \hspace{6px}-\hspace{6px} & ${0.53}$ $\,|\,$ ${0.45}$ $\,|\,$ - &  ${0.19}$ $\,|\,$ ${0.30}$ $\,|\,$ \hspace{6px}-\hspace{6px} \\
       RR-ClArC & ${0.86}$ $\,|\,$ ${0.97}$ $\,|\,$ ${0.82}$ & $\hspace{-1px}\mathbf{0.76}\hspace{-1px}$ $\,|\,$ $\hspace{-1px}\mathbf{0.96}\hspace{-1px}$ $\,|\,$ $\hspace{-1px}\mathbf{0.79}\hspace{-1px}$ &  $\hspace{-1px}\mathbf{0.42}\hspace{-1px}$ $\,|\,$ ${0.34}$ $\,|\,$  -  & $\hspace{-1px}\mathbf{0.00}\hspace{-1px}$ $\,|\,$ $\hspace{-1px}\mathbf{0.07}\hspace{-1px}$ $\,|\,$ $\hspace{-1px}\mathbf{0.00}\hspace{-1px}$ \\
        P-ClArC & ${0.82}$ $\,|\,$ ${0.83}$ $\,|\,$ ${0.72}$ & ${0.59}$ $\,|\,$ ${0.92}$ $\,|\,$ ${0.76}$ &  ${0.44}$ $\,|\,$ $\hspace{-1px}\mathbf{0.18}\hspace{-1px}$ $\,|\,$  -  & ${0.07}$ $\,|\,$ ${0.11}$ $\,|\,$ ${0.30}$ \\
rP-ClArC & ${0.87}$ $\,|\,$ ${0.97}$ $\,|\,$ ${0.81}$ & ${0.60}$ $\,|\,$ ${0.92}$ $\,|\,$ ${0.76}$ &  $\hspace{-1px}\mathbf{0.42}\hspace{-1px}$ $\,|\,$ ${0.19}$ $\,|\,$  -  & ${0.07}$ $\,|\,$ ${0.11}$ $\,|\,$ ${0.30}$ \\

    \bottomrule
\end{tabular}
    \label{tab:mitigation_resnet_full}
    \end{table*}
    
\begin{table*}[t]\centering
    \caption{Bias mitigation results with \gls{rrr} using ground truth masks (gt), \gls{rrclarc}, and \gls{pclarc} (plain and reactive) for \gls{vit} models for controlled spurious correlations, specifically ISIC2019 (\texttt{microscope}) $|$ HyperKvasir (\texttt{timestamp}) $|$ CheXpert (\texttt{brightness}).  We report accuracy on a clean and biased test set, artifact relevance computed and $\Delta\text{TCAV}$ and arrows indicate whether high ($\uparrow$) or low ($\downarrow$) are better.}
\begin{tabular}{@{}
l@{\hspace{1em}}
c@{\hspace{1em}}
c@{\hspace{1em}}
c@{\hspace{1em}}
c@{\hspace{1em}}
}
        \toprule
Method &
Accuracy (clean) $\uparrow$ & Accuracy (biased) $\uparrow$ & Art. relevance  $\downarrow$& $\text{TCAV}_{\text{sens}}$ 
$\downarrow$\\ 
\midrule
\emph{Vanilla} & ${0.81}$ $\,|\,$ ${0.93}$ $\,|\,$ ${0.80}$ & ${0.27}$ $\,|\,$ ${0.42}$ $\,|\,$ ${0.25}$ &  ${0.16}$ $\,|\,$ ${0.21}$ $\,|\,$  -  & ${1.47}$ $\,|\,$ ${0.45}$ $\,|\,$ ${0.38}$ \\
 RRR (gt) &  ${0.77}$ $\,|\,$ ${0.92}$ $\,|\,$ \hspace{6px}-\hspace{6px} &  $\hspace{-1px}\mathbf{0.59}\hspace{-1px}$ $\,|\,$ $\hspace{-1px}\mathbf{0.54}\hspace{-1px}$ $\,|\,$ \hspace{6px}-\hspace{6px} & $\hspace{-1px}\mathbf{0.07}\hspace{-1px}$ $\,|\,$ $\hspace{-1px}\mathbf{0.12}\hspace{-1px}$ $\,|\,$ - &  ${0.24}$ $\,|\,$ ${0.14}$ $\,|\,$ \hspace{6px}-\hspace{6px} \\
       RR-ClArC & ${0.81}$ $\,|\,$ ${0.93}$ $\,|\,$ ${0.80}$ & ${0.31}$ $\,|\,$ ${0.42}$ $\,|\,$ ${0.31}$ &  ${0.14}$ $\,|\,$ ${0.21}$ $\,|\,$  -  & $\hspace{-1px}\mathbf{0.00}\hspace{-1px}$ $\,|\,$ $\hspace{-1px}\mathbf{0.00}\hspace{-1px}$ $\,|\,$ $\hspace{-1px}\mathbf{0.00}\hspace{-1px}$ \\
        P-ClArC & ${0.81}$ $\,|\,$ ${0.93}$ $\,|\,$ ${0.80}$ & ${0.30}$ $\,|\,$ ${0.43}$ $\,|\,$ $\hspace{-1px}\mathbf{0.32}\hspace{-1px}$ &  ${0.14}$ $\,|\,$ ${0.21}$ $\,|\,$  -  & ${0.42}$ $\,|\,$ ${0.04}$ $\,|\,$ ${0.16}$ \\
rP-ClArC & ${0.81}$ $\,|\,$ ${0.93}$ $\,|\,$ ${0.80}$ & ${0.30}$ $\,|\,$ ${0.42}$ $\,|\,$ $\hspace{-1px}\mathbf{0.32}\hspace{-1px}$ &  ${0.14}$ $\,|\,$ ${0.21}$ $\,|\,$  -  & ${0.42}$ $\,|\,$ ${0.04}$ $\,|\,$ ${0.16}$ \\
    \bottomrule
\end{tabular}
    \label{tab:mitigation_vit}
    \end{table*}

\begin{table*}[t]\centering
    \caption{Chosen $\lambda$ values for bias mitigation runs with \gls{rrr} and \gls{rrclarc} for our controlled experiments with the datasets HyperKvasir, ISIC2019, CheXpert, and PTB-XL. We selected the best performing values on the validation set and report results on an unseen test set. 
    }
\begin{tabular}{@{}
l@{\hspace{1em}}|
c@{\hspace{.3em}}c@{\hspace{.3em}}c@{\hspace{.3em}}|
c@{\hspace{.3em}}c@{\hspace{.3em}}c@{\hspace{.3em}}|
c@{\hspace{.3em}}c@{\hspace{.3em}}c@{\hspace{.3em}}|
c@{\hspace{.5em}}
}
        \toprule
    &\multicolumn{3}{c}{HyperKvasir} & \multicolumn{3}{c}{ISIC2019} & \multicolumn{3}{c}{CheXpert} & PTB-XL\\
Approach & VGG & ResNet & ViT & VGG & ResNet & ViT & VGG & ResNet & ViT & XResNet\\ 

\midrule
\gls{rrr} (hm)  & $5 \cdot10^{3}$ & $5 \cdot10^{4}$& -           & $5 \cdot10^{3}$ & $5 \cdot 10^{5}$& -& - & -& - &- \\
\gls{rrr} (bin) & $5 \cdot10^{3}$ & $5 \cdot10^{4}$& -           & $10^{4}$ & $5 \cdot 10^{5}$& -& - & -& - &- \\
\gls{rrr} (gt)  & $10^{3}$ & $5 \cdot10^{4}$& $5 \cdot10^{2}$                  & $10^{4}$ & $10^{5}$& $10^{4}$& - & -& - &$5 \cdot10^{3}$ \\
\gls{rrclarc}   & $5 \cdot10^{9}$ & $10^{9}$& $10^{1}$                  & $10^{10}$ & $5 \cdot 10^{9}$& $10^{1}$& $10^{10}$ & $5 \cdot10^{12}$& $5 \cdot10^{1}$ &$5 \cdot10^{3}$ \\
\bottomrule
\end{tabular}
    \label{tab:mitigation_hyperparameters}
    \end{table*}
\clearpage 
\twocolumn
\bibliographystyleapp{plainnat}
\bibliographyapp{main} 

\end{document}